\documentclass{article}

\PassOptionsToPackage{numbers, compress}{natbib}
    \usepackage[preprint]{neurips_2022}

\usepackage[utf8]{inputenc} % allow utf-8 input
\usepackage[T1]{fontenc}    % use 8-bit T1 fonts
\usepackage[colorlinks,
            linkcolor=red,
            anchorcolor=red,
            citecolor=brown
            ]{hyperref}      % hyperlinks
\usepackage{url}            % simple URL typesetting
\usepackage{booktabs}       % professional-quality tables
\usepackage{amsfonts}       % blackboard math symbols
\usepackage{nicefrac}       % compact symbols for 1/2, etc.
\usepackage{microtype}      % microtypography
\usepackage{xcolor}         % colors
\usepackage{times}
\usepackage{soul}
\usepackage{graphicx}
\usepackage{amsmath}
\usepackage{amsthm}
\usepackage{amssymb}
\usepackage{booktabs}
\usepackage{algorithm}
\usepackage{algpseudocode}
\usepackage{subfig}
\usepackage{chngpage}
\usepackage{wrapfig}
\usepackage{enumitem}
\usepackage{multirow}
\usepackage{subfloat}
\usepackage{color}  %%%%%%
\usepackage{makecell}
\usepackage{bbding}
\usepackage{pifont}

\newtheorem{theorem}{Theorem}

\newtheorem{lemma}{Lemma}

\sodef\allcapsspacing{\upshape}{0.15em}{0.65em}{0.6em}%

\title{PROTO: Iterative Policy Regularized Offline-to-Online Reinforcement Learning 
}

\author{Jianxiong Li$^{1}$, Xiao Hu$^1$, Haoran Xu$^{1}$,\\ \textbf{Jingjing Liu}$^{1}$, \textbf{Xianyuan Zhan}\footnotemark[1]~~$^{1}$, \textbf{Ya-Qin Zhang}\footnotemark[1]~~$^{1}$\\
$^1$ Institute for AI Industry Research (AIR), Tsinghua University, Beijing, China \\
\texttt{li-jx21@mails.tsinghua.edu.cn}\\
\texttt{zhanxianyuan, zhangyaqin@air.tsinghua.edu.cn}\\
}

\begin{document}

\maketitle
\renewcommand{\thefootnote}{\fnsymbol{footnote}}
\footnotetext[1]{Corresponding Authors}

\begin{abstract}
Offline-to-online reinforcement learning (RL), by combining the benefits of offline pretraining and online finetuning, promises enhanced sample efficiency and policy performance. However, existing methods, effective as they are, suffer from suboptimal performance, limited adaptability, and unsatisfactory computational efficiency. We propose a novel framework, \textit{PROTO}, which overcomes the aforementioned limitations by augmenting the standard RL objective with an iteratively evolving regularization term. Performing a trust-region-style update, \textit{PROTO} yields stable initial finetuning and optimal final performance by gradually evolving the regularization term to relax the constraint strength.
% Instead of a fixed constraint set, resulting in both stable initial finetuning and optimal performances at the later finetuning stage\zhan{revise and add depth}. 
By adjusting only a few lines of code, \textit{PROTO} can bridge any offline policy pretraining and standard off-policy RL finetuning to form a powerful offline-to-online RL pathway, birthing great adaptability to diverse methods.
% be integrated into standard online off-policy RL finetuning and any pretraining approaches non-intrusively, yielding great generalizability to diverse methods. 
Simple yet elegant, \textit{PROTO} imposes minimal additional computation and enables highly efficient online finetuning. Extensive experiments demonstrate that \textit{PROTO} achieves superior performance over SOTA baselines, offering an adaptable and efficient offline-to-online RL framework.
% Undoubtedly, \textit{PROTO} offers a simple yet effective, generalizable and competitive Offline-to-Online approach.

% Overall, \texttt{PROTO} enjoys both stable, optimal, and efficient finetuning performances, which offers a simple yet competitive baseline for future works.

% imposes minimal \textbf{\texttt{LEOG}}assumptions on both pretraining and finetuning approaches, thereby enhancing extension-ability to diverse methods. By adjusting a few lines of code, standard off-policy RL finetuning approaches and pretraining methods, including offline RL and offline IL, can be integrated with PROTO. The proposed framework introduces negligible computational costs, achieving similar computational efficiency as standard off-policy RL approaches.

\end{abstract}
\vspace{-1pt}
\section{Introduction}
\vspace{-1pt}
% Reinforcement Learning (RL) provides the potential to obtain promising performances to reach or surpass human-level performances in many fields~\cite{mnih2015human, silver2017mastering, degrave2022magnetic}. Train a high-performing policy from scratch, however, requires a huge amount of expensive or dangerous online data collections, which limits the applicability in many real-world applications. Offline RL~\cite{levine2020offline} and offline imitation learning (IL)~\cite{randlov1998learning} provide alternatives to train policies without any environment interactions by exploiting only a fixed offline dataset. However, without further online interactions, offline RL and IL are largely restricted by offline data for their full resilience on given data.

Reinforcement learning (RL) holds the potential to surpass human-level performances by solving complex tasks autonomously~\cite{mnih2015human, silver2017mastering, degrave2022magnetic}. However, collecting a large amount of online data, especially the initial random explorations, can be expensive or even hazardous~\cite{nair2020awac}.
% Collecting large amounts of online data, however, can be expensive or hazardous.
Offline RL~\cite{levine2020offline} and offline imitation learning (IL)~\cite{randlov1998learning} offer alternatives to training policies without environment interactions, by exploiting fixed offline datasets generated by a behavior policy $\mu$. However, their performances are heavily limited by the quality and state-action space coverage of pre-existing offline datasets~\cite{kumar2019stabilizing, jin2021pessimism}. This largely inhibits learning-based decision-making approaches in real-world applications, where both sample efficiency and optimal performance are required~\cite{kiran2021deep, kormushev2013reinforcement, dulac2019challenges}.

Offline-to-online RL~\cite{nair2020awac, lee2022offline, zhang2023policy} has emerged as a promising solution, by pretraining a policy $\pi_0$ using offline RL/IL
% \footnote{In this paper, we are not restricted to offline RL pretraining but also consider offline IL pretraining. \jj{Why include this footnote? Seems redundant}\ljx{Previous works only consider offline RL and ignore offline IL pretraining. But, it is ok to delete this footnote.}}
and then finetuning with online RL. Ideally, offline-to-online RL can improve sample efficiency with favorable initialization for online RL. Further, by exploring more high-quality data, it overcomes the suboptimality of offline RL/IL caused by the over-restriction on a fixed suboptimal dataset. However, directly finetuning a pretrained policy often suffers from severe (even non-recoverable) performance drop at the initial finetuning stage, caused by distributional shift and the over-estimation error of value function at out-of-distribution (OOD) regions~\cite{nair2020awac, lee2022offline, uchendu2022jump}.

Existing works typically adopt conservative learning to alleviate this initial performance drop, which has three major drawbacks: $1)$ \textit{Suboptimal performance.} The majority of current methods introduce policy constraints to combat performance drop due to distributional shift~\cite{nair2020awac, iql, zhang2023policy, zheng2023adaptive}. Optimizing policy with an additional constraint, however, impedes the online learning process and can cause non-eliminable suboptimality gap~\cite{kumar2019stabilizing, wusupported, li2022distance}. $2)$ \textit{Limited adaptability.} Another way is to initialize online RL with pessimistic value functions to alleviate the over-estimation error accumulations~\cite{lee2022offline, nakamoto2023cal, hong2023confidenceconditioned}, which can only apply to CQL-style~\cite{kumar2020conservative} pretraining and lacks adaptability to diverse pretraining frameworks. $3)$ \textit{Computational inefficiency.} Moreover, some works require ensemble models to obtain near-optimal performance~\cite{lee2022offline}, which inevitably introduces tremendous computational costs and is unscalable to larger models.

In this paper, we propose a generic and adaptive framework, \textit{iterative Policy Regularized Offline-To-Online RL (PROTO)}, which incorporates an iterative policy regularization term into the standard RL objective. Performing a trust-region-style update~\cite{schulman2017proximal, schulman2015trust, zhuang2023behavior}, our method encourages the finetuning policy to remain close to $\pi_k$ (policy at last iteration, with $\pi_0$ being the pretrained policy). Compared to existing methods, \textit{PROTO} adopts appropriate conservatism to overcome the initial performance drop, while gradually relaxing excessive restrictions by casting constraint on an evolved $\pi_k$ rather than the fixed $\pi_0$, which leads to stable and optimal finetuning performance. Moreover, theoretical analysis proves that the introduced iterative regularization term induces no suboptimality and hence is far more \textit{optimistic} compared to previous policy constraints that typically cause suboptimal performance due to over-conservatism. %Therefore, we dub our method as \textit{PROTO: Let It Go!} \jj{Name is a bit strange and tacky. Why not simple 'PROTO'?}, \ljx{PROTO seems too simple to catch the eyes, but it is ok to use PROTO if PROTO is a bit strange and tacky. :)} \jj{PROTO and Let it go are too common phrases, and seems a bit stretchy here. How about PROTO until we reach a better acronym?} 
Therefore, \textit{PROTO} recognizes the necessity of giving enough freedom to finetuning in order to obtain near-optimal policies. It imposes minimal assumptions on pretraining and finetuning methods, allowing for seamless extension to diverse methods accomplished by adding just a few lines of code to standard off-policy RL finetuning. Simple yet effective, \textit{PROTO} achieves state-of-the-art performance on D4RL benchmarks~\cite{fu2020d4rl} and introduces negligible computational costs, retaining high computational efficiency on par with standard off-policy RL approaches and offering a competitive offline-to-online RL framework for future work.

% At last, we conduct extensive experiments to demonstrate the efficiency of \textit{PROTO}.  \textit{PROTO} also shows the strong generalizability to extend to diverse pretraining methods and enjoys high computational efficiency, which demonstrates the superiority of \textit{PROTO} over other baselines and offers a simple yet competitive baseline for future works.

% One natural solution of the above problems is offline-to-online RL, which pretrains a policy using offline RL/IL\footnote{In this paper, we are not limited to offline RL and also consider offline IL pretraining.} and then finetune it with online RL approaches. Hopefully, offline-to-online RL can not only provide a favorable initialization for online RL to improve sample-efficiency, but also mitigate the limitation induced by suboptimal offline dataset. Directly finetuning a pretrained policy, however, suffers severe and even non-recoverable performance drop caused by distributional shift at the initial finetuning stage~\cite{nair2020awac, lee2022offline, uchendu2022jump}.
\vspace{-1pt}
\section{Related Work}
\vspace{-1pt}
%Existing offline-to-online RL approaches typically alleviate the initial performance drop by adopting different types of conservatism such as policy constraint and pessimistic value initialization.

\textbf{Policy Constraint (PC).} The most straightforward way to mitigate the initial finetuning performance drop is to introduce policy constraints to combat the distributional shift. Existing methods, however, are over-conservative as they typically constrain the policy in a fixed constraint set (e.g., offline dataset support~\cite{kumar2019stabilizing}), which can lead to severely suboptimal performance~\cite{kumar2019stabilizing, li2022distance, wusupported}. \citet{nair2020awac} is the first offline-to-online RL approach that obtains stable finetuning performance. It introduces advantage weighted regression (AWR)~\cite{peng2019advantage} to extract policy, which is equivalent to implicitly constraining the policy \textit{w.r.t.} the replay buffer $\mathcal{B}$ that is updated by filling in newly explored transitions. Some offline RL approaches adopt AWR-style policy extraction to learn policies that can be directly utilized for online finetuning~\cite{iql, garg2023extreme, xiao2023the}. AWR, however, cannot be plugged into diverse online RL approaches non-intrusively, limiting its adaptability. Sharing similar philosophy, some works constrain the policy to stabilize training, but using a pluggable regularization~\cite{wusupported, zhao2022adaptive, zheng2023adaptive} such as simply adding one additional IL loss~\cite{wusupported}, which is easily adaptable to diverse online finetuning approaches. All these methods are over-conservative since the constraint on a mixed replay buffer $\mathcal{B}$ or a behavior policy $\mu$ may be severely suboptimal~\cite{kumar2019stabilizing, li2022distance, wusupported}. Some recent works partially reduce the over-conservatism by constraining on a potentially well-performing pretrained policy $\pi_0$~\cite{yu2023actorcritic, agarwalreincarnating, zhang2023policy}. However, $\pi_0$ may still be severely suboptimal when pretrained on a suboptimal offline dataset~\cite{kumar2019stabilizing, jin2021pessimism}. 

\textbf{Pessimistic Value Initialization (PVI).} One alternative to address performance drop is to initialize online RL with a pessimistic value function, to alleviate the side effect of overestimation errors. By doing so, the value function already attains low values at OOD regions and one can directly finetune online RL without introducing any conservatism, which has the potential to obtain near-optimal finetuning performance. \citet{lee2022offline} is the first to adopt pessimistic value initialization and introduces a balanced experience replay scheme. \citet{nakamoto2023cal} further improves upon~\cite{lee2022offline} by conducting a simple value surgery to ameliorate training instability caused by the over-conservative value initialization at OOD regions. However, these methods heavily rely on CQL~\cite{kumar2020conservative} pretraining framework, which inherit the main drawbacks of CQL such as being over-conservative and computationally inefficient~\cite{kostrikov2021offline, li2022distance}. Thus, when tasks are too difficult for CQL to obtain reasonable initialization, inferior performance may occur. Moreover, an ensemble of pessimistic value functions is generally required to better depict the manifold of OOD regions~\cite{lee2022offline}, which again inevitably imposes tremendous computational costs during both offline pretraining and online finetuning.

\textbf{Goal-Conditioned Supervised Learning (GCSL).}
A recent study~\cite{zheng2022online} considers the decision transformer (DT)~\cite{chen2021decision} finetuning setting and introduces entropy regularization to improve exploration. However, DT is formulated as a conditioned-supervised learning problem, which can be perceived as implicitly constraining policies on the replay buffer $\mathcal{B}$ similar to AWR, hence also suffering suboptimal performance when $\mathcal{B}$ is severely suboptimal.

\vspace{-1pt}
\section{PROTO RL Framework}
\vspace{-1pt}
\label{sec:Preliminaries}
\subsection{Problem Definition} 
We consider the infinite-horizon Markov Decision Process (MDP)~\cite{puterman2014markov}, which is represented by a tuple $\mathcal{M}:=\left<\mathcal{S}, \mathcal{A}, r, \rho, \mathcal{P}, \gamma\right>$, where $\mathcal{S}$ and $\mathcal{A}$ denote the state and action space, respectively. $r: \mathcal{S}\times\mathcal{A}\rightarrow\mathbb{R}$ represents a reward function, $\rho$ denotes initial distribution, $\mathcal{P}: \mathcal{S}\times\mathcal{A}\rightarrow\mathcal{S}$ is the transition kernel, and $\gamma\in(0, 1)$ is a discount factor. 
% We denote the discounted visitation distribution generated by a policy $\pi$ as $\rho_\pi:=\left(1-\gamma\right)\sum_0$

% \subsection{Online Reinforcement Learning}
Standard RL aims to learn a policy $\pi^*: \mathcal{S}\rightarrow\mathcal{A}$ that maximizes the expected discounted return $J(\pi)=\mathbb{E}\left[\sum_{t=0}^\infty \gamma^t r(s_t,a_t)|s_0\sim\rho, a_t\sim\pi(\cdot|s_t), s_{t+1}\sim\mathcal{P}(\cdot|s_{t}, a_{t})\right]$, \textit{i.e.}, $\pi^*\leftarrow\arg\max_\pi J(\pi)$.
% Directly optimizing this objective via policy gradient $\nabla J(\pi)$, however, may suffer exceptionally high variance and low sample-efficiency~\cite{kakade2001natural}. 
One popular approach to solving the above problem is approximate dynamic programming (ADP)~\cite{powell2007approximate}, which typically approximates the action-value function $Q^{\pi_k}(s, a)$
% , and optionally, the state-value function $V^\pi(s)$
of the policy $\pi_k$ at the last iteration by repeatedly applying the following policy evaluation operator $\mathcal{T}^{\pi_k}, k\in N$:
\begin{equation}
\label{equ:Q_operator}
    (\mathcal{T}^{\pi_k} Q)(s,a):=r(s,a)+\gamma\mathbb{E}_{s'\sim\mathcal{P}(\cdot|s,a), a'\sim{\pi_k}(\cdot|s')}\left[Q(s', a')\right]
\end{equation}
% where
% \begin{equation}
% \label{equ:V_operator}
%     V(s):=\mathbb{E}_{a\sim\pi_k(\cdot|s)}\left[Q(s,a)\right]
% \end{equation}

Then, standard actor-critic
% \footnote{We focus on actor-critic methods for their higher sample-efficiency compared to on-policy RL approaches.}
RL approaches introduce one additional policy improvement step to further optimize the action-value function $Q^{\pi_k}(s,a)$~\cite{chenrandomized, fujimoto2018addressing, haarnoja2018soft, lillicrap2015continuous}:
\begin{equation}
\label{equ:original_policy}
    \pi_{k+1}\leftarrow\arg\max_\pi \mathbb{E}_{a\sim\pi(\cdot|s)}[Q^{\pi_k}(s,a)]
\end{equation}
In high-dimensional or continuous space, $Q^{\pi_k}$ is generally learned by enforcing the single-step Bellman consistency, \textit{i.e.}, $\min_Q J^{\pi_k}(Q)=\frac{1}{2}\mathbb{E}_{(s,a,s')\sim\mathcal{B}}\left[(\mathcal{T}^{\pi_k} Q-Q))(s,a)\right]^2$, where  $\mathcal{B}$ is a replay buffer that is updated by filling in new transitions during the training process.
% and $\min_V \frac{1}{2}\mathbb{E}_{(s,a)\sim\mathcal{D}}\left[(\mathcal{T}^\pi V-V))(s)\right]^2$. 
The policy improvement step is also performed on this replay buffer, \textit{i.e.}, $\pi_{k+1}\leftarrow\arg\max_\pi\mathbb{E}_{s\sim\mathcal{B},a\sim\pi(\cdot|s)}\left[Q^{\pi_k}(s,a)\right]$.

\subsection{Offline-to-Online RL}
% Online RL algorithms~\cite{haarnoja2018soft, fujimoto2018addressing, chenrandomized, schulman2017proximal, mnih2015human} typically require a substantially large amount of initial random explorations to warm up, which is sample-inefficient (and potentially unsafe) in real-world applications~\cite{zhan2021deepthermal}. Offline RL~\cite{kumar2020conservative, iql, garg2023extreme, por, sql}/IL~\cite{randlov1998learning, xu2022discriminator, kostrikov2019imitation} provides an alternative to learn policy without environment interactions, but suffers from suboptimal performances caused by the over-restriction on a suboptimal offline dataset $\mathcal{D}$
% \jj{Specify what D is. Offline dataset?}.
Offline-to-online RL ensures favorable initialization for online RL~\cite{haarnoja2018soft, fujimoto2018addressing, chenrandomized, schulman2017proximal, mnih2015human} with a pretrained policy, meanwhile overcomes the suboptimality of offline RL~\cite{kumar2020conservative, iql, garg2023extreme, por, sql, xu2021offline, li2023mind} or IL~\cite{randlov1998learning, xu2022discriminator, kostrikov2019imitation} by exploring more high-quality data with online finetuning.
\begin{table}[t]
    \centering
    \addtolength{\leftskip} {-0.3cm}
    \footnotesize
    \setlength{\tabcolsep}{2pt}
    \caption{\small Comparison of existing offline-to-online RL methods. See Table~\ref{tab:related_work} in Appendix for a more detailed comparison of other offline-to-online RL methods. $\mu$: behavior policy that generats the offline dataset~$\mathcal{D}$.  $\mathcal{B}$: replay buffer. $\pi_0$:  pretrained policy.
    % \jj{Add 'Type', 'Method' to the first two rows?}\ljx{OK}
    %\jj{Explain beta, pie, etc.?} \ljx{beta, pie, etc have been explained in related work and intro, should we explain them more detailed here?} \jj{People might already forgot when Table 1 is mentioned?}\ljx{OK!}
        % of other \jj{RL?} methods.
    % \jj{Table is too wide. Move IQL to below AWAC? Use 'PT' 'FT' for pretraining and finetuning?}
    }
    \label{tab:related_work_main}
    \begin{tabular}{lccccccc}
    \toprule
       % \multirow{2}{*}[-2ex]{\quad \quad \quad \quad \quad \quad Method}
       \quad \quad \quad \quad \quad \quad Type & \multicolumn{3}{c}{PC} & \multicolumn{1}{c}{PVI} & \multicolumn{1}{c}{GCSL}\\
       \cmidrule(r){1-1}\cmidrule(r){2-4}\cmidrule(r){5-5}\cmidrule(r){6-6}
       \quad \quad \quad \quad \quad \ \ Method & SPOT~\cite{wusupported}& \thead{{{AWAC}}~\cite{nair2020awac} {{IQL}}~\cite{iql}} &{{PEX}}~\cite{zhang2023policy} & \thead{{{Off2On}}~\cite{lee2022offline}} &{{ODT}}~\cite{zheng2022online} \\
    \midrule
    {a. Constraint policy set} &$\mu$ & $\mathcal{B}$ &$\pi_0$ & No Constraint &$\mathcal{B}$\\
    {b. Stable and optimal policy learning} &\ding{55} &\ding{55} &\ding{55} &\Checkmark &\ding{55}   \\
    {c. Adaptable to diverse pretraining methods} &\Checkmark &\Checkmark &\Checkmark &\ding{55} &\ding{55}  \\
    {d. Adaptable to diverse finetuning methods} &\Checkmark &\ding{55} &\Checkmark &\Checkmark &\ding{55}  \\
    {e. Computationally efficient} &\Checkmark &\Checkmark &\Checkmark &\ding{55} &\ding{55}  \\
    \bottomrule
    % \multicolumn{6}{l}{\small \thead{(a):  policy set, (b):  (c):  \\ methods.  (d):  (e): .}}
    % \jj{Move the descriptions of (a) to (e) to the table}}}
    \end{tabular}
    \vspace{-10pt}
\end{table}
%\textbf{Stable yet optimal policy learning}. \
However, directly finetuning offline pretrained policy with online RL often suffers from severe performance drop caused by distributional shift and over-estimation error at OOD regions~\cite{nair2020awac, lee2022offline}. Thus, additional regularization is required to stabilize the finetuning process. Since optimizing policy with additional regularization can lead to suboptimal performance~\cite{kumar2019stabilizing, wusupported, li2022distance},  the primary goal of the offline-to-online RL pathway is to balance stability and optimality during online finetuning. This requires policy finetuning to be initially stable while avoiding excessive conservatism to achieve near-optimal policies. 

\begin{wrapfigure}{r}{5.6cm}
    \vspace{-8pt}
    \includegraphics[width=0.4\textwidth]{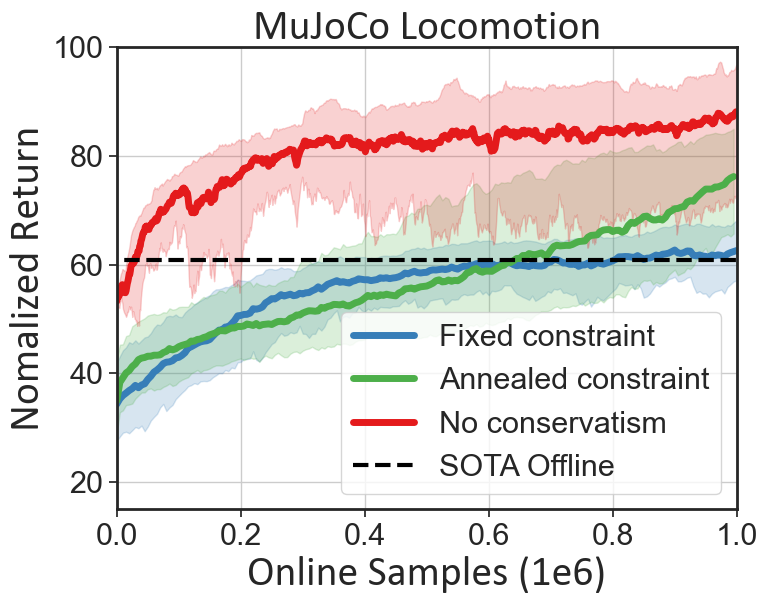}
    \caption{\small Aggregated learning curves of online finetuning with different policy constraints on 9 MuJoCo Locomotion tasks in D4RL benchmark~\cite{fu2020d4rl}. When policy constraints  are involved, severely suboptimal performance persists. Fixed constraint: IQL~\cite{iql}; Annealed constraint: Frozen (see Section~\ref{sec:ablation}); No conservatism: Off2On~\cite{lee2022offline}.}
    % \ljx{to be updated}\zhan{font size too small}}
    \label{fig:intro_optimal_gap}
    \vspace{-8pt}
\end{wrapfigure}

\textbf{Limitations of SOTA}. \ \
As summarized in Table~\ref{tab:related_work_main}, previous offline-to-online RL studies all directly borrow conservatism from offline RL to stabilize online finetuning. %However, we believe existing conservatism treatment in offline RL might not be best suited for offline-to-online RL.
Current methods, especially those based on policy constraint and goal-conditioned supervised learning, prioritize stability over policy performance optimality~\cite{nair2020awac, iql, zheng2022online},
% They typically apply commonly used constraints in offline RL such as constraining \textit{w.r.t} the behavior policy $\mu$~\cite{wusupported} or the slowly evolved replay buffer $\mathcal{B}$~\cite{iql, nair2020awac}. 
by keeping policy constraints fixed (e.g., behavior policy $\mu$ and pretrained policy $\pi_0$) or changed slowly during online finetuning (e.g., mixed replay buffer $\mathcal{B}$). Thus, if the initial constraints are severely suboptimal, they may restrain the online finetuning process to \textit{suboptimal performance with poor online sample efficiency}, as illustrated in Figure~\ref{fig:intro_optimal_gap}. 

Some works gradually anneal the constraint strength to alleviate over-conservatism~\cite{wusupported, agarwalreincarnating}. However, even with constraint annealing, suboptimal and slow online finetuning still occurs if the initial over-conservatism is too strong, as shown in Figure~\ref{fig:intro_optimal_gap}. Therefore, directly using fixed policy constraints may not be the best choice for offline-to-online RL.
Recent pessimistic value initialization method provides stable and optimal policy learning without introducing additional conservatism into online finetuning~\cite{lee2022offline, nakamoto2023cal, hong2023confidenceconditioned}, but at the expense of inefficiency, as it requires ensemble models to achieve reasonable performance with significant computational overhead. So far, it still remains a challenge how to strike a \textit{balance between stability and optimality} in a \textit{computationally efficient} way.

%The ideal offline-to-online RL should provide a universal solution that bridges a wide range of offline pretraining and online finetuning approaches to achieve the best possible performance and applicability.
% , resulting in powerful offline-to-online RL approaches. 
In addition, most previous studies focus on a specific pretraining or finetuning method~\cite{nair2020awac, iql, lee2022offline}, with \textit{limited adaptability} to diverse RL approaches. An ideal offline-to-online RL, however, should provide a universal solution that bridges a wide range of offline pretraining and online finetuning approaches to achieve the best possible performance and applicability. The simplest way to achieve this is to adopt a pluggable regularization, such as adding a BC term into the original policy loss~\cite{fujimoto2021minimalist, wusupported}, which can be easily integrated into diverse RL methods. However, such regularization often suffers from large suboptimality gaps~\cite{wusupported} due to lack of consideration on policy optimality.

%\textbf{Computational efficient}. \ 
%Lastly, %an important consideration for offline-to-online RL is achieving high performance with minimal computational costs, enabling agile and lightweight applications. However, pessimistic value initialization~\cite{lee2022offline}, while effective, often requires ensemble models to achieve reasonable performance, leading to significant computational overhead.

In summary, conservative designs commonly used in offline RL, such as policy constraints, pessimistic value regularization, and goal-conditioned supervised learning, inevitably suffer from suboptimal performance, limited adaptability, and inefficiency. Next, we introduce a new easy-to-use regularization that effectively retains stability while enabling adaptive and efficient online policy learning. 

\subsection{Iterative Policy Regularization}
\label{subsec:PROTO}
Trust-region update has gained some success in online RL~\cite{schulman2015trust, schulman2017proximal, nachum2018trustpcl} and thanks to its potential for unifying offline and online policy learning, has recently been extended to solve offline RL problems~\cite{zhuang2023behavior}. Inspired by this, we propose a generic and adaptive framework, \textit{iterative Policy Regularized
Offline-To-Online RL (PROTO)}, which augments the standard RL objective $J(\pi)$ with an \textit{Iterative Policy Regularization} term:
% will show in this section that we can easily meet all the desired properties via optimizing the following , which simply augments the standard RL objective $J(\pi)$ with an iteratively evolved regularization term:
\begin{equation}
\pi_{k+1}\leftarrow\arg\max_\pi\mathbb{E}\left[\sum_{t=0}^\infty\gamma^t\left(r(s_t,a_t)-\alpha\cdot \log\left(\frac{\pi(a_t|s_t)}{\pi_k(a_t|s_t)}\right)\right)\right], k\in N,
\label{equ:PROTO_objective}
\end{equation}
% \begin{wrapfigure}{r}{4cm}
%     \centering
%     \includegraphics[width=0.23\textwidth]{Figure/Main/illustration_PROTO.jpg}
%     \caption{\small {Illustration of \textit{PROTO}.}}
%     \label{fig:illustration_PROTO}
% \end{wrapfigure}

% where we treat $\hat\pi$ as the updated policy $\pi_k$ at last iteration and $f(x):=\log(x)$\footnote{$f(x)$ is not limited to $\log(x)$ as long as it comes from a type of $f$-divergence. For instance, $f(x):=x-1$ corresponds to $\chi^2$-divergence.}.
where $\pi_k$ is the policy at the last iteration, with $\pi_0$ being the pretrained policy.
This objective seeks to simultaneously maximize the reward and minimize the KL-divergence \textit{w.r.t.} the policy obtained at the last iteration $\pi_k$, which is equivalent to optimizing the original objective within the log-barrier of $\pi_k$,  hence can be interpreted as a trust-region-style learning objective~\cite{schulman2015trust, schulman2017proximal, nachum2018trustpcl, zhuang2023behavior}. 

Similar to the treatment in Max-Entropy RL~\cite{haarnoja2018soft, haarnoja2017reinforcement}, this \textit{Iterative Policy Regularized} MDP gives the following policy evaluation operator by simply adding a regularization term into Eq.~(\ref{equ:Q_operator}):
\begin{equation}
\label{equ:Q_operator_trust}
    (\mathcal{T}^{\pi_k}_{\pi_{k-1}}Q)(s,a):=r(s,a)+\gamma\mathbb{E}_{s'\sim\mathcal{P}(\cdot|s,a), a'\sim\pi_k(\cdot|s')}\left[Q(s',a')-\alpha\cdot \log\left(\frac{\pi_k(a'|s')}{\pi_{k-1}(a'|s')}\right)\right], k\in N^+,
\end{equation}
% where
% \begin{equation}
% \label{equ:V_operator_trust}
%     V(s)=\mathbb{E}_{a\sim\pi}\left[Q(s,a)-\alpha\cdot \log\left(\frac{\pi(a|s)}{\pi_k(a|s)}\right)\right],
% \end{equation}
The policy improvement step can be realized by adding a similar regularization term into Eq.~(\ref{equ:original_policy}):
\begin{equation}
\label{equ:policy_trust}
    \pi_{k+1}\leftarrow\arg\max_{\pi} \mathbb{E}_{a\sim\pi(\cdot|s)}\left[Q^{\pi_k}(s,a)-\alpha\cdot \log\left(\frac{\pi(a|s)}{\pi_k(a|s)}\right)\right], k\in N.
\end{equation}

 % Trust region policy optimization~\cite{schulman2015trust, schulman2017proximal, nachum2018trustpcl, zhuang2023behavior} restricts the policy in the trust region of the updated policy at last iteration, which has shown great success on stabilizing the policy learning process in online RL. Recently, trust region is also introduced to solve pure offline RL problem~\cite{zhuang2023behavior} and achieves superior performance. These successful applications on both online RL and offline RL show the potential probability to bridge offline pretraining to online finetuning with trust region technic.

Despite its simplicity, we will show that \textit{PROTO} can naturally balance the stability and optimality of policy finetuning in an effective manner, therefore is more suited for offline-to-online RL compared to existing conservative learning schemes that directly borrowed from offline RL methods.

\textbf{Stability and Optimality}. Performing a trust-region-style update, \textit{PROTO} constrains the policy \textit{w.r.t.} an iteratively evolving policy $\pi_k$, which smartly serves dual purposes by ensuring that:  $1)$ the finetuned policy remains close to the pre-trained policy $\pi_0$ during the initial finetuning stage, to avoid distributional shift; $2)$ gradually allowing the policy to deviate far from the potentially suboptimal constraint induced by $\pi_0$ at the later stage, to find the optima as long as it stays within the trust region. Therefore, this objective enables stable and optimal policy learning, which is different from and far more optimistic than existing methods with constraints on a potentially suboptimal and fixed $\mu$, $\pi_0$ or $\mathcal{B}$~\cite{wusupported, nair2020awac, iql, garg2023extreme, yu2023actorcritic, xiao2023the, zhao2022adaptive, zhang2023policy, agarwalreincarnating}. 

% \ljx{I added more theories here to provide a deeper understanding of the problem and prevent our work from appearing trivial. But now, section 3.3 seems too long. I'll keep thinking about how can we better organize the flow}

Furthermore, we present Theorem~\ref{theorem:gap}, which provides a concrete theoretical interpretation and proves that \textit{PROTO} principally enjoys both stable and optimal policy finetuning. Define $Q^*$ is the optimal value of optimal policy $\pi^*$. $\pi_0$ is the pretrained policy and $Q^0$ is its corresponding action-value. Let $v_{\rm max}^{\alpha}:=\frac{r_{\rm max}+\alpha{\rm ln}\mathcal{|A|}}{1-\gamma}, v_{\rm max}:=v_{\rm max}^0$, and $\epsilon_j$ is the approximation error of value function at $j$-th iteration. Assume that $\|Q^k\|_\infty\le v_{\rm max}, k\in N$, then we have (The proof is presented in Appendix~\ref{subsec:theorem_1_proof}):

\begin{theorem}
(Performance bound of \textit{Iterative Policy Regularization})
Define $Q^{k}$ as the action-value of policy $\pi_k$ obtained at $k$-th iteration by iterating Eq.~(\ref{equ:Q_operator_trust})-(\ref{equ:policy_trust}), then:
% (refer to Appendix~\ref{subsec:theorem_1_proof} for proof):
\begin{equation}
\label{equ:suboptimal_gap_trust_main}
    \|Q^*-Q^{k}\|_\infty\le{\frac{2}{1-\gamma}\left\|\frac{1}{k+1}\sum_{j=0}^{k}\epsilon_{j}\right\|_{\infty}}+{\frac{4}{1-\gamma}\frac{v_{\max}^{\alpha}}{k+1}}, k\in N.
\end{equation}
\label{theorem:gap}
% \vspace{-5pt}
\end{theorem}

% First, we review the performance of finetuning without any regularization (Lemma~\ref{lemma:stable_no_ipr}) and with fixed regularization (Lemma~\ref{lemma:fixed_gap}). 

For the RHS, the first term reflects how approximation error affects the final performance, and the second term impacts the convergence rate.
Note that the approximation error term in Eq.~(\ref{equ:suboptimal_gap_trust_main}) is the norm of average error, \textit{i.e.}, $\|\frac{1}{k+1}\sum_{j=0}^{k}\epsilon_j\|_\infty$, which might converge to 0 by the law of large numbers. Therefore, \textit{PROTO} will be less influenced by approximation error accumulations and enjoys stable finetuning processes. By contrast, the performance bound of finetuning without any regularization attains the following form (see Lemma~\ref{lemma:stable_no_ipr_appendix} in Appendix for detailed discussion):
\begin{equation}
    \|Q^*-Q^{k}\|_\infty\le\frac{2\gamma}{1-\gamma}\sum_{j=0}^k\gamma^{k-j}\|\epsilon_{j}\|_\infty+\frac{2}{1-\gamma}\gamma^{k+1}v_{\rm max}, k\in N.
    \label{equ:no_regularization}
\end{equation}

The error term $\sum_{j=0}^k\gamma^{k-j}\|\epsilon_j\|_\infty\ge0$ in Eq.~(\ref{equ:no_regularization}) cannot converge to 0 and initially decays slowly ($\gamma$ often tends to $1$, so $\gamma^k$ changes slowly initially). Therefore, directly finetuning without any regularization may result in severe instability due to the initial approximation error at OOD regions induced during offline pretraining. Previous methods typically introduce additional fixed regularization to stabilize finetuning~\cite{nair2020awac, iql}. However, fixed regularization might lead to a non-eliminable suboptimality gap in the form of (see Lemma~\ref{lemma:fixed_gap} in Appendix for detailed discussion):
\begin{equation}
    \|Q^*-Q^k\|\le\frac{\|Q^*-Q^*_\Pi\|_\infty}{1-\gamma}, k\in N,
    \label{equ:fixed_gap_main}
\end{equation}

where $Q^*_\Pi$ is the optimal action-value obtained at the constraint set $\Pi$. The RHS of Eq.~(\ref{equ:fixed_gap_main}) cannot converge to 0 unless $\Pi$ already contains the optimal policy~
\cite{kumar2019stabilizing, wusupported, li2022distance}, but the constraint set $\Pi$ typically only covers suboptimal policies due to the limited coverage of $\mathcal{B}, \mu$ or $\pi_0$.
%\jj{This seems redundant to the last paragraph}. \ljx{Thanks for pointing it. I revised it and please see if the latest version is ok.} 
Whereas, the RHS in Eq.~(\ref{equ:suboptimal_gap_trust_main}) can converge to 0 as $k\rightarrow\infty$, indicating that the \textit{PROTO} can converge to optimal as $k\rightarrow\infty$, which underpins the optimistic nature of \textit{PROTO}.

The comparison between Theorem~\ref{theorem:gap} and Eq.~(\ref{equ:no_regularization})-(\ref{equ:fixed_gap_main}) demonstrates that \textit{PROTO} serves as a seamless bridge between \textit{fixed policy regularization} and \textit{no regularization}, allowing for stability while retaining the optimality 
%\jj{Restraining optimality seems a bad thing?}\ljx{Sorry, I misuse retrain and it should be retain. This is a good thing, means that PROTO can balance stability and optimality.}
of finetuning performance. This indicates that \textit{Iterative Policy Regularization} offers a more reasonable level of conservatism for the offline-to-online RL setting compared to existing policy regularization that directly borrowed from offline RL or no regularization.

\textbf{Adaptability and Computational Efficiency.} \ \ 
% For the choice of online finetuning and offline pretraining methods, 
\textit{PROTO} bridges diverse offline RL/IL and online RL methods, offering a universal proto-framework for offline-to-online RL approaches. It imposes no assumption on how $\pi_0$ is pretrained, and can be non-intrusively incorporated into diverse off-policy RL finetuning methods~\cite{chenrandomized, fujimoto2018addressing, haarnoja2018soft, lillicrap2015continuous} by simply modifying several lines of code in the original actor-critic framework, according to Eq.~(\ref{equ:Q_operator_trust})-(\ref{equ:policy_trust}). 
% Such a great adaptability illustrates that \textit{PROTO} can serve as a prototype for constructing offline-to-online RL
In our experiments, unless otherwise specified, we pretrain policy using a recent SOTA offline RL method EQL~\cite{sql} (equivalent to XQL~\cite{garg2023extreme}) for its superior pretraining performances and finetune with SAC~\cite{haarnoja2018soft} for its high sample efficiency and superior performance among off-policy RL methods. In addition, calculating the additional regularization term introduces negligible computational cost compared to ensemble networks~\cite{lee2022offline} or transformer-based finetuning approaches~\cite{zheng2022online}, enabling agile and lightweight applications.

\subsection{Practical Implementation}
To further stabilize the finetuning process and meanwhile retain optimality, % We find that regulating the learning process with an overly fast-evolving $\pi_k$ may still cause instability. Therefore, 
we introduce Polyak averaging trick, a widely adopted technique in modern RL to address potential instability caused by fast target-value update~\cite{mnih2015human, haarnoja2018soft, fujimoto2018addressing, kumar2020conservative, fujimoto2021minimalist, chenrandomized}, by replacing $\pi_k$ with its delayed updates $\bar\pi_k$, \textit{i.e.}, $\bar\pi_k\leftarrow \tau\pi_k+(1-\tau)\bar{\pi}_{k-1}$. Here, $\tau\in(0, 1]$ is a hyper-parameter to control the update speed. As apparent, replacing $\pi_k$ with $\bar\pi_k$ retains optimal performance since it still allows large deviation from the pretrained policy (with a slower deviation speed). We also gradually anneal the $\alpha$ value with a linear decay schedule~\cite{wusupported, deng2021score, agarwalreincarnating} for the purpose of weaning off conservatism. Although introducing two hyper-parameters, we show in Appendix~\ref{subdec:hyperparameters} and Appendix~\ref{sec:ablation_appendix} that \textit{PROTO} is robust to changes in hyperparameters within a large range, and parameter tuning  can be reduced by adopting a non-parametric approach and setting the annealing speed as a constant.

\vspace{-1pt}
\section{Experiments}
\vspace{-1pt}
\label{sec:Experiments}
We conduct extensive experiments on MuJoCo locomotion, AntMaze navigation and Adroit manipulation tasks with D4RL~\cite{fu2020d4rl} datasets to demonstrate the stable and optimal policy learning, adaptability, and computational efficiency of \textit{PROTO} compared to state-of-the-art offline-to-online RL methods.
% investigate whether \textit{PROTO} can achieve both stable and optimal policy learning, adaptable to diverse approaches and is computational efficient.

% 1) Can PROTO achieve stable and optimal finetuning performance? 

% 2) Can PROTO generalize to diverse pretraining methods? 

% 3) Is PROTO computational efficient?

%\subsection{Evaluations on Benchmarks}
\vspace{-1pt}
\subsection{Baselines}
\vspace{-1pt}
%We first evaluate on benchmark tasks to demonstrate the stable and optimal finetuning performance of \textit{PROTO}.
%Specifically, 
We compare \textit{PROTO} with the following baselines: \textit{(i)} \textit{AWAC}~\cite{nair2020awac}: an offline-to-online method that implicitly constrains the finetuning policy \textit{w.r.t.} the replay buffer $\mathcal{B}$ using AWR-style~\cite{peng2019advantage} policy learning. \textit{(ii)} \textit{IQL}~\cite{iql}: a SOTA offline RL approach that can directly transfer to online finetuning since it implicitly constrains \textit{w.r.t.} the replay buffer $\mathcal{B}$ via AWR-style policy learning. \textit{(iii)} \textit{Off2On}~\cite{lee2022offline}: a SOTA offline-to-online RL method that uses an ensemble of pessimistic value functions together with a balanced experience replay scheme, but it is only applicable for CQL~\cite{kumar2020conservative} pretraining. \textit{(iv)} \textit{ODT}~\cite{zheng2022online}: a recent offline-to-online approach that is specifically designed for decision transformer~\cite{chen2021decision} pretraining and finetuning. \textit{(v)} \textit{PEX}~\cite{zhang2023policy}: a recent SOTA offline-to-online approach that adaptively constrains the finetuning policy \textit{w.r.t.} the pretrained policy $\pi_0$ by introducing a policy expansion and Boltzmann action selection scheme. \textit{(vi)} \textit{Offline}: performances of SOTA offline RL approaches without online finetuning that are adopted from~\cite{bai2021pessimistic, sql, iql, kumar2020conservative, li2022distance}.
% \textbf{\textit{(v)} \texttt{SAC}}~\cite{haarnoja2018soft}: train SAC from scratch, without offline pretraining. \textbf{\textit{(vi)} \texttt{SAC+Buffer}}~\cite{zheng2022online}: train SAC from scratch with its replay buffer initialized by offline dataset.
\vspace{-1pt}
\subsection{Main Results}
\begin{figure}[h]
    \centering
    \includegraphics[width=0.8\textwidth]{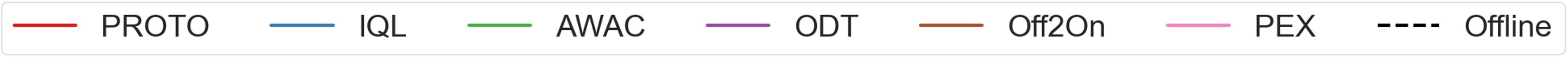}
    \includegraphics[width=0.32\textwidth]{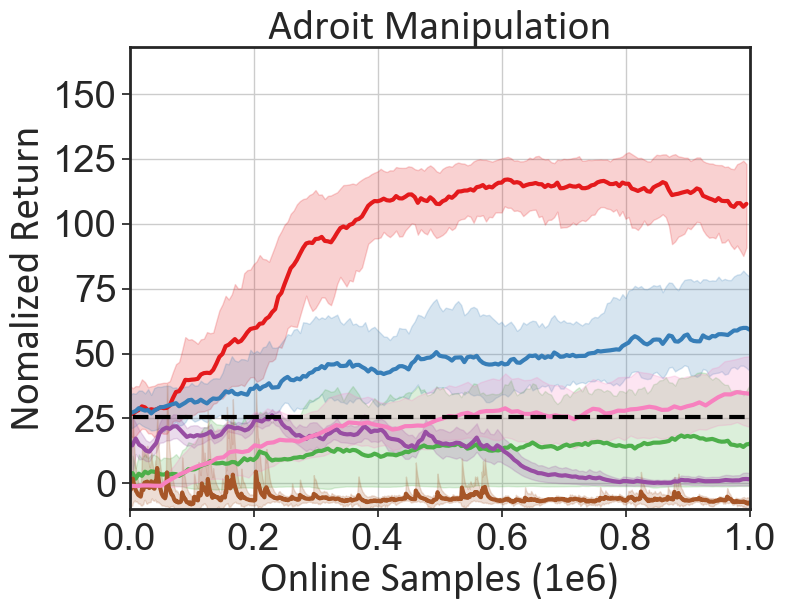}
    \includegraphics[width=0.32\textwidth]{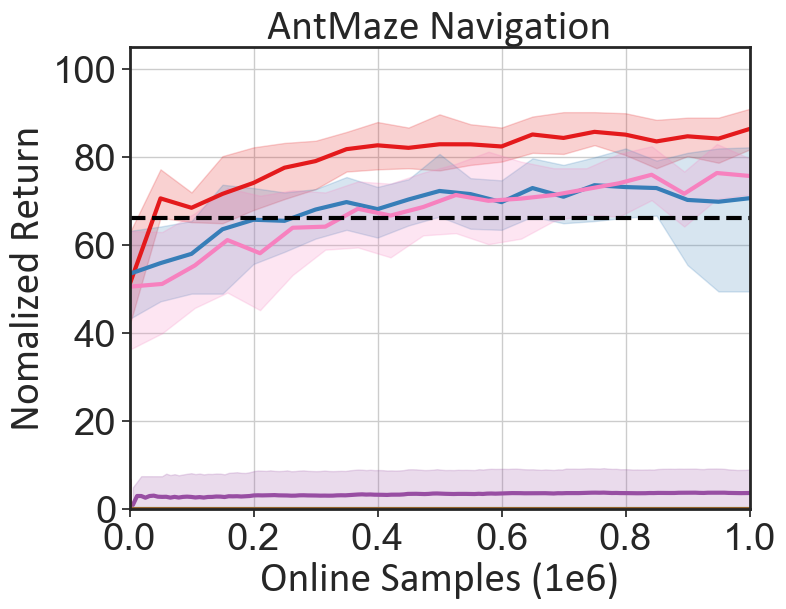}
    \includegraphics[width=0.32\textwidth]{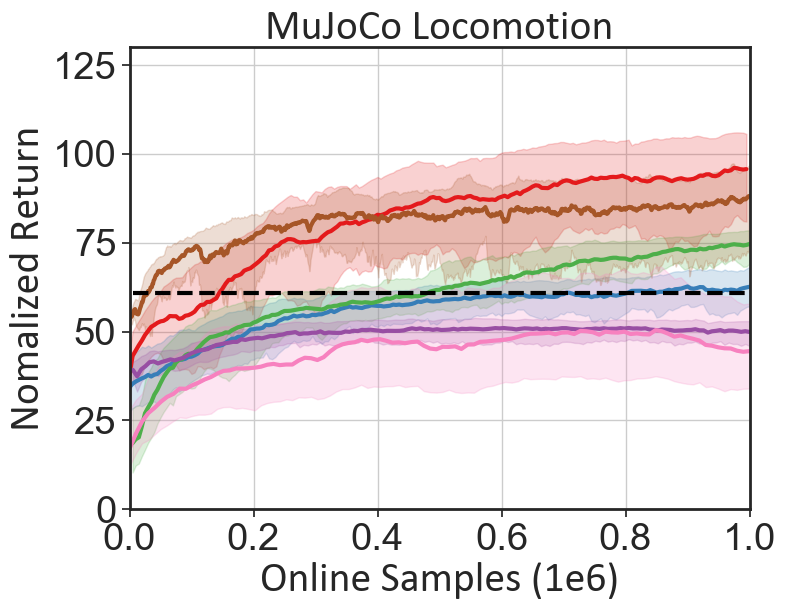}
    \caption{\small {Aggregated learning curves} of different approaches on Adroit manipulation, AntMaze navigation, and MuJoCo locomotion tasks from D4RL~\cite{fu2020d4rl} benchmark.}
    \label{fig:learning_curves_all_aggregation}
    \vspace{-5pt}
\end{figure}

\begin{figure}[h]
    \centering
    \includegraphics[width=0.8\textwidth]{Figure/Main/Learning_curves_main/legend_main.png.jpg}
    \includegraphics[width=0.24\textwidth]{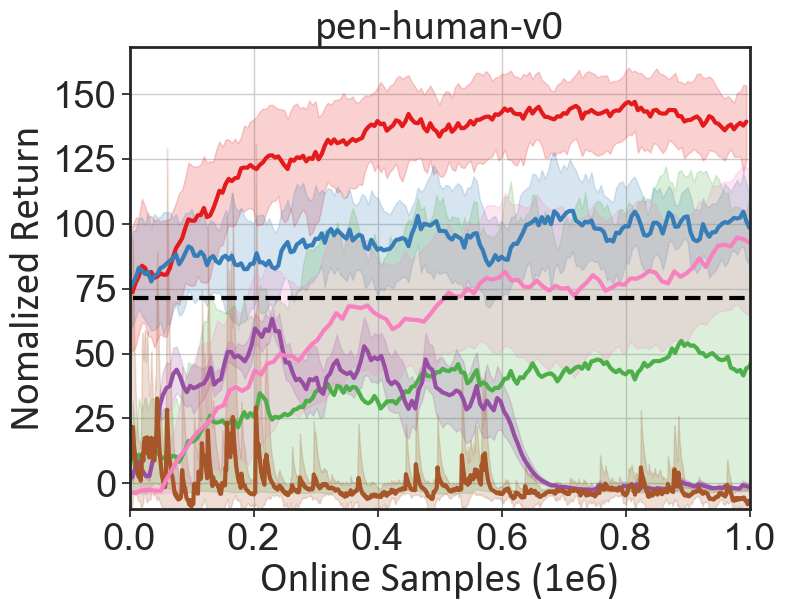}
    \includegraphics[width=0.24\textwidth]{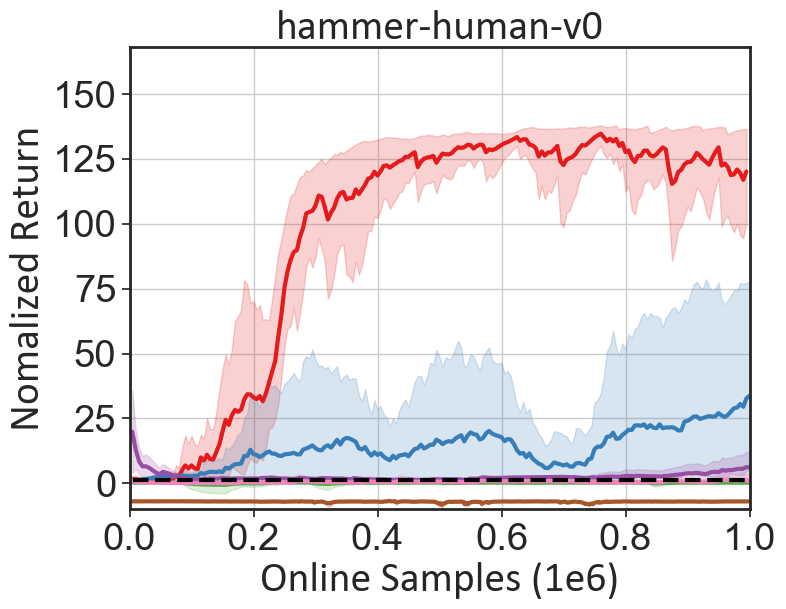}
    \includegraphics[width=0.24\textwidth]{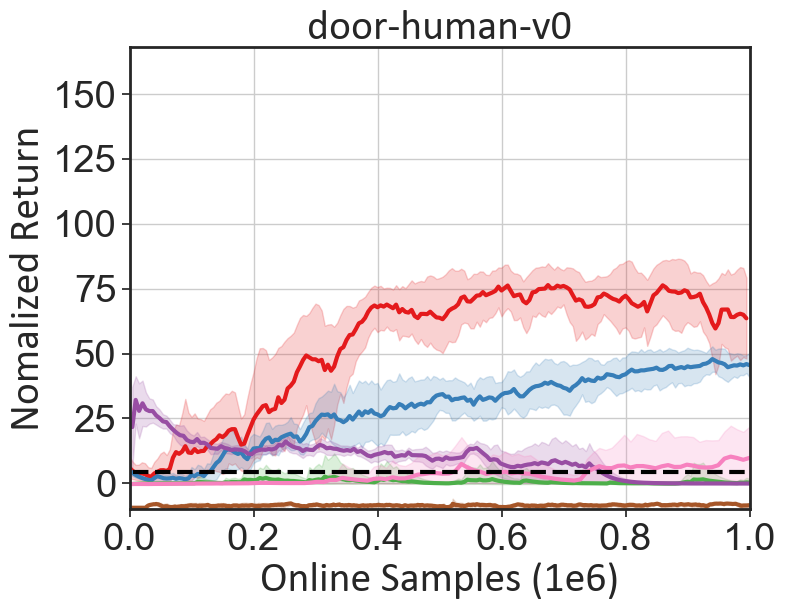}
    \includegraphics[width=0.24\textwidth]{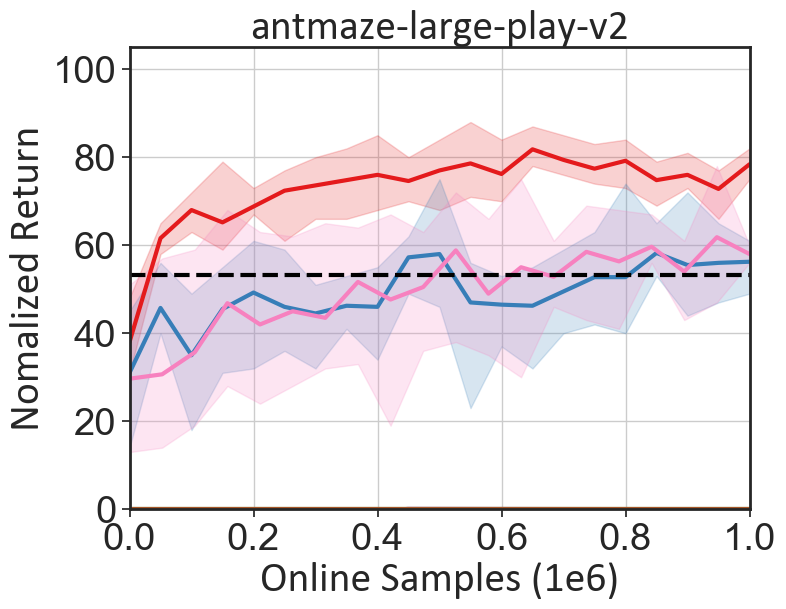}
    \includegraphics[width=0.24\textwidth]{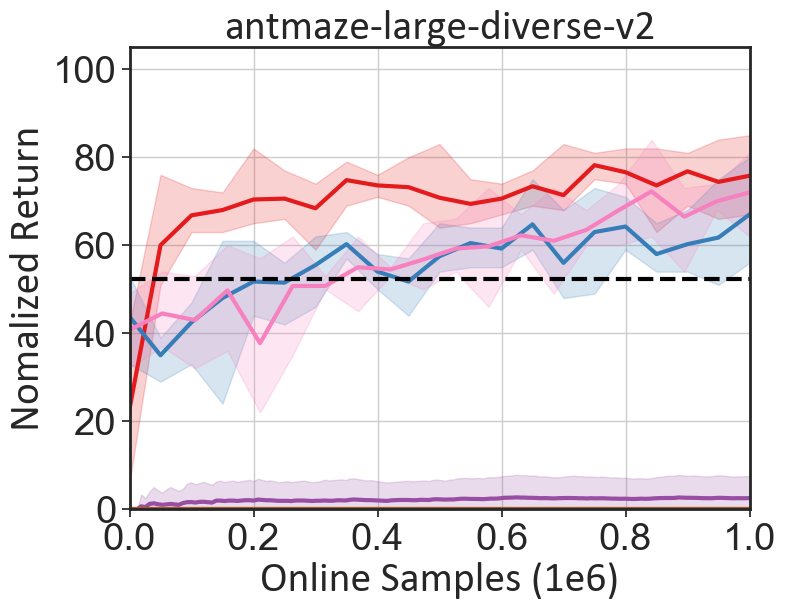}
    \includegraphics[width=0.24\textwidth]{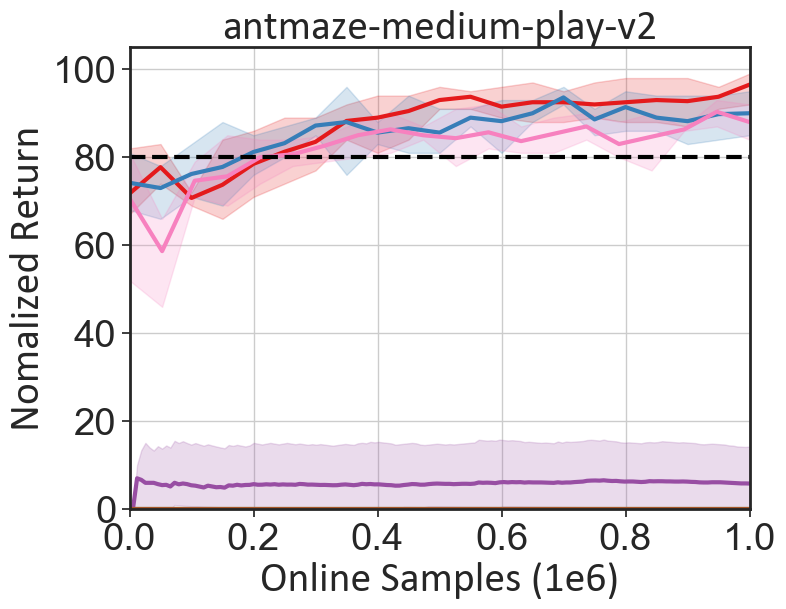}
    \includegraphics[width=0.24\textwidth]{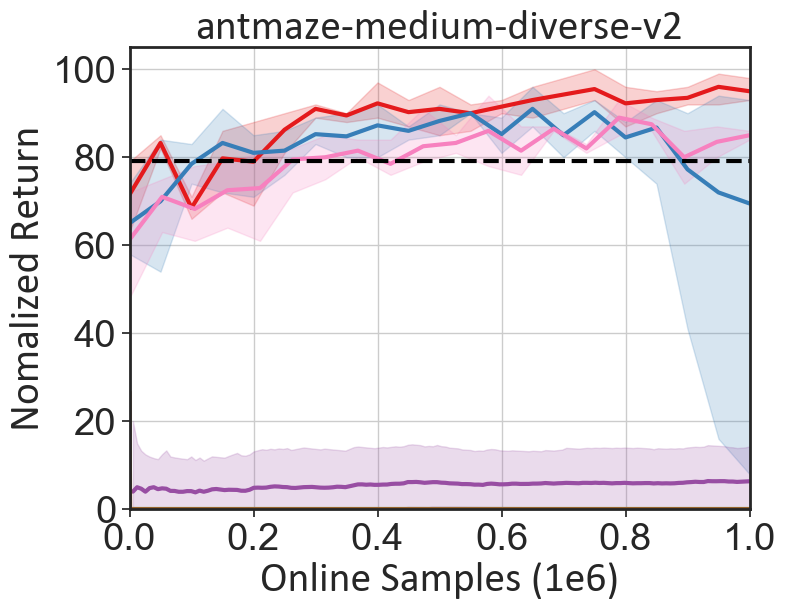}
    \includegraphics[width=0.24\textwidth]{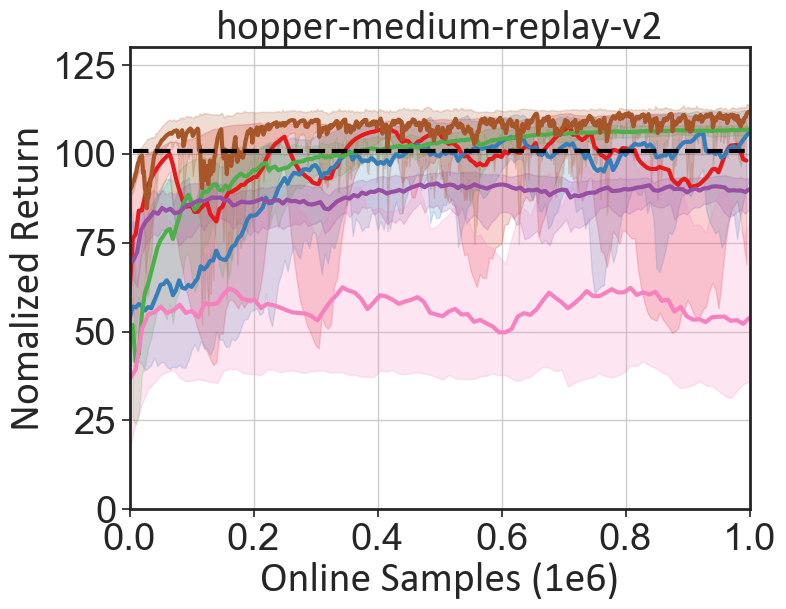}
    \includegraphics[width=0.24\textwidth]{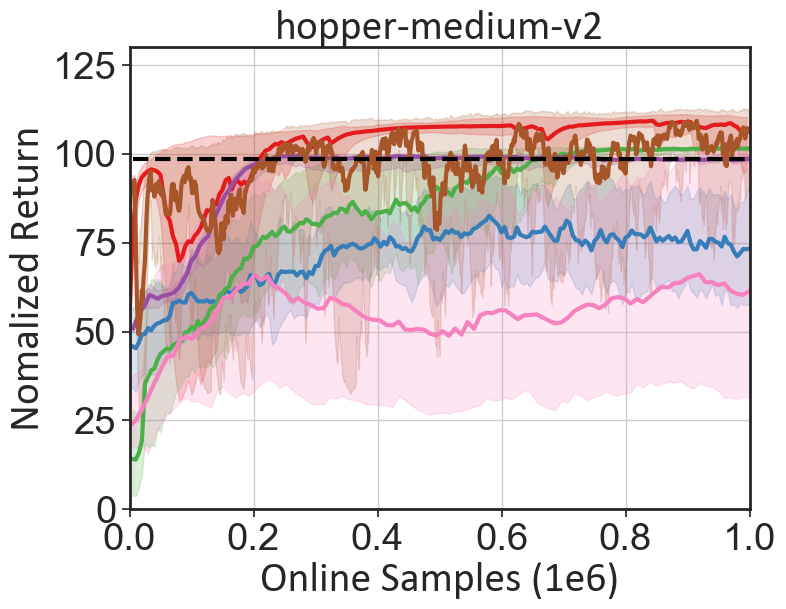}
    \includegraphics[width=0.24\textwidth]{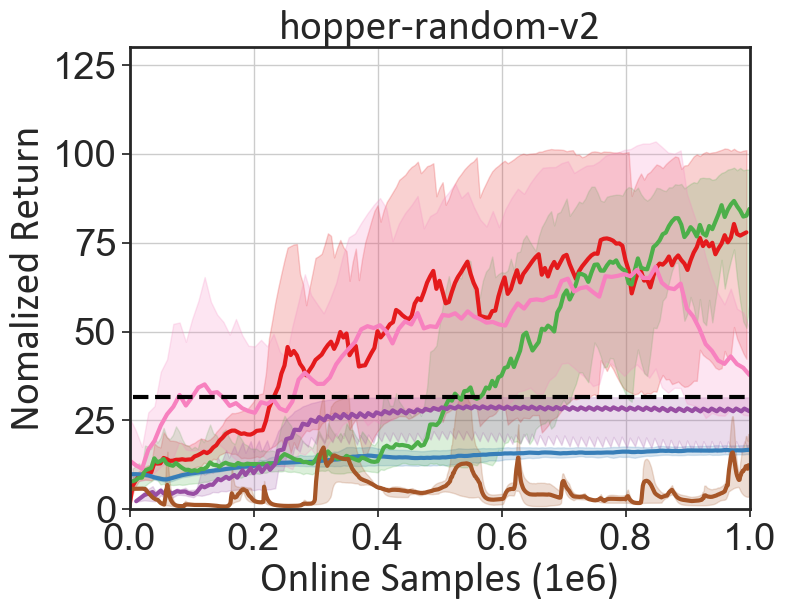}
    \includegraphics[width=0.24\textwidth]{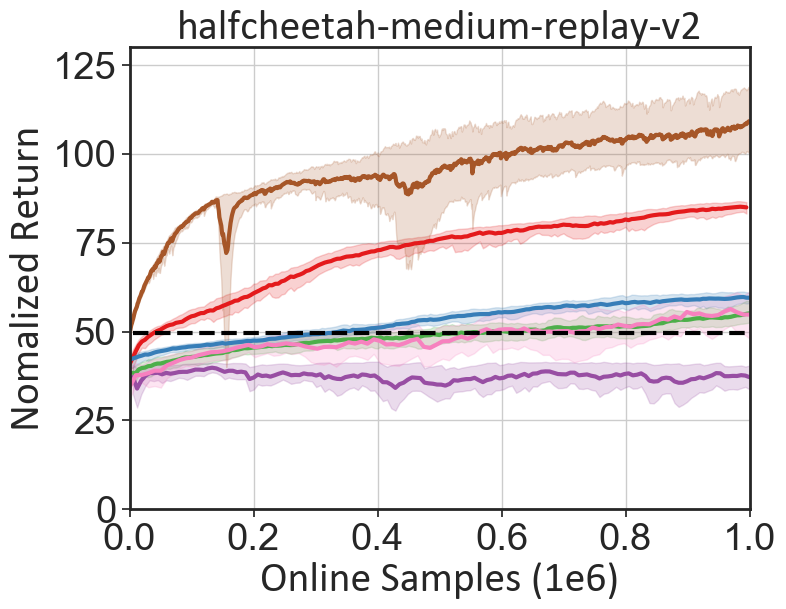}
    \includegraphics[width=0.24\textwidth]{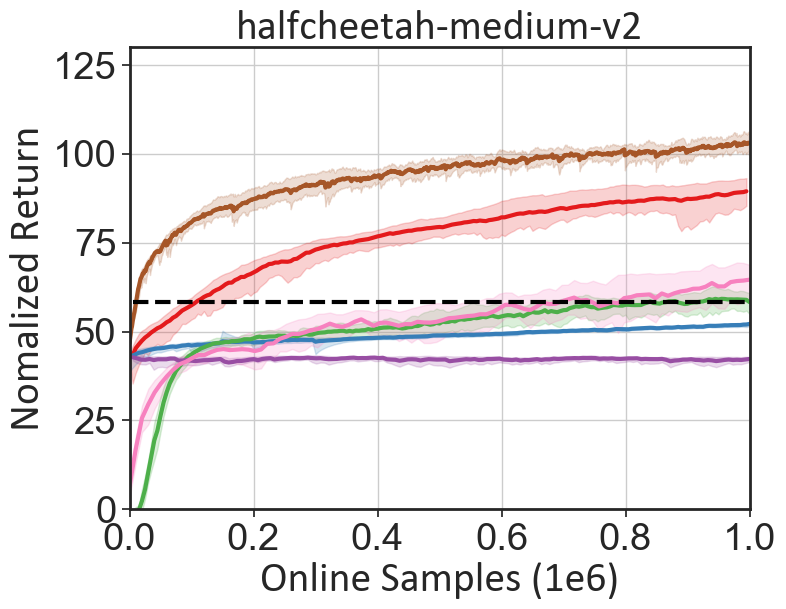}
    \includegraphics[width=0.24\textwidth]{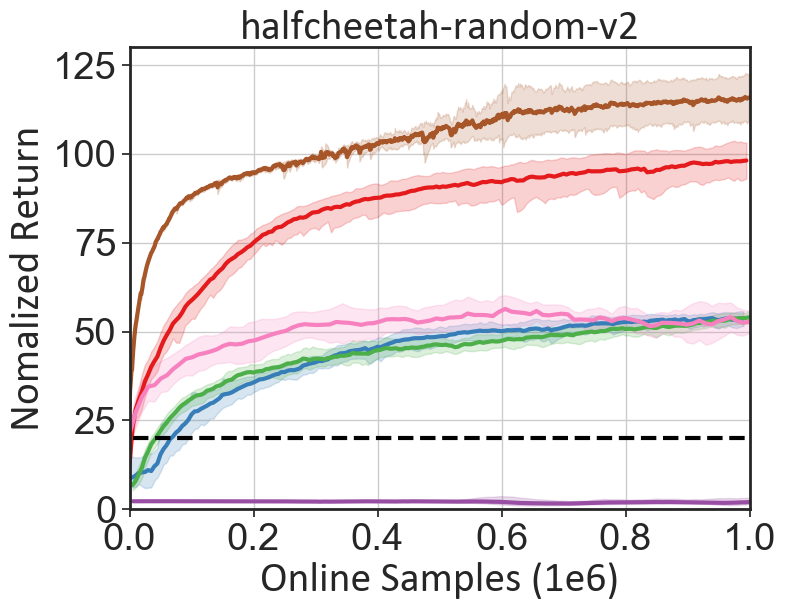}
    \includegraphics[width=0.24\textwidth]{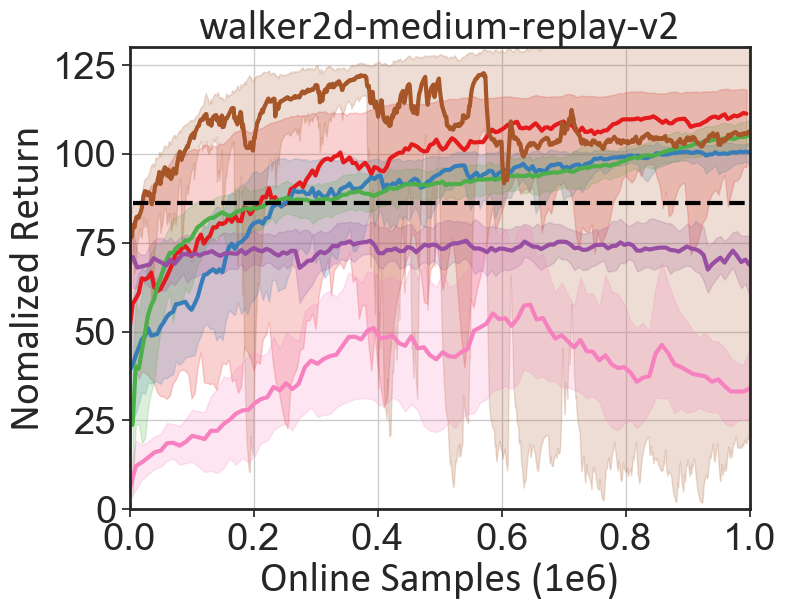}
    \includegraphics[width=0.24\textwidth]{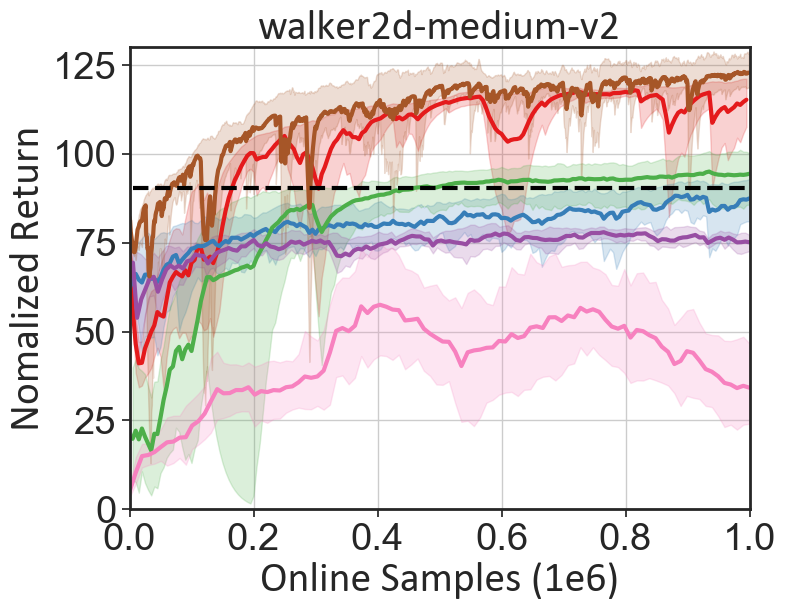}
    \includegraphics[width=0.24\textwidth]{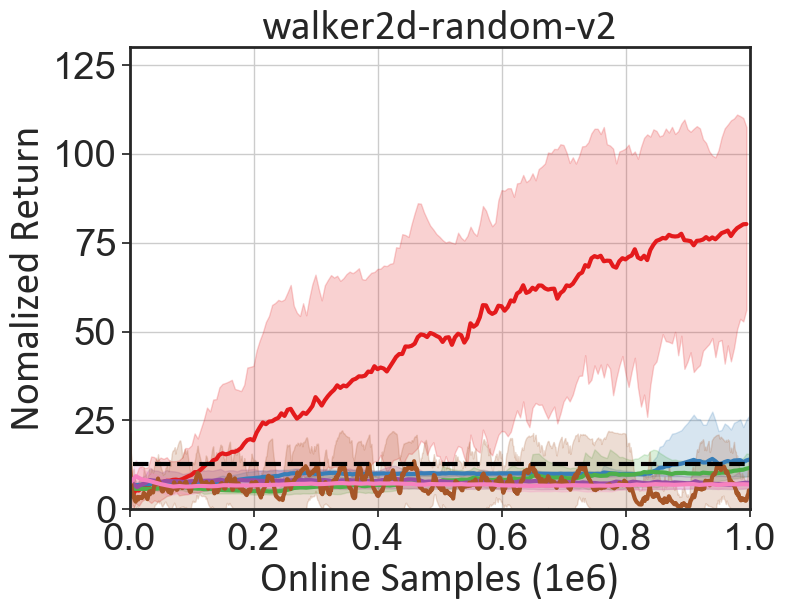}
    \caption{\small {Learning curves} of different approaches on Adroit manipulation, AntMaze navigation, and MuJoCo locomotion tasks from D4RL~\cite{fu2020d4rl} benchmark.
}
\vspace{-10pt}
    \label{fig:learning_curves_all}
\end{figure}

\vspace{-1pt}
Learning curves of \textit{PROTO} are illustrated in Figure~\ref{fig:learning_curves_all_aggregation} and \ref{fig:learning_curves_all}. Returns are normalized, where 0 and 100 represent random and expert policy performances, respectively. The error bars indicate min and max over 5 different random seeds unless specified. The results of most baselines are reproduced by the authors' open-source implementations, while the results of AWAC are reproduced by d3rlpy~\cite{seno2022d3rlpy}.

\begin{figure}[t]
    \centering
\vspace{10pt}
    \includegraphics[width=0.7\textwidth]{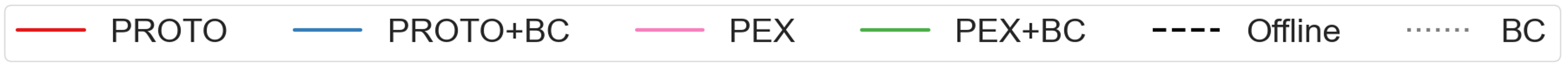}
    
    \includegraphics[width=0.32\textwidth]{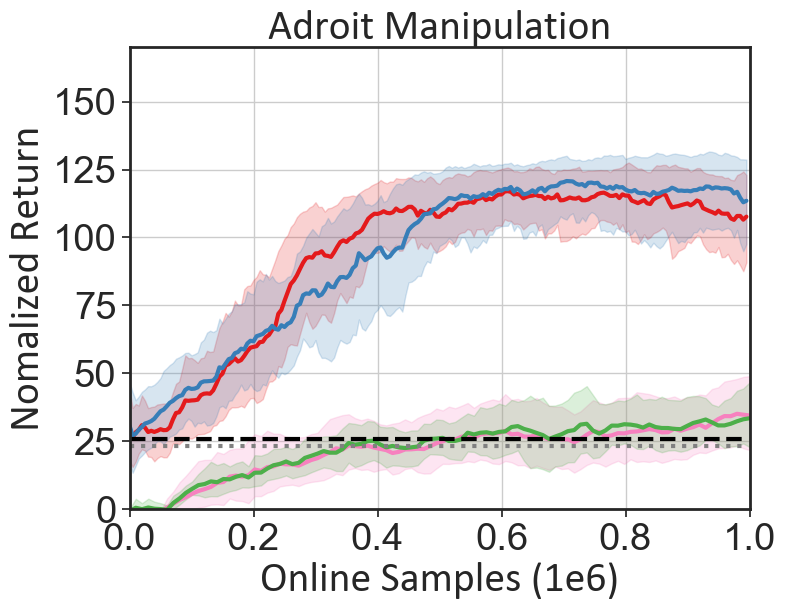}
    \includegraphics[width=0.32\textwidth]{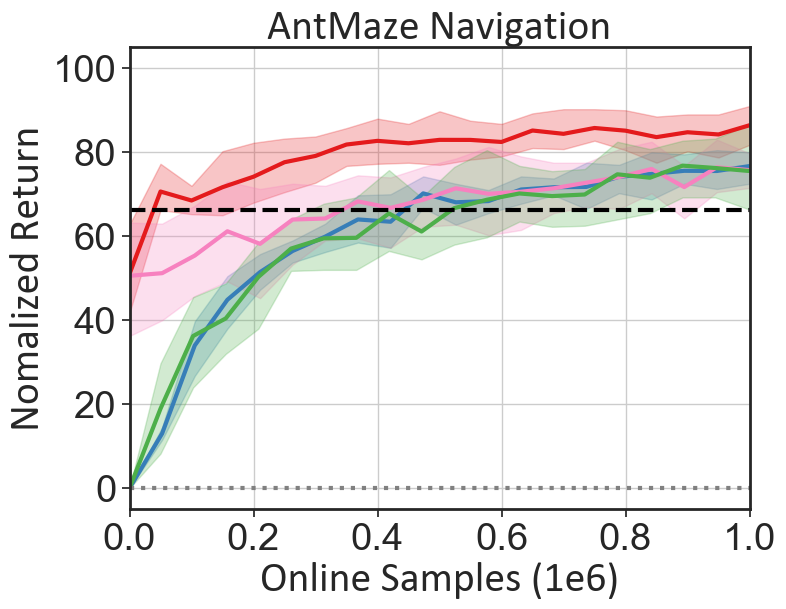}
    \includegraphics[width=0.32\textwidth]{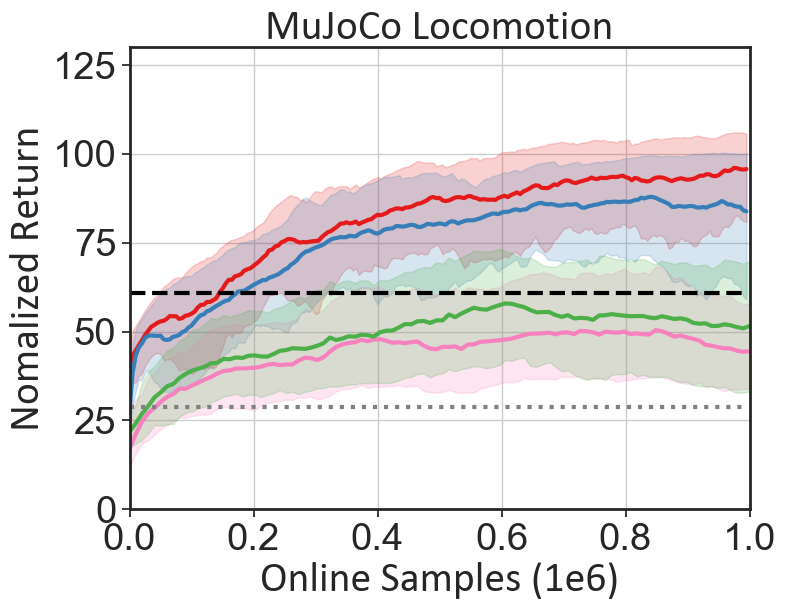}
    \caption{\small {Learning curves} of different approaches with BC pretraining. See Figure~\ref{fig:appendix_bc_pretrained} for full results.
}
\vspace{-10pt}
    \label{fig:bc_pretraining}
\end{figure}

Figure~\ref{fig:learning_curves_all_aggregation} and \ref{fig:learning_curves_all} show that existing policy constraint-based approaches ({\textit{IQL}}, {\textit{AWAC}} and {\textit{PEX}}) in most cases can only marginally outperform or cannot surpass SOTA offline RL approaches, due to the over-conservatism introduced by the policy constraint that largely hinges the finetuning process. This is especially pronounced when offline dataset or pretrained policy is highly-suboptimal such as Adroit manipulation, Antmaze navigation, and MuJoCo locomotion random tasks. In contrast, {\textit{PROTO}} enjoys both a stable initial finetuning stage and superior final performance owing to the optimistic nature of the proposed iterative policy regularization. Note that {\textit{Off2On}} also obtains great performance for most MuJoCo locomotion tasks, since it imposes no conservatism during policy finetuning and the tasks are relatively simple. {\textit{Off2On}}, however, is limited to CQL pretraining, in which case it is hard to yield reasonable performance when the tasks are too difficult for CQL to obtain stable pretrained policies and value functions (\textit{e.g.}, Adroit and Antmaze-large tasks).

% requires an ensemble of pessimistic action-value function pretrained only by CQL, which imposes tremendous computational cost and is hard to pretrain when the tasks are extremely difficult for CQL to solve. Observe from Figure~\ref{fig:learning_curves_all} and Figure~\ref{fig:learning_curves_all_aggregation} that \textbf{\texttt{Off2On}} the extremely hard , pretraining a well-behaved CQL is.

\vspace{-1pt}
\subsection{Evaluation on Adaptability}
\begin{figure}[t]
    \centering
    \includegraphics[width=0.9\textwidth]{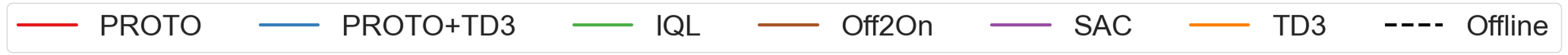}
    
    \includegraphics[width=0.32\textwidth]{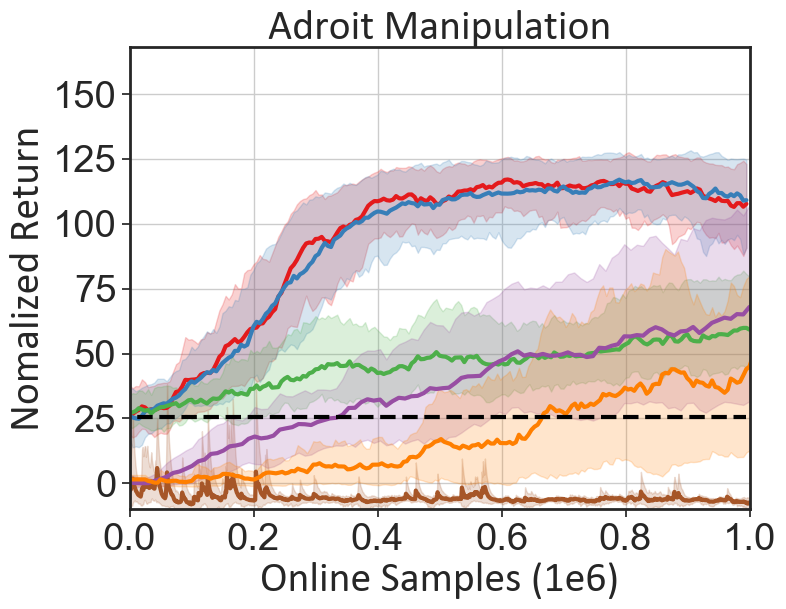}
    \includegraphics[width=0.32\textwidth]{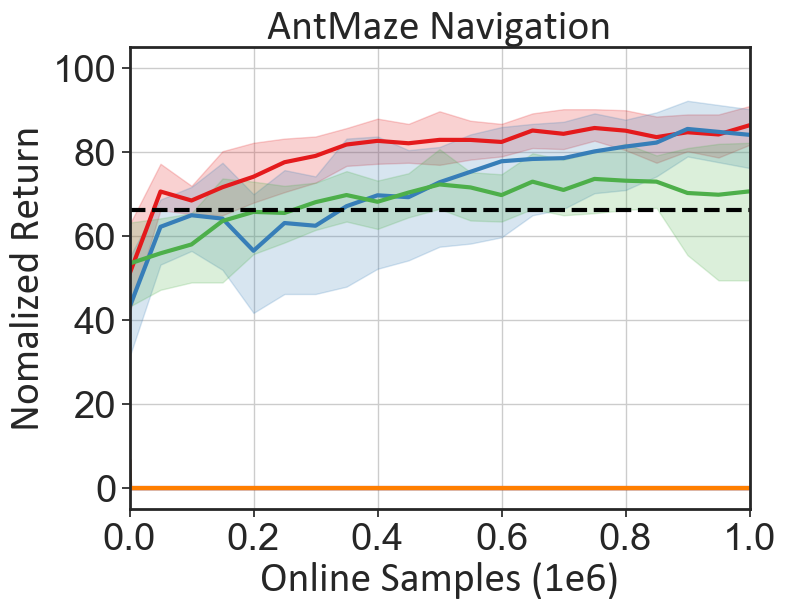}
    \includegraphics[width=0.32\textwidth]{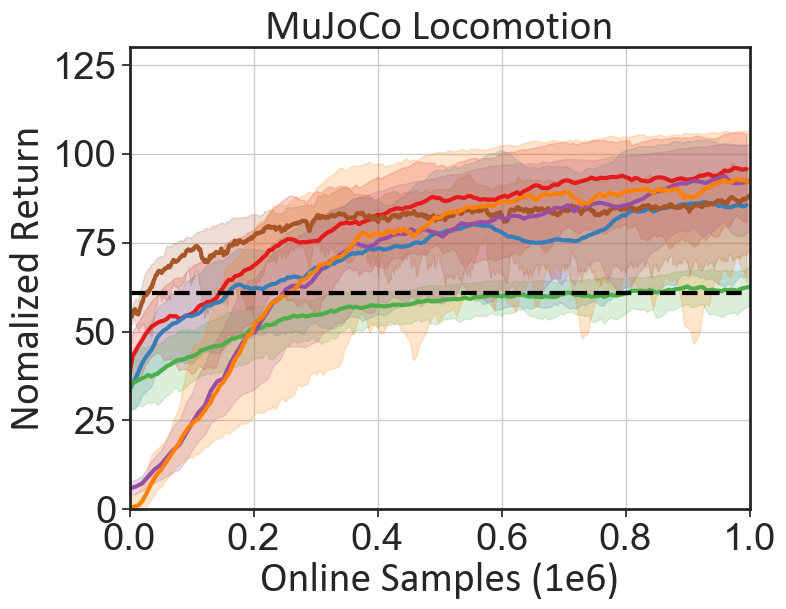}
    \caption{\small Learning curves of online finetuning for \textit{PROTO+TD3}. Refer to Figure~\ref{fig:appendix_td3_finetuning} for full results.}
    \label{fig:td3_online_main}
    \vspace{-5pt}
\end{figure}

\vspace{-1pt}
% The majority of previous works are limited to several specific offline RL pretraining approaches. A desired offline-to-online approach, however, should be flexible enough to fit in as much pretraining approaches as possible.
We construct experiments to demonstrate the versatility of \textit{PROTO}. To evaluate its adaptibility for pretraining, we finetune {\textit{PROTO}} with BC~\cite{pomerleau1988alvinn}, the simplest offline IL approach. {\textit{PEX}} is the only related work that explicitly considers adaptability in offline-to-online RL. Therefore, we consider {\textit{PEX}} with BC pretraining as the main baseline. Learning curves are illustrated in Figure~\ref{fig:bc_pretraining}, where {\textit{BC}} denotes the performance of pretrained BC policy. To evaluate its adaptibility for online RL, we plug \textit{PROTO} into {\textit{TD3}}~\cite{fujimoto2018addressing} and summarize the results in Figure~\ref{fig:td3_online_main}.

As showed in Figure~\ref{fig:bc_pretraining}, {\textit{PROTO+BC}} surpasses {\textit{PEX+BC}} by a large margin. It also shows although BC pretraining may not perform as well as offline RL pretraining, {\textit{PROTO+BC}} and {\textit{PEX+BC}} still boost the finetuning performances and obtain similar results to {\textit{PROTO}} and {\textit{PEX}} pretrained by offline RL, respectively. It is known that offline RL methods are generally hyperparameter-sensitive and suffer from unstable training process caused by exploitation error accumulation, while BC is much more stable~\cite{kumar2021should}. In our experiment, pretraining using the simplest BC and then finetuning by {\textit{PROTO}} can already obtain great performance, bypassing complex offline RL training. 
%\jj{Why is it surprising? What's the relation with the difficulty of offline RL pretraining?}\ljx{BC typically performs poor than the \textbf{well-tuned} SOTA offline RL methods, but enjoys super stable and fast training process since it is trained on a simple supervised loss. This loss have no error bootstrapping issue and requires minimal hyper-parameter tuning. However, to obtain reasonable performances, offline RL typically requires tedious parameter tuning. Without a proper hyperparameter, offline RL may not even outperform BC. During the offline pretraining, however, we cannot evaluate policy and also cannot tune parameter since no online evaluation is allowed. Therefore, for most of the real applications, offline RL may not outperform BC. So, this is a surprising finding that we can actually pretrain with the most stable BC and then finetune it with online RL, which can obtain similar performances as pretraining by SOTA offline RL methods.} \jj{I understand the logic between BC and RL. What I was asking is the logic from 'offline RL is difficult to train' to 'BC performs better', which seems natural and not surprising.}\ljx{Thanks! I see. I want to convey to others that perhaps we should not focus on the complex offline RL pretraining, using the simplest BC to pretrain already attains good performance}
In addition, Figure~\ref{fig:td3_online_main} shows that \textit{PROTO+TD3} also obtains SOTA performances compared to baselines. 

Altogether, Figure~\ref{fig:bc_pretraining} and \ref{fig:td3_online_main} demonstrate that we can construct competitive offline-to-online RL algorithms by simply combining diverse offline pretraining and online finetuning RL approaches via \textit{PROTO}, which offers a flexible and adaptable framework for future practitioners. % with applications that have special requirements on specific pretraining and finetuning methods.
% other researchers to construct their own offline-to-online RL algorithms correspondingly.

% In summary, the advantages of \textbf{\texttt{PROTO}} compared with other baselines are listed in Table~\ref{tab:related_work_main}.
\vspace{-1pt}
\subsection{Ablation Study}
\label{sec:ablation}
\vspace{-1pt}

\begin{figure}[h]
% \begin{minipage}{0.75\linewidth}
    \centering
    \includegraphics[width=0.5\textwidth]{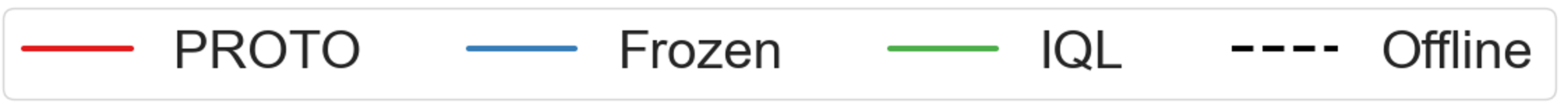}

    \begin{minipage}[t]{0.99\textwidth}
    \centering
    \includegraphics[width=0.32\textwidth]{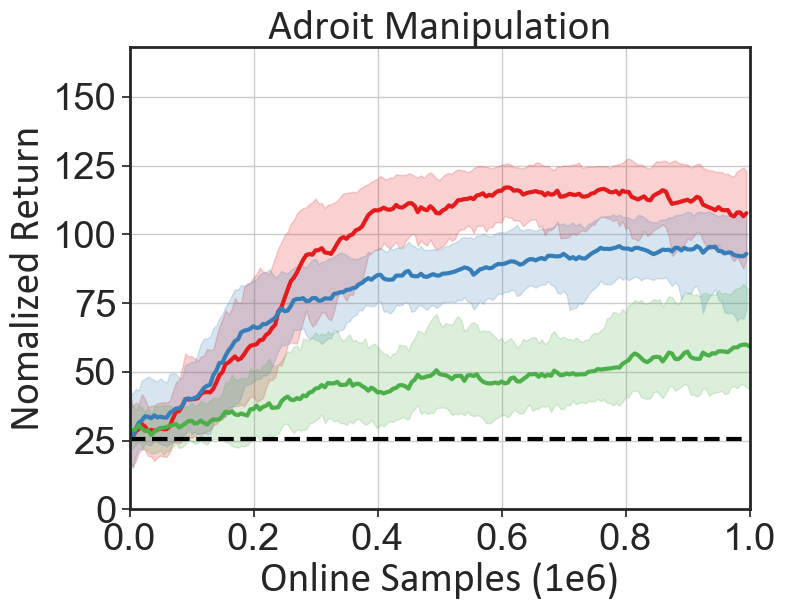}
    \includegraphics[width=0.32\textwidth]{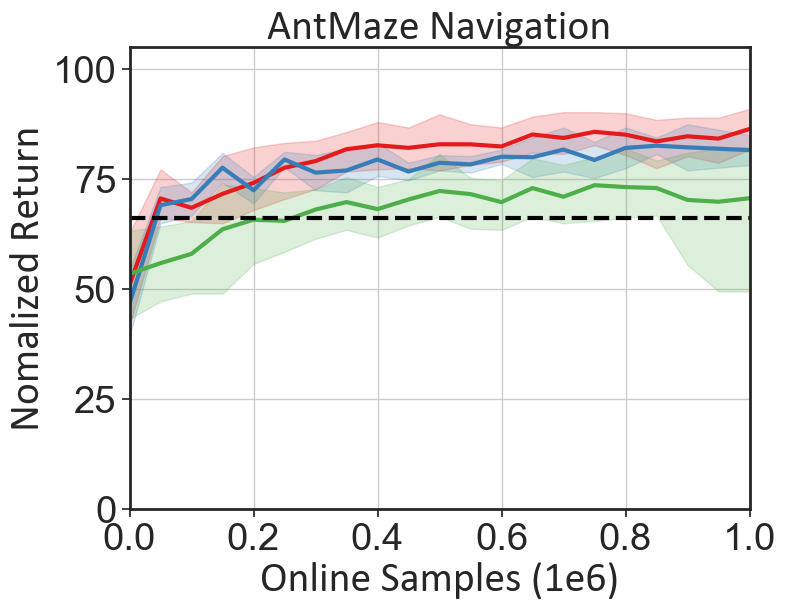}
    \includegraphics[width=0.32\textwidth]{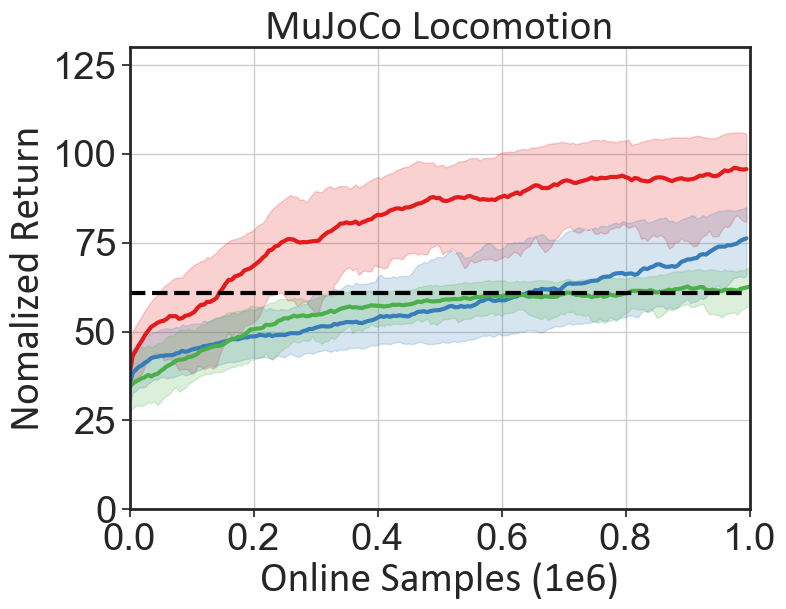}
% \end{minipage}
% \begin{minipage}{0.24\linewidth}
%     \centering
%     \small
%     \begin{tabular}{ll}
%     \toprule
%          & AntMaze Complete Speed \\
%     \midrule
%          &  \\
%     \bottomrule
%     \end{tabular}
% \end{minipage}
    \caption{\small Comparison between iterative policy regularization ({\textit{PROTO}}) and fixed policy regularization ({\textit{Frozen}}) and other baselines. Refer to Figure~\ref{fig:appendix_iterative_vs_fixed} for full results.}
    \label{fig:frozon}
    \end{minipage}
    % \hspace{+4pt}
    % \begin{minipage}[t]{0.24\textwidth}
    % \includegraphics[width=0.99\textwidth]{Figure/Main/lc_frozen/aggrated-return_antmaze_length.png}
    % \caption{Completion speeds for Antmaze tasks.}\label{fig:speed}
    % \end{minipage}
    \vspace{-5pt}
\end{figure}

\textbf{Iterative Policy Regularization vs. Fixed Policy Regularization}. \ \ To further demonstrate the advantages of \textit{Iterative Policy Regularization}, we replace the iterative policy ${\pi}_k$ in Eq.~(\ref{equ:PROTO_objective}) with the fixed pretrained policy ${\pi}_0$ while retains all the other experimental setups such as conservatism annealing and denote this simplest variant as \textit{Frozen}. Similar to previous policy constraint approaches, {\textit{Frozen}} aims to solve a \textit{pretrained policy constrained RL} problem. 
% \textit{i.e.}, $\max_\pi \mathbb{E}\left[\sum_{t=0}^\infty\gamma^t\left(r(s_t, a_t)-\alpha\cdot\log\left(\frac{\pi(a_t|s_t)}{\pi_0(a_t|s_t)}\right)\right)\right]$. 
Figure~\ref{fig:frozon} illustrates the aggregated learning curves of {\textit{Frozen}} and {\textit{PROTO}}. We also compare with {\textit{IQL}} for its strong performances among other baselines in Figure~\ref{fig:learning_curves_all_aggregation}.

\begin{figure}[t]
    \centering
\begin{minipage}[t]{0.315\textwidth}
\includegraphics[width=0.99\textwidth]{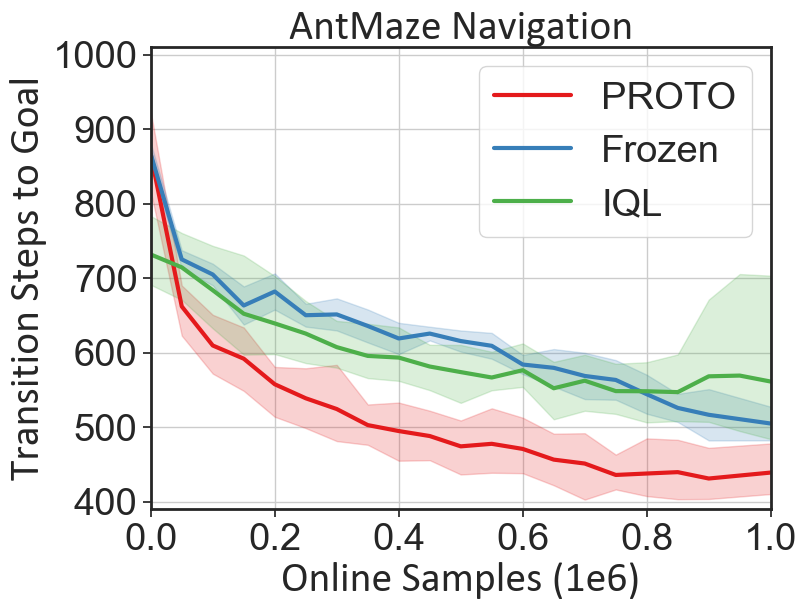}
\caption{\small Completion speeds for AntMaze Navigation tasks. Refer to Figure~\ref{fig:appendix_iterative_vs_fixed} for full results.}\label{fig:speed}
\label{fig:my_label}
\end{minipage}
\hfill
\begin{minipage}[t]{0.315\textwidth}
\includegraphics[width=0.99\textwidth]{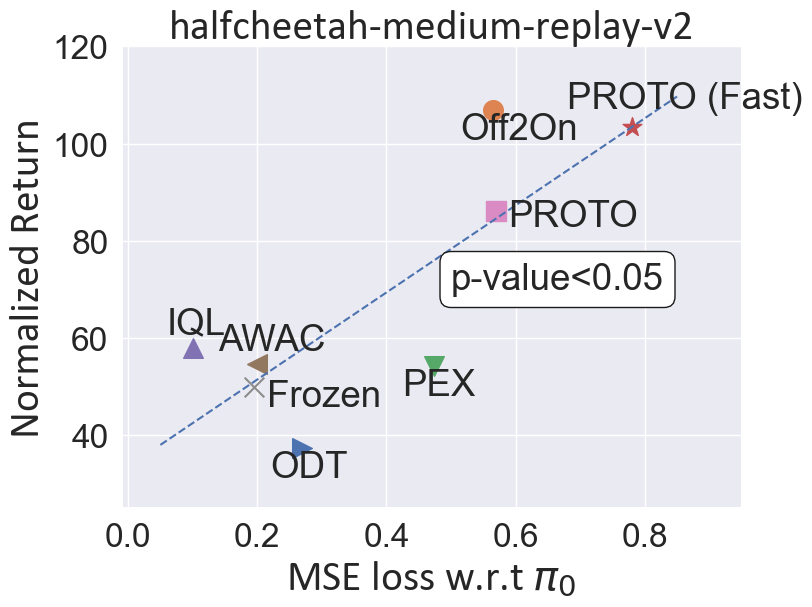}
\caption{\small Positive correlation between policy deviation \textit{w.r.t.} $\pi_0$ and policy performance.}\label{fig:perform_vs_strength}
\label{fig:my_label}
\end{minipage}
\hfill
\begin{minipage}[t]{0.315\textwidth}
% \vspace{-2pt}
\includegraphics[width=0.99\textwidth]{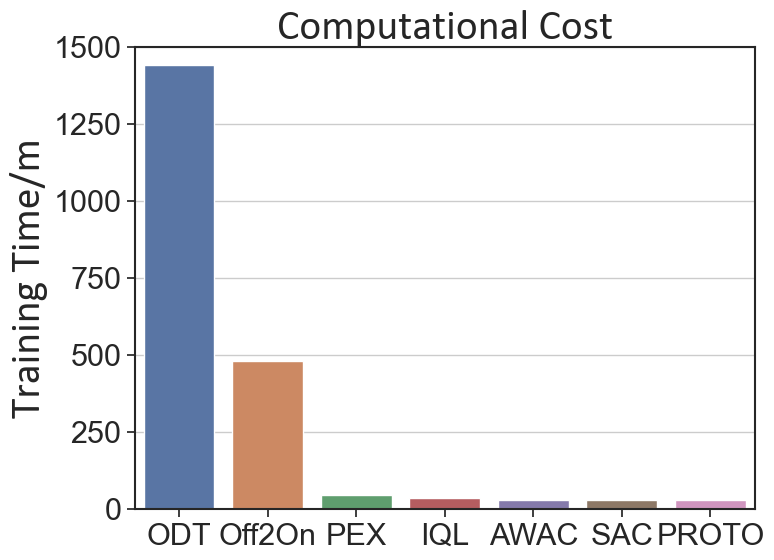}
\caption{\small Computational cost when performing 1M online samples and gradient steps.}\label{fig:computational_cost}
\label{fig:my_label}
\end{minipage}
\vspace{-7pt}
\end{figure}

Observe in Figure~\ref{fig:frozon} that {\textit{PROTO}} obtains superior performances compared with {\textit{Frozen}}, which demonstrates the advantage of \textit{iterative} over \textit{fixed} policy regularization. The iterative nature
% \begin{wrapfigure}{r}{4.5cm}
% % \vspace{-10pt}
%     \includegraphics[width=0.32\textwidth]{Figure/Main/lc_frozen/aggrated-return_antmaze_length.png}
%     \caption{\small Completion speeds for Antmaze tasks.}\label{fig:speed}
%     \vspace{-14pt}
% \end{wrapfigure}
of {\textit{PROTO}} facilitates a reasonable deviation from the pretrained policy or offline data distribution to guarantee optimal performance.
% supports the advantages of \textit{Iterative Policy Regularization} over \textit{Fixed Policy Regularization} and the necessity of large deviation from the pretrained policy or data distribution to find optima. 
For AntMaze navigation tasks, although {\textit{PROTO}} obtains similar success rates to other baselines, 
{\textit{PROTO}} completes the navigation tasks with much fewer transition steps and higher speed (see Figure~\ref{fig:speed}), translating to much better learned policies.

Also, note that the simplest variant {\textit{Frozen}} already surpasses or achieves on-par performances as {\textit{IQL}}. We believe that this is because the in-sample learning paradigm of {\textit{IQL}} learns from in-sample data only, which lacks supervision on OOD regions and hence hinges exploration. Additionally, we employ a linear decay schedule to wean off conservatism, while the conservatism in {\textit{IQL}} cannot be fully turned off since {\textit{IQL}} recovers the maximum of action-value function only when the inverse temperature in its policy extraction step goes to infinity~\cite{iql}.

% \begin{wrapfigure}{r}{4.5cm}
% \centering
% \vspace{-10pt}
%     \includegraphics[width=0.32\textwidth]{Figure/Main/score_mse.png}
%     \caption{\small Positive correlation between policy deviation \textit{w.r.t.} $\pi_0$ and policy performance. }
%     % \jj{Font in figure is too small}}\ljx{Thanks, I will update it soon.}
%     \label{fig:perform_vs_strength}
%     \vspace{-15pt}
% \end{wrapfigure}
\textbf{Finetuning Performance vs. Constraint Strength}. We investigate how constraint strength affects finetuning performance of different approaches. Figure~\ref{fig:perform_vs_strength} demonstrates that the final performance and constraint strength exhibit a negative correlation, where {\textit{PROTO}} attains relaxed constraints and near-optimal performances. Furthermore, we can obtain a better policy %\jj{What does it mean 'performant'?}\ljx{I want to represent "better performance", perhaps the word needs to be revised.} 
by adjusting the polyak averaging speed and conservatism annealing speed, to accelerate the policy deviation speed (reported  as {\textit{PROTO (Fast)}}), which further demonstrates the necessity of relaxing the conservatism when finetuning policies.

We also conduct ablation experiments to analyze the effect of polyak averaging update speed and the conservatism annealing speeds, and find \textit{PROTO} robust to parameter tuning (due to space limits, please refer to Appendix~\ref{sec:ablation_appendix} for detailed results).

\vspace{-1pt}
\subsection{Computational Cost}
% \jj{Move after 6.5?}
\vspace{-1pt}
% \begin{table}[h]
% \small
%     \caption{Computational cost to perform 1M online samples and updates of different approaches}
%     \centering
%     \begin{tabular}{cllllll}
%     \toprule
%         Methods & \textbf{\texttt{Off2On}} & \textbf{\texttt{ODT}} &\textbf{\texttt{AWAC}} & \textbf{\texttt{IQL}} &\textbf{\texttt{PEX}} &\textbf{\texttt{PROTO}}\\
%     \midrule
%         % Time &  \\
%         \multirow{2}{*}{Introduced Cost} & Ensemble Value Function & \multirow{2}{*}{Transformer} & \multirow{2}{*}{-} & \multirow{2}{*}{-} & \multirow{2}{*}{-}\\
%         ~ & Balanced Experience Replay \\
%         \midrule
%         Time & 8h & 24h &30m & 30m &45m & 30m\\
%     \bottomrule
%     \end{tabular}

%     \label{tab:computational_cost}
% \end{table}
% \begin{wrapfigure}{r}{4.5cm}
% \centering
%     \vspace{-12pt}
%     \includegraphics[width=0.32\textwidth]{Figure/Main/computational_cost/computational_cost.png}
%     \caption{\small {Computational cost when  performing 1M online samples and gradient steps.}}
%     \vspace{-10pt}
%     \label{fig:computational_cost}
% \end{wrapfigure}
% Apart from stable and optimal performances and generalizable ability, a desired offline-to-online approach should also be efficient and do not impose too much additional computational costs.
In Figure~\ref{fig:computational_cost}, we report the computation time of performing 1M online samples and gradient steps, to compare the computational efficiency of different methods. It is not surprising that {\textit{ODT}} requires the most computational resources since it is a transformer-based approach, while other methods build on simple MLPs. {\textit{Off2On}} requires an ensemble of pessimistic Q-functions and a complex balanced experience replay scheme, which imposes high computational cost. In addition, the CQL pretraining in {\textit{Off2On}} explicitly requires performing computationally-expensive numerical integration to approximate the intractable normalization term in continuous action spaces~\cite{kumar2020conservative, kostrikov2021offline}.
% {\textit{IQL}} and {\textit{AWAC}} build on top of implicit policy regularization, hence enjoy similar efficiency to standard off-policy RL methods. {\textit{PEX}} expands the policy set by introducing a new online policy which introduces some extra computation but within an acceptable range. 
By contrast, {\textit{PROTO}} only requires calculating the additional regularization term, computational overhead of which is negligible, therefore enjoys the same computational efficiency as standard off-policy RL methods.
% (SAC in our experiments).

\vspace{-3pt}
\section{Conclusion and Future Work}
\vspace{-3pt}
To address major drawbacks of existing offline-to-online RL methods (suboptimal performance, limited adaptability, low computational efficiency), we propose \textit{PROTO} that incorporates an iteratively evolved regularization term to stabilize the initial finetuning and bring enough flexibility to yield strong policies. \textit{PROTO} seamlessly bridges diverse offline RL/IL and online off-policy RL methods with a non-intrusively modification, offering a flexible and efficient offline-to-online RL proto-framework. %Experimental evaluations demonstrate that \textit{PROTO} achieves new SOTA performances on multiple RL benchmarks. 
% One limitation is that \textit{PROTO} has two hyper-parameters, which can largely be addressed by a non-parametric treatment to set the annealing speed as constants, as discussed in Appendix~\ref{subdec:hyperparameters}.
Following existing works, this paper only focuses on off-policy RL finetuning, which has high-sample efficiency but may not enjoy monotonic policy improvement guarantees. One appealing future direction is to introduce \textit{PROTO} into on-policy RL finetuning or marry off-policy sample efficiency with on-policy monotonic improvement to devise more advanced offline-to-online RL. 

% Offline-to-Online reinforcement learning (RL) offers promising potential to combine the benefits of offline pretraining and online finetuning, yielding improved sample efficiency and optimal performances. However, significant challenges remain in terms of suboptimal performances, limited generalizabiliey to diverse methods and poor computational efficiency. This paper proposes a novel framework, \textit{Iterative Policy Regularization}, and dub it as \textbf{\texttt{PROTO}}, which overcomes all the limitations of existing works in a concise manner. \textbf{\texttt{PROTO}} regularizes the finetuning policy \textit{w.r.t} the policy at last iteration instead of a fixed constraint set, resulting in both stable initial finetuning and optimal performances at the later finetuning stage. \textbf{\texttt{PROTO}} imposes minimal assumption on the pretraining methods, enhancing extension-ability to diverse pretraining methods. Moreover, by adjusting just a few lines of code, \textbf{\texttt{PROTO}} can be integrated in standard off-policy RL finetuning approaches, showing the great generalizability to diverse methods. Simple yet effective, extensive experiments demonstrate the superior performances of \textbf{\texttt{PROTO}} compared to other baselines, represents its stable, optimal and efficient finetuning performances, which offers a simple yet competitive baseline for future works.

% \newpage

\bibliographystyle{named}
\bibliography{main}

\newpage
\appendix

\section{Detailed Discussions on Related Works}
\label{sec:comparisons_table}

This section provides detailed comparisons with existing offline-to-online RL methods in Table~\ref{tab:related_work}.

\begin{table}[h]
    \centering
    \addtolength{\leftskip} {-2.5cm}
    \caption{Detailed comparisons with related offline-to-online RL methods. $\mu$: behavior policy that generats the offline dataset~$\mathcal{D}$.  $\mathcal{B}$: replay buffer. $\pi_0$:  pretrained policy.}
    \footnotesize
    \begin{tabular}{lccccccccccc}
    \toprule
       Type & \multicolumn{3}{c}{PC} & \multicolumn{1}{c}{PVI} & \multicolumn{1}{c}{GCSL} & Others &~\\
       \cmidrule(r){1-1}\cmidrule(r){2-4}\cmidrule(r){5-5}\cmidrule(r){6-6}\cmidrule(r){7-7}
       Method& {{SPOT}}~\cite{wusupported} & \thead{{{AWAC}}~\cite{nair2020awac}, {{IQL}}~\cite{iql}, \\ {{XQL}}~\cite{garg2023extreme}, {{InAC}}~\cite{xiao2023the}} &{{PEX}}~\cite{zhang2023policy} & \thead{{{Off2On}}~\cite{lee2022offline}, {{Cal-QL}}~\cite{nakamoto2023cal} \\ CCVL~\cite{hong2023confidenceconditioned}} &{{ODT}}~\cite{zheng2022online} &{{APL}}~\cite{zheng2023adaptive} &{{PROTO}} (Ours)\\
    \midrule
    a. &$\mu$ & $\mathcal{B}$ &$\pi_0$ & No Constraint &$\mathcal{B}$ &$\bigcirc$ &$\pi_k$\\
    b. &\ding{55} &\ding{55} &\ding{55} &\Checkmark &\ding{55} &\ding{55} &\Checkmark\\
    c. &\Checkmark &\Checkmark &\Checkmark &\ding{55} &\ding{55} &$\bigcirc$ &\Checkmark \\
    d. &\Checkmark &\ding{55} &\Checkmark &\Checkmark &\ding{55} &\ding{55} &\Checkmark\\
    e. &\Checkmark &\Checkmark &\Checkmark &\ding{55} &\ding{55} &$\bigcirc$ &\Checkmark\\
    \bottomrule
    \multicolumn{7}{c}{\quad \quad\thead{a. Constraint policy set; b. Stable and optimal policy learning; c. Adaptable to diverse pretraining methods;  d. Adaptable \\ to diverse finetuning methods; 
    e. Computational efficient. \quad \Checkmark: Yes, \ding{55}: No, $\bigcirc$: It depends. \quad \quad }}
    \end{tabular}
    \label{tab:related_work}
\end{table}

All existing policy constraint and goal conditioned supervised learning methods constrain the finetuning policy \textit{w.r.t.} a fixed policy set induced by $\mu$, $\mathcal{B}$ or $\pi_0$, which may be highly-suboptimal and induce large optimality gap. {{APL}} finetunes the policy with the existence of pretraining conservatism, which may also lead to suboptimal performances. By contrast, {{PROTO}} casts constraint on an iteratively evolving constraint set induced by the policy $\pi_k$ at the last iteration, which relaxes the conservatism in the later stage and thus enjoys similar optimality as no constraints. 

In terms of adaptability to diverse finetuning and pretraining methods, {{SPOT}} provides a pluggable policy regularization as {{PROTO}} dose, thus can flexibly extend to diverse methods, but requires the hard estimation of the unknown behavior policy. {{AWAC, IQL, XQL, InAC}} use AWR to extract policy without the need for explicitly behavior policy estimation. AWR, however, is difficult to plug in other online RL approaches non-intrusively, limiting its adaptability for diverse online finetuning methods. {{PEX}} offers the most flexible framework among all the baselines and is adaptable to diverse methods. {{Off2On}}, {{Cal-QL}} and CCVL can only apply to {{CQL}}-style~\cite{kumar2020conservative} pretraining, but can finetune with diverse online RL approaches. {{ODT}} is specifically designed for {{DT}}~\cite{chen2021decision} finetuning and hence suffers from the most limited applicability. {APL} also offers a general framework for offline-to-online RL, but can only apply to offline RL methods that have explicit regularization terms such as {{TD3+BC}}~\cite{fujimoto2021minimalist} and {{CQL}}~\cite{kumar2020conservative}.

In terms of computational efficiency, {{Off2On}} requires an ensemble of value functions and hence is computation inefficient. Moreover, pretraining with CQL explicitly requires performing computationally-expensive numerical integration to approximate the intractable normalization term in continuous action spaces~\cite{kumar2020conservative, kostrikov2021offline}, which inevitably introducing tremendous computational costs. {{ODT}} builds on Transformer~\cite{vaswani2017attention} architecture, which requires far more computational resources compared to other methods that builds on simple MLPs.

There is also a large amount of work focusing on accelerating online RL training using offline dataset~\cite{ball2023efficient}, representation learning~\cite{laskin2020curl}, or a guiding policy~\cite{uchendu2022jump}, which lie in the orthogonal scope of our paper and thus are not discussed here. Furthermore, similar to all these related works, this paper contains no potential negative societal impact.

\newpage
\section{Theoretical Interpretations}
% \subsection{Proof of Theorem~\ref{theorem:optimal_gap_previous}}
\label{subsec:theorem_1_proof}
This section provides the proof of Theorem~\ref{theorem:gap} and detailed discussions on the inherent stability and optimality of \textit{PROTO}. Note that we do not seek to devise tighter and more complex bounds but to give insightful interpretations for our proposed approach \textit{PROTO}.

\subsection{Proof of Theorem~\ref{theorem:gap}}

\begin{proof}
First, we introduce Lemma~\ref{lemma:KL_gap} (Corollary of Theorem 1 in~\cite{vieillard2020leverage}), which builds the foundation of our theoretical interpretation.
% \ljx{proof sketch}
\begin{lemma}
Define $Q^*$ is the action-value of optimal policy $\pi^*$, $Q^{k}$ is the action-value of policy $\pi_k$ obtained at $k$-th iteration by iterating Eq.~(\ref{equ:Q_operator_trust})-(\ref{equ:policy_trust}). $v_{\rm max}^{\alpha}:=\frac{r_{\rm max}+\alpha{\rm ln}\mathcal{|A|}}{1-\gamma}$. $\epsilon_j$ is the approximation error of the action-value function. Assume that $\pi_0$ is a uniform policy, $Q^0$ is initialized such that $\|Q^0\|_\infty\le v_{\rm max}$ and $\|Q^k\|_\infty\le v_{\rm max}$. We have~\cite{vieillard2020leverage}:
\begin{equation}
\label{equ:suboptimal_gap_trust_appendix}
    \|Q^*-Q^{k}\|_\infty\le{\frac{2}{1-\gamma}\left\|\frac{1}{k}\sum_{j=1}^k\epsilon_{j}\right\|_{\infty}}+{\frac{4}{1-\gamma}\frac{v_{\max}^{\alpha}}{k}}, k\in N^+.
\end{equation}
\label{lemma:KL_gap}
\end{lemma}

At first glance, Lemma~\ref{lemma:KL_gap} is quite similar to Theorem~\ref{theorem:gap}.
However, note that the assumption of Lemma~\ref{lemma:KL_gap} is slightly different from Theorem~\ref{theorem:gap}. With a slight abuse in notation, in Lemma~\ref{lemma:KL_gap}, $\pi_0$ is assumed as a uniform policy and $Q^0$ can be any initialized action-value function that satisfies $\|Q^0\|_\infty\le v_{\rm max}$, while in Theorem~\ref{theorem:gap}, $\pi_0$ is the pretrained policy and $Q^0$ is the corresponding action-value. Nevertheless, we will show that this lemma can seamlessly extend to Theorem~\ref{theorem:gap} by introducing an additional negligible requirement on $Q^0$ initialization. 

Since the original assumption in Lemma~\ref{lemma:KL_gap} allows any form of $Q^0$ initialization as long as it satisfies $\|Q^0\|_\infty\le v_{\rm max}$, we introduce one additional constraint on $Q^0$ initialization that the pretrained policy should be obtained via one policy improvement step upon $Q^0$, \textit{i.e.}, $\arg \max_{\pi}\mathbb{E}_{a\sim\pi(\cdot|s)}[Q^0(s,a)-\alpha\cdot\log\left(\frac{\pi(s,a)}{\pi_0(s,a)}\right)$. This assumption is negligible since it only introduce an additional condition under the premise of satisfying the original assumption.

% Assume the policy is parameterized as a Boltzmann distribution~\cite{haarnoja2018soft}. Then, we assume 

% Assume the policy is parameterized as a Boltzmann distribution~\cite{haarnoja2018soft} and $\pi_1(a|s)=\pi_0(a|s)\frac{Q^0(s,a)}{\mathbb{E}_{a\in\mathcal{A}}[Q^0(s,a)]}$.

% \begin{assumption}
% $Q^0$ is initialized such that the pretrained policy can be obtained via $\arg \max_{\pi}\mathbb{E}_{a\sim\pi(\cdot|s)}[Q^0(s,a)-\alpha\cdot\log\left(\frac{\pi(s,a)}{\pi_0(s,a)}\right)]$ and $\|Q^0\|_\infty\le v_{\rm max}$.
% \end{assumption}

% This assumption is negligible since the original assumption in Lemma~\ref{lemma:KL_gap} covers any form of $Q^0$ initialization that satisfies $\|Q^0\|_\infty\le v_{\rm max}$. We just specifically introduce one additional constraint on $Q^0$ initialization that the pretrained policy should be obtained via one policy improvement step upon $Q^0$, which still satisfies the original assumption Lemma~\ref{lemma:KL_gap}.

Under this mild assumption, the pretrained policy and its action-value in offline-to-online setting become $\pi_1$ and $Q^1$ in Lemma~\ref{lemma:KL_gap}, respectively. Note that the pretrained policy is $\pi_0$ and its corresponding action-value function is $Q^0$ in Theorem~\ref{theorem:gap}. Therefore, the conclusion of Lemma~\ref{lemma:KL_gap} is ready to transfer to Theorem~\ref{theorem:gap} with some simple modifications to meet the requirement that $k=1$ in Lemma~\ref{lemma:KL_gap} equivalents to $k=0$ in Theorem~\ref{theorem:gap}:

\textbf{Theorem 1}. \textit{Define $Q^{k}$ as the action-value of policy $\pi_k$ obtained at $k$-th iteration by iterating Eq.~(\ref{equ:Q_operator_trust})-(\ref{equ:policy_trust}), and $Q^*$ is the optimal value of optimal policy $\pi^*$. $\pi_0$ is the pretrained policy and $Q^0$ is its corresponding action-value function. Let $v_{\rm max}^{\alpha}:=\frac{r_{\rm max}+\alpha{\rm ln}\mathcal{|A|}}{1-\gamma}$, and $\epsilon_j$ is the approximation error of the action-value function. Assume that $\|Q^k\|_\infty\le v_{\rm max}^0$, we have:}
\begin{equation}
\label{equ:suboptimal_gap_trust}
    \|Q^*-Q^{k}\|_\infty\le{\frac{2}{1-\gamma}\left\|\frac{1}{k+1}\sum_{j=0}^{k}\epsilon_{j}\right\|_{\infty}}+{\frac{4}{1-\gamma}\frac{v_{\max}^{\alpha}}{k+1}}, k\in N.
\end{equation}
\end{proof}

\subsection{Stability and Optimality Compared with Previous Methods}
\textbf{Stability}. \ \ The approximation error term in Eq.~(\ref{equ:suboptimal_gap_trust}) is the infinity norm of the average estimation errors, \textit{i.e.}, $\|\frac{1}{k+1}\sum_{j=0}^{k}\epsilon_j\|_\infty$, which can converge to 0 by the law of large numbers. This indicates that \textit{PROTO} will be less influenced by approximation error accumulations and enjoys stable finetuning processes, owing to the stabilization of iterative policy regularization. 

Then, we recall the typical approximation error propagation without any regularization in Lemma~\ref{lemma:stable_no_ipr_appendix}~(can be found at Section 4 in~\cite{vieillard2020leverage}).

\begin{lemma}
Assume $\pi_0$ is a uniform policy, $Q^0$ is initialized such that $\|Q^0\|_\infty\le v_{\rm max}$ and $\|Q^k\|_\infty\le v_{\rm max}$, where $v_{\rm max}:=\frac{r_{\rm max}}{1-\gamma}$. Then we have
\begin{equation}
    \|Q^*-Q^{k}\|_\infty\le\frac{2\gamma}{(1-\gamma)^2}\left((1-\gamma)\sum_{j=1}^k\gamma^{k-j}\|\epsilon_{j}\|_\infty\right)+\frac{2}{1-\gamma}\gamma^kv_{\rm max}, k \in N^+.
\end{equation}
\label{lemma:stable_no_ipr_appendix}
\end{lemma}

Similar to the difference between Lemma~\ref{lemma:KL_gap} and Theorem~\ref{theorem:gap}, Lemma~\ref{lemma:stable_no_ipr_appendix} cannot be directly applied in the offline-to-online RL setting (Eq.~(\ref{equ:no_regularization})). However, it can be easily transferred by imposing a minimal assumption on $Q^0$ initialization akin to the proof of Theorem~\ref{theorem:gap}. We will not elaborate on this again and instead directly focus on its approximation error term. As shown in Lemma~\ref{lemma:stable_no_ipr_appendix} or Eq.~(\ref{equ:no_regularization}), the approximation error term is the discounted sum of the infinity norm of the approximation error at each iteration $\sum_{j=1}^k\gamma^{k-j}\|\epsilon_{j}\|_\infty$, which cannot converge to 0 and thus is non-eliminable. Furthermore, this term often initially decays slowly since $\gamma$ usually tends to 1. Therefore, if the initial approximation error $\epsilon_0$ caused by offline pretraining is pretty large, the effects of $\gamma^k\|\epsilon_0\|_\infty$ might cause severe instability to the initial finetuning.
% Therefore, directly finetuning online policy without any regularization may lead to severe instability due to the approximation error at OOD regions induced during offline pretraining. 
This explains why directly finetuning an offline pretrained policy with online RL typically leads to an initial performance drop and requires additional regularization to stabilize the training process.

\textbf{Optimality}. \ \ To stabilize the online finetuning, previous policy constraint based offline-to-online RL approaches typically constrain the finetuning policy in a fixed constraint set $\Pi$  that is induced by the behavior policy $\mu$, the pretrained policy $\pi_0$ or the replay buffer $\mathcal{B}$\footnote{The constraint set induced by $\mathcal{B}$ slowly changes with filling in new transitions, but is much slower than the one induced by $\pi_k$.}. However, as discussed in previous works~\cite{kumar2019stabilizing, li2022distance, wusupported}, optimizing in a fixed constraint set typically lead to large optimality gap as Lemma~\ref{lemma:fixed_gap} (Theorem 4.1 in~\cite{kumar2019stabilizing}, Theorem 3 in~\cite{li2022distance}) shows:

\begin{lemma}
(Suboptimality induced by fixed policy regularization).
% \ljx{shorten the lemma description}
Define $Q^*_\Pi$ is the optimal value obtained at a constrained policy set $\Pi$. $Q^k$ is the action-value of policy $\pi_k$ obtained at $k$-th iteration by iterating Eq.~(\ref{equ:Q_operator})-(\ref{equ:original_policy}), but in the set $\Pi$, \textit{i.e.},$\pi_{k}\leftarrow \arg\max_{\pi\in\Pi}\mathbb{E}_{a\sim\pi}[Q^{k-1}(s,a)]$. Then we always have a suboptimality gap:

\begin{equation}
    \|Q^*-Q^{k}\|_\infty\le\frac{\|Q^*-Q^*_\Pi\|_\infty}{1-\gamma}, k\in N
    \label{equ:fixed_gap}
\end{equation}
\label{lemma:fixed_gap}
\end{lemma}

Note that the RHS of Eq.~(\ref{equ:fixed_gap}) has no correlation with the iteration times $k$ and hence can not be eliminated, which may cause a large suboptimality gap when the constraint policy set $\Pi$ is highly suboptimal. On the contrary, Theorem~\ref{theorem:gap} provides an intuitive insight that the suboptimality gap can be minimized to zero as $k\rightarrow\infty$ even with the existence of iterative policy regularization. Therefore, \textit{PROTO} can be perceived as an "optimistic" conservatism, which will not lead to suboptimal performances caused by over-conservatism. This can be observed in Figure~\ref{fig:frozon} that constraining on a fixed potentially suboptimal constraint set may result in suboptimal performances while allowing the constraint set to actively update can yield near-optimal performances. It is also worth mentioning that the complete version of Eq.~(\ref{equ:fixed_gap}) contains how approximation error and distributional shifts affect final performances, but we leave them behind to ease the readers to catch the main difference between fixed policy regularization and iterative policy regularization.

\begin{figure}[h]
    \centering
    \includegraphics[width=0.8\textwidth]{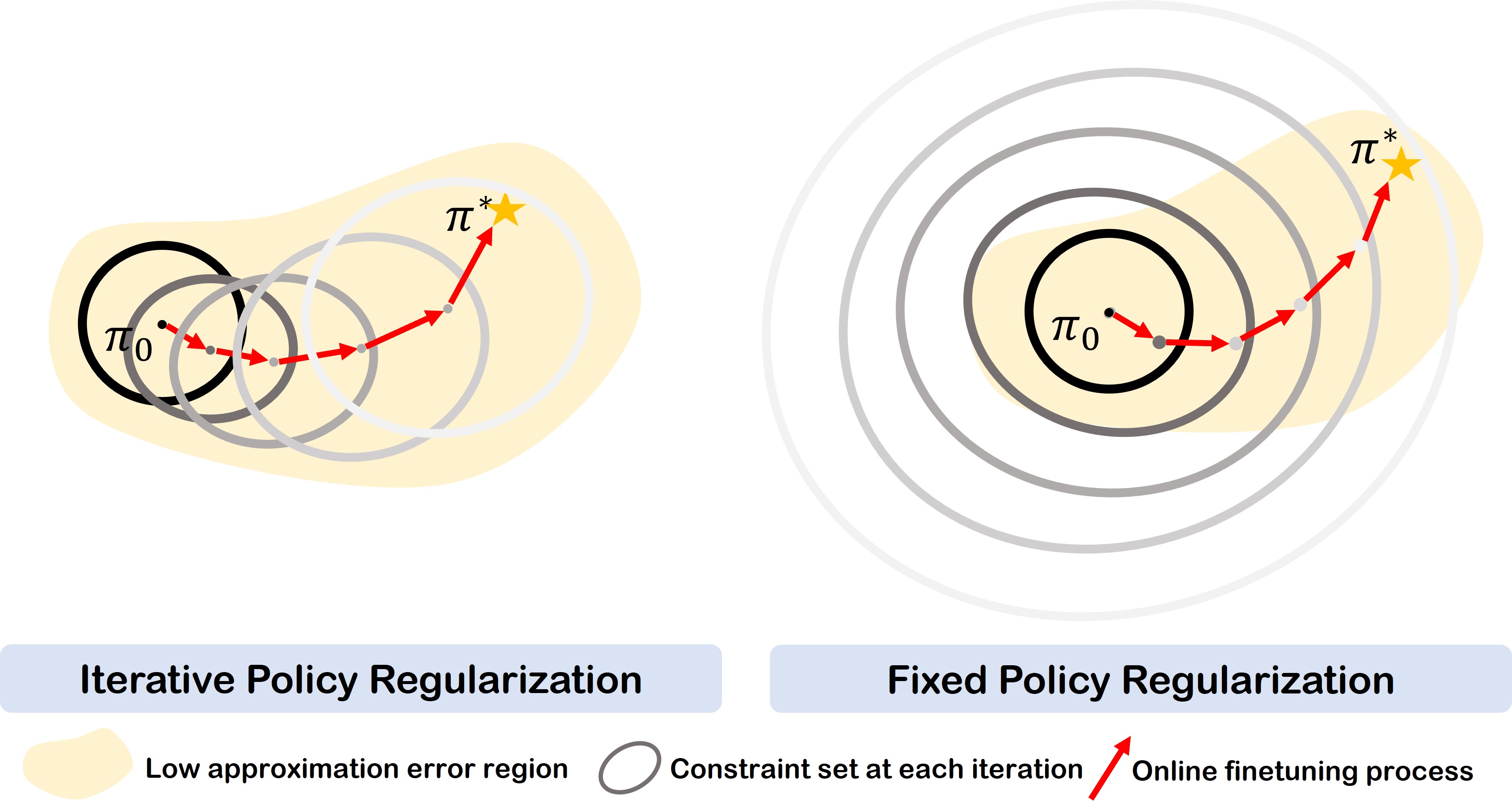}
    \caption{Illustration of iterative policy regularization \textit{v.s} fixed policy regularization. In order to achieve the same optimal policy, fixed policy regularization requires a relaxation of conservatism strength while also being more susceptible to potential approximation errors compared to iterative policy regularization.}
    \label{fig:trust_region_illustration}
\end{figure}

\textbf{Intuitive Illustration}. To alleviate the over-conservatism caused by fixed policy regularization, previous studies typically anneal the conservatism strength. However, observe from Figure~\ref{fig:frozon}, Figure~\ref{fig:appendix_iterative_vs_fixed_aggregated} and Figure~\ref{fig:appendix_iterative_vs_fixed} that even with the conservatism annealing, fixed policy regularization still underperforms iterative policy regularization. Apart from the theoretical interpretation, we give an intuitive illustration in Figure~\ref{fig:trust_region_illustration} to further explain why simply annealing the conservatism strength is not enough to obtain optimal policy.

% In Figure~\ref{fig:trust_region_illustration}, the size of the constraint set represents the conservatism strength determined by $\alpha$, where a large $\alpha$ induces strong conservatism, thus a strict and small constraint set, a small $\alpha$ otherwise. As illustrated in Figure~\ref{fig:trust_region_illustration}, to achieve the optimal performance, iterative policy regularization can achieve optimal performance via gradually deviating from the initial policy while keep the conservatism strength less changed, while fixed policy regularization requires a far more relaxed conservatism to obtain the same optimal performance. However, relaxed conservatism will include more regions that attains large approximation error, making the online finetuning more susceptible to potential approximation errors compared to iterative policy regularization.

In Figure~\ref{fig:trust_region_illustration}, the size of the constraint set represents the conservatism strength determined by $\alpha$. A larger $\alpha$ induces stronger conservatism, resulting in a strict and small constraint set, while a smaller $\alpha$ implies a more relaxed conservatism and a larger constraint set. As depicted in Figure~\ref{fig:trust_region_illustration}, to achieve optimal performance, iterative policy regularization gradually deviates from the initial policy while maintaining relatively less-changed conservatism strength. In contrast, fixed policy regularization requires significantly relaxed conservatism to attain the same optimal performance. Moreover, this relaxed conservatism includes more regions that experience large approximation errors, rendering online finetuning more susceptible to potential approximation errors compared to iterative policy regularization.

% Theorem~\ref{theorem:gap} provides an intuitive insight that the suboptimality gap can be minimized to zero as $k\rightarrow\infty$ even with the existence of iterative policy regularization. Therefore, \textit{PROTO} can be perceived as an "optimistic" conservatism, which will not lead to suboptimal performances caused by over-conservatism. This is obviously different compared with previous policy constraint based offline-to-online RL approaches.

\section{Experimental Details}
\label{sec:experimental_details}
This section outlines the experimental details to reproduce the main results in our paper. We'll also open source our code for other researchers to better understand our paper.

\subsection{Task Description}
\label{subsec:task_description}

% \begin{figure}[h]
% \begin{minipage}[t]{0.375\textwidth}
%     \centering
%     \subfloat[Pen]{\includegraphics[width=0.32\textwidth]{Figure/Task/pen.png}}
%     \subfloat[Hammer]{\includegraphics[width=0.32\textwidth]{Figure/Task/hammer.png}}
%     \subfloat[Door]{\includegraphics[width=0.32\textwidth]{Figure/Task/door.png}}
%     \caption{\small Adroit manipulation}
% \end{minipage}
% \begin{minipage}{0.25\textwidth}
%     \centering
%     \subfloat[Antmaze medium]{\includegraphics[width=0.495\textwidth]{Figure/Task/ant_medium.png}}
%     \subfloat[Antmaze large]{\includegraphics[width=0.495\textwidth]{Figure/Task/ant_large.png}}
% \caption{\small Antmaze Navigation}
% \end{minipage}
% \begin{minipage}{0.37\textwidth}
%     \centering
%     \subfloat[Hopper]{\includegraphics[width=0.32\textwidth]{Figure/Task/hopper.jpg}}
%     \subfloat[Halfcheetah]{\includegraphics[width=0.32\textwidth]{Figure/Task/halfcheetah.jpg}}
%     \subfloat[Walker2d]{\includegraphics[width=0.32\textwidth]{Figure/Task/walker2d.jpg}}
% \caption{Mujoco Locomotion}
% \end{minipage}
% \end{figure}

% \begin{wrapfigure}{r}{5cm}
%     \begin{minipage}[t]{0.375\textwidth}
%     \centering
%     {\includegraphics[width=0.32\textwidth]{Figure/Task/pen.png}}
%     {\includegraphics[width=0.32\textwidth]{Figure/Task/hammer.png}}
%     {\includegraphics[width=0.32\textwidth]{Figure/Task/door.png}}
%     \caption{\small Adroit manipulation}
% \end{minipage}
% \end{wrapfigure}
\textbf{Adroit Manipulation}. Adroit manipulation contains 3 domains: \textit{pen, door, hammer}, where the RL agent is required to solve dexterous manipulation tasks including rotating a pen in specific directions, opening a door, and hammering a nail, respectively. The offline datasets are \textit{human-v0} datasets in D4RL~\cite{fu2020d4rl} benchmark, which only contain a few successful non-markovian human demonstrations and thus is pretty difficult for most offline RL approaches to acquire reasonable pretraining performances.

% \begin{figure}[h]
%     \centering
%     \subfloat[Pen]{\includegraphics[width=0.28\textwidth]{Figure/Task/pen.png}}
%     \hfill
%     \subfloat[Hammer]{\includegraphics[width=0.275\textwidth]{Figure/Task/hammer.png}}
%     \hfill
%     \subfloat[Door]{\includegraphics[width=0.28\textwidth]{Figure/Task/door.png}}
%     \caption{Adroit manipulation tasks}
%     \label{fig:adroit_tasks}
% \end{figure}

% \begin{wrapfigure}{r}{3.5cm}
% \begin{minipage}{0.25\textwidth}
%     \centering
%     {\includegraphics[width=0.48\textwidth]{Figure/Task/ant_medium.png}}
%     {\includegraphics[width=0.48\textwidth]{Figure/Task/ant_large.png}}
% \caption{\small Antmaze Navigation}
% \end{minipage}
% \end{wrapfigure}
\textbf{Antmaze Navigation}. 
Antmaze navigation consists of two domains, namely \textit{medium} and \textit{large}, each with two datasets from the D4RL~\cite{fu2020d4rl} benchmark: \textit{play-v2} and \textit{diverse-v2}. In each domain, the objective is for an ant to learn how to walk and navigate from the starting point to the destination in a maze environment, with only sparse rewards provided. This task poses a challenge for online RL algorithms to explore high-quality data effectively without the support of offline datasets or additional domain knowledge.

% \begin{figure}[h]
%     \centering
%     \subfloat[Antmaze-medium]{\includegraphics[width=0.28\textwidth]{Figure/Task/ant_medium.png}}
%     \hspace{+5mm}
%     \subfloat[Antmaze-large]{\includegraphics[width=0.275\textwidth]{Figure/Task/ant_large.png}}
%     \caption{Antmaze mavigation tasks}
%     \label{fig:antmaze_tasks}
% \end{figure}

% \begin{wrapfigure}{r}{5cm}
% \begin{minipage}{0.37\textwidth}
%     \centering
%     {\includegraphics[width=0.32\textwidth]{Figure/Task/hopper.jpg}}
%     {\includegraphics[width=0.32\textwidth]{Figure/Task/halfcheetah.jpg}}
%     {\includegraphics[width=0.32\textwidth]{Figure/Task/walker2d.jpg}}
% \caption{Mujoco Locomotion}
% \end{minipage}
% \end{wrapfigure}
\textbf{MuJoCo Locomotion}. MuJoCo locomotion encompasses several standard locomotion tasks commonly utilized in RL research, such as \textit{Hopper, Halfcheetah, Walker2d}. In each task, the RL agent is tasked with controlling a robot to achieve forward movement. The D4RL~\cite{fu2020d4rl} benchmark provides three types of datasets with varying quality for each task: \textit{random-v2, medium-v2, medium-replay-v2}.

% \begin{figure}[h]
%     \centering
%     \subfloat[Hopper]{\includegraphics[width=0.28\textwidth]{Figure/Task/hopper.jpg}}
%     \hfill
%     \subfloat[Halfcheetah]{\includegraphics[width=0.275\textwidth]{Figure/Task/halfcheetah.jpg}}
%     \hfill
%     \subfloat[Walker2d]{\includegraphics[width=0.28\textwidth]{Figure/Task/walker2d.jpg}}
%     \caption{Mujoco locomotion tasks}
%     \label{fig:D4RL_MuJoCo_tasks}
% \end{figure}

\subsection{Hyper-parameters}
\label{subdec:hyperparameters}
\textbf{Online finetuning hyper-parameters.} \ \textit{PROTO} has two hyper-parameters: the Polyak averaging speed $\tau$ and the conservatism linear decay speed $\eta$. Although with 2 hyper-parameters, we find that choosing $\eta$ around 0.9 can achieve substantially stable performances for all 16 tasks (see Figure~\ref{fig:annealing_speed} for detailed results), where $\eta=1$ means $\alpha$ anneals to 0 with $10^6$ online samples and $\eta=0.9$ means $\alpha$ anneals to 0 with $\frac{10^6}{0.9}$ online samples. Therefore, we adopt a non-parametric
treatment by setting $\eta=0.9$. We report the detailed setup in Table~\ref{tab:tau_para}.

\begin{table}[h]
    \centering
    \caption{Online finetuning hyper-parameters}
    \begin{tabular}{cccc}
    \toprule
      Task & $\tau$ & $\eta$\\
    \midrule
      All MuJoCo locomotion & $5e-3$ & 0.9\\
      All Antmaze navigation & $5e-5$ & 0.9\\
      All Adroit manipulation & $5e-5$ & 0.9\\
    \bottomrule
    \end{tabular}
    \label{tab:tau_para}
\end{table}

\begin{table}[h]
    \centering
    \caption{Offline pretraining hyper-parameters}
    \begin{tabular}{cccc}
    \toprule
      Task & Initial $\alpha$ & Pretraining Steps\\
    \midrule
      All MuJoCo locomotion &2.0 & 0.1M\\
      All Antmaze navigation &0.5 & 0.2M\\
      All Adroit manipulation &2.0 & 0.1M\\
    \bottomrule
    \end{tabular}
    \label{tab:alpha}
\end{table}

Table~\ref{tab:tau_para} shows that $\textit{PROTO}$ has only two groups of hyper-parameters across 16 tasks with only varying the Polyak averaging speed $\tau$. Moreover, we ablate on the Polyak averaging speed $\tau$ in the range of $[0.5\tau, 2\tau]$ in Figure~\ref{fig:tau_speed} and find that \textit{PROTO} is robust to such large hyper-parameter variations. As a result, \textit{PROTO} is easy to tune, which is critical and desired for RL community since online evaluations for parameter tuning are generally costly.

\textbf{Offline pretraining hyper-parameters.} \ In our paper, we pretrain the policy using EQL~\cite{sql} (equivalents to XQL~\cite{garg2023extreme}), and hence we directly adopt the conservatism strength coefficient $\alpha$ in EQL paper~\cite{sql} to pretrain policy and use it to initialize $\alpha$ in our paper. In terms of pretraining steps, We find that performing 0.1M pretraining steps for all MuJoCo locomotion and Adroit manipulation tasks and 0.2M pretraining steps for all Antmaze navigation tasks already attain good initialization. Therefore, we do not pretrain further to reduce computational costs. Please see Table~\ref{tab:alpha} for detailed parameter choice\footnote{We also initialize $\alpha$ according to Table~\ref{tab:alpha} when using BC to pretrain policy.}. Similar to EQL, both PEX and IQL pretrain based on in-sample learning. Therefore, we also pretrain PEX and IQL with the pretraining steps according to Table~\ref{tab:alpha}.

\subsection{Additional Experimental Details}
\label{subsec:additional_experimental_details}

\textbf{Initialization of online replay buffer}. \ We initialize the online replay buffer with three different types: (1) Initialize the buffer with the entire offline dataset akin to~\cite{nair2020awac, iql}. (2) Initialize the buffer with a small set of offline datasets. (3) Conduct a separate online buffer and sample symmetrically from both offline dataset and online buffer akin to~\cite{ball2023efficient}. However, we observe that these three different types of initialization have little effect on finetuning performances. 
Whereas, we recommend to symmetrically sample from offline dataset and together a separate online dataset akin to~\cite{ball2023efficient} for future consideration to design online finetuning approaches with higher sample efficiency.
% Finally, we choose to initialize the online replay buffer with a small set of offline dataset, which enables
% symmetrically sample from offline dataset and together a separate online dataset akin to~\cite{ball2023efficient} for future consideration to design online finetuning approaches with high sample-efficiency.

\textbf{Network architecture and optimization hyper-parameters}. \ We implement all the function approximators with 2-layer MLPs with ReLU activation functions. To stabilize both offline pretraining and online finetuning processes, we add Layer-Normalization~\cite{ba2016layer} to the action-value networks and state-value networks akin to previous works~\cite{sql, garg2023extreme, ball2023efficient}. We find that Layer-Normalization may cause over-conservatism for all halfcheetah tasks, and thus we only drop Layer-Normalization when experimenting on all halfcheetah tasks. We choose Adam~\cite{kingma2015adam} as optimizer, 3e-4 as learning rate and 256 as batch size for all networks and all tasks.

\textbf{Clip-double Q and value regularization backup}. Similar to~\cite{ball2023efficient}, we also find that clip-double Q and the value regularization backup may introduce over-conservatism and cause inferior performances for some extremely difficult tasks. Therefore, we do not use these trick for some experiments as Table~\ref{tab:clip_double_Q} shows. 

\begin{table}[h]
    \centering
    \caption{Additional experiment details}
    \begin{tabular}{cccc}
    \toprule
      Task & Clip-double Q &Value regularization backup\\
    \midrule
      All Mujoco-Locomotion &\Checkmark &\Checkmark\\
      All Antmaze-Navigation &\ding{55} &\ding{55}\\
      All Adroit-Manipulation &\Checkmark &\ding{55}\\
    \bottomrule
    \end{tabular}
    \label{tab:clip_double_Q}
\end{table}

% Note that without the value regularization backup, the optimization objective in Eq.~(\ref{equ:PROTO_objective}) degenerates to an iterative policy constrained objective and still can :

% \begin{equation}
%     \begin{aligned}
% \pi_{k+1}\leftarrow\arg&\max_{\pi}\mathbb{E}\left[\sum_{t=0}^{\infty}\gamma^tr(s_t,a_t)\right], k\ge0, \\
% &{\rm s.t.} \quad {\rm D}_{\rm KL}(\pi\|\pi_k)\le\epsilon
%     \end{aligned}
% \end{equation}

\subsection{Pseudocode and computational cost}
\label{sucsec:sudocode}
This subsection presents the pseudocode when finetuning using SAC.

\begin{algorithm}[h]
    \caption{PROTO with SAC finetuning}
    \label{alg:cap}
\begin{algorithmic}
    \State \textbf{Input:} Offline dataset $\mathcal{D}$, online replay buffer $\mathcal{B}$, pretrained value networks $Q^0$, pretrained policy $\pi_0$, initial conservatism strength $\alpha$.
    % \State \textcolor{purple}{/ / Discriminator learning}
    % \State Train $h_\tau$ using $\mathcal{D}^E$ and $\mathcal{D}$ using Eq.~(\ref{discriminator}).
    \For {$k=0, 1, 2, 3,..., N$}
        \State {Collect new transition, $\mathcal{B}\leftarrow\mathcal{B}\cup\{(s, a, r, s')\}$}.
        \State {Sample mini-batch transitions $B\sim\mathcal{D}\cup\mathcal{B}$}.
        \State Update SAC action-value networks based on $B$ by subtracting $\alpha\cdot\log\frac{\pi}{\bar\pi_k}$ from target value.
        \State Update SAC policy based on $B$ by adding $\alpha\cdot\log\frac{\pi}{\bar\pi_k}$ from from actor loss.
        \State Update target value networks via polyak averaging trick
        \State Update target actor network $\bar\pi_k$ via polyak averaging trick
        \State Anneal $\alpha$ until 0.
    \EndFor
\end{algorithmic}
\end{algorithm}

We implement our approach using the JAX framework~\cite{jax2018github}. On a single RTX 3080Ti GPU, we can perform 1 million online samples and gradient steps in approximately 20 minutes for all tasks.
% \begin{algorithm}[h]
%     \caption{PROTO with TD3 finetuning}
%     \label{alg:cap}
% \begin{algorithmic}
%     \State \textbf{Input:} Offline dataset $\mathcal{D}$, online replay buffer $\mathcal{B}$, pretrained action-value networks $Q^0$, pretrained policy $\pi_0$, initial conservatism strength $\alpha$.
%     % \State \textcolor{purple}{/ / Discriminator learning}
%     % \State Train $h_\tau$ using $\mathcal{D}^E$ and $\mathcal{D}$ using Eq.~(\ref{discriminator}).
%     \For {$k=0, 1, 2, 3,..., N$}
%         \State {Collect new transition, $\mathcal{B}\leftarrow(s, a, r, s')$}.
%         \State {Sample mini-batch transitions $B\sim\mathcal{D}\cup\mathcal{B}$}.
%         \State Update action-value networks based on $B$ using Eq.~(\ref{equ:Q_operator_trust}).
%         \State Update policy based on $B$ using Eq.~(\ref{equ:policy_trust}).
%         % \iff {$k\%$}
%         \If {$k\%$update freq$=0$}
%         \State Update target value networks via polyak averaging trick
%         \State Update target actor network via polyak averaging trick
%         \EndIf
%         \State Anneal $\alpha$ until 0.
%     \EndFor
% \end{algorithmic}
% \end{algorithm}

% \newpage
\section{Ablation Study}
\label{sec:ablation_appendix}

In this section, we conduct ablation studies on the two hyper-parameters of \textit{PROTO}, the Polyak averaging speed $\tau$ and the conservatism strength annealing speed $\eta$, to investigate whether \textit{PROTO} is hyper-parameter robust. 

\begin{figure}[h]
    \centering
    \includegraphics[width=0.32\textwidth]{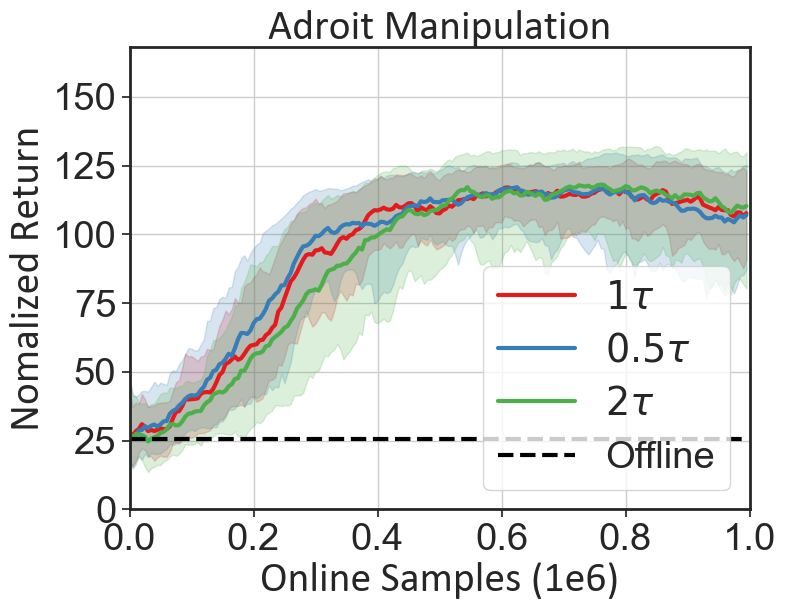}
    \includegraphics[width=0.32\textwidth]{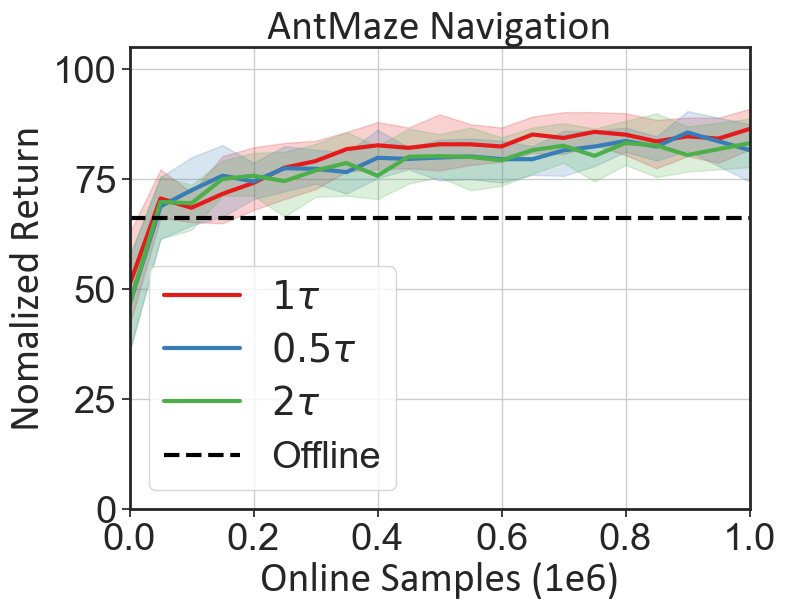}
    \includegraphics[width=0.32\textwidth]{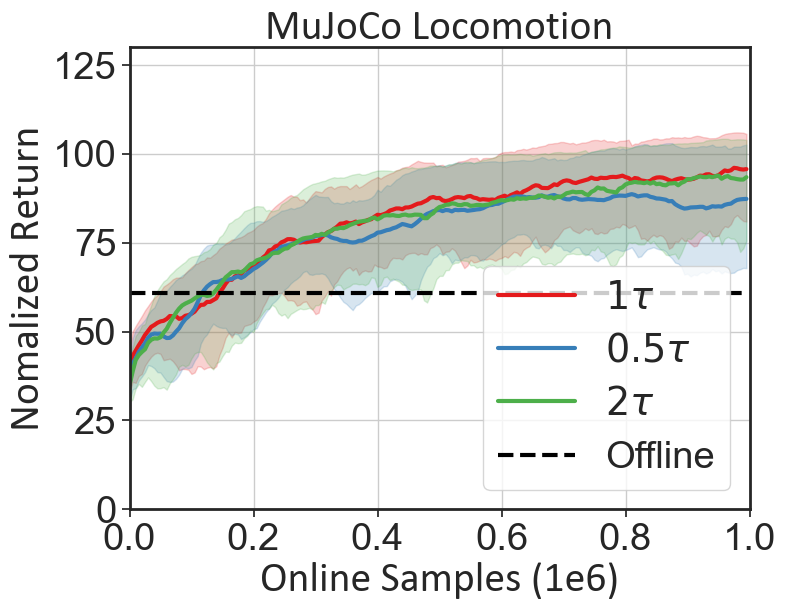}
    \caption{{Aggregated learning curves} of ablations on the polyak averaging speed.}
    \label{fig:tau_speed_aggregated}
    \includegraphics[width=0.24\textwidth]{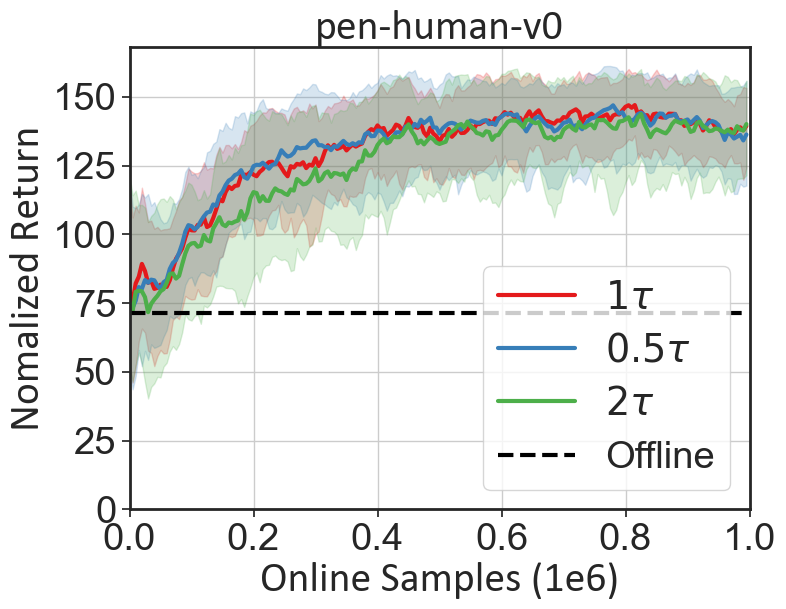}
    \includegraphics[width=0.24\textwidth]{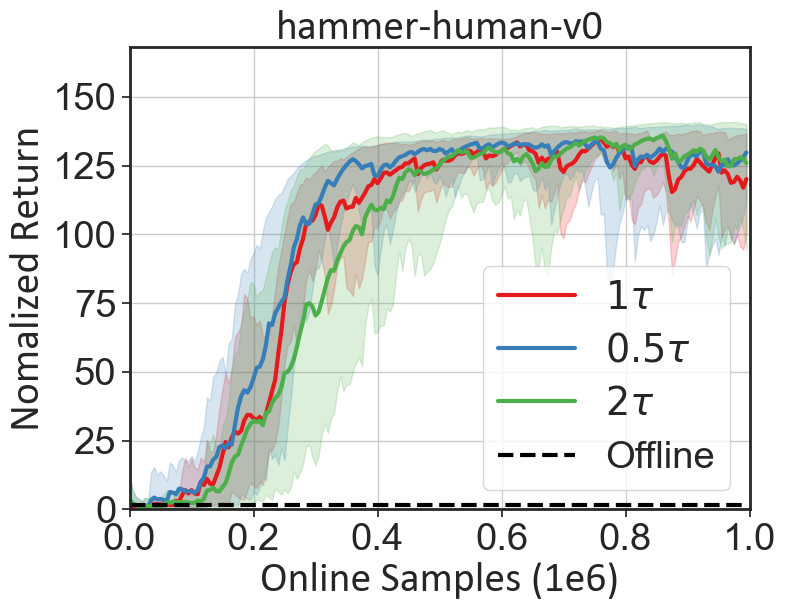}
    \includegraphics[width=0.24\textwidth]{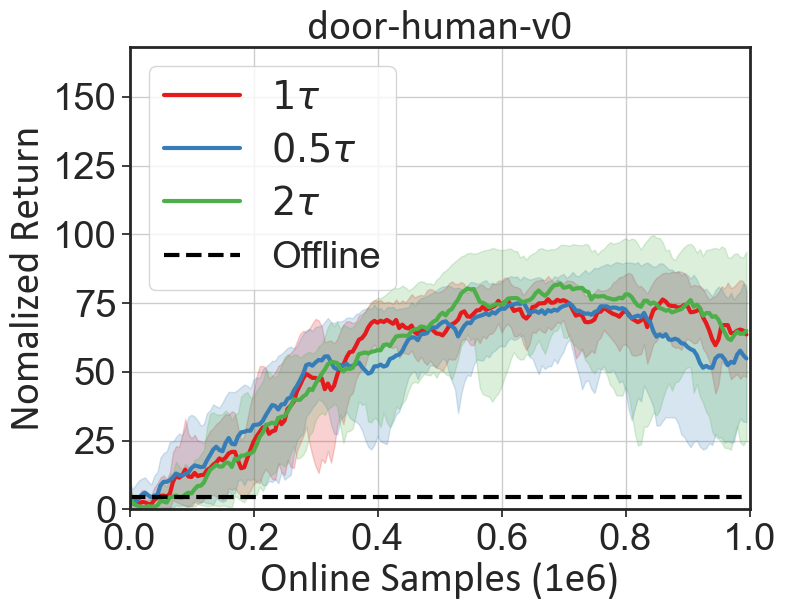}
    \includegraphics[width=0.24\textwidth]{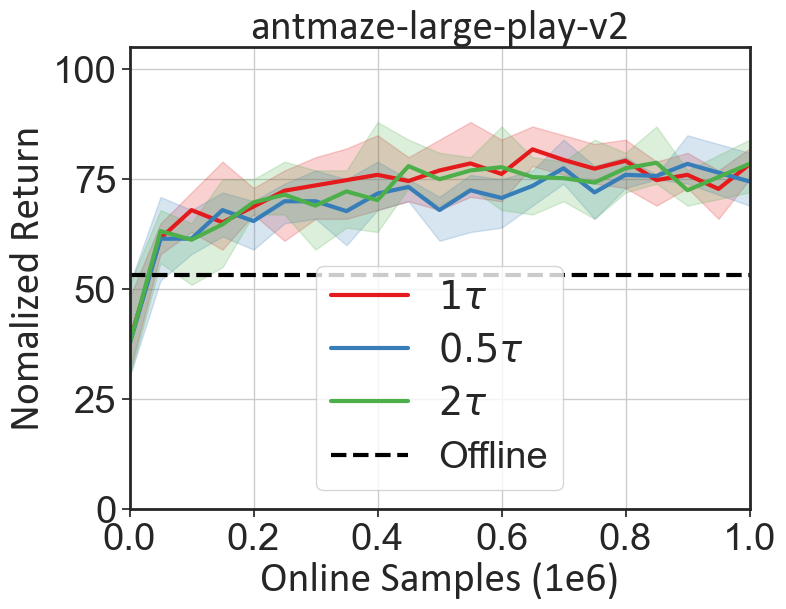}
    \includegraphics[width=0.24\textwidth]{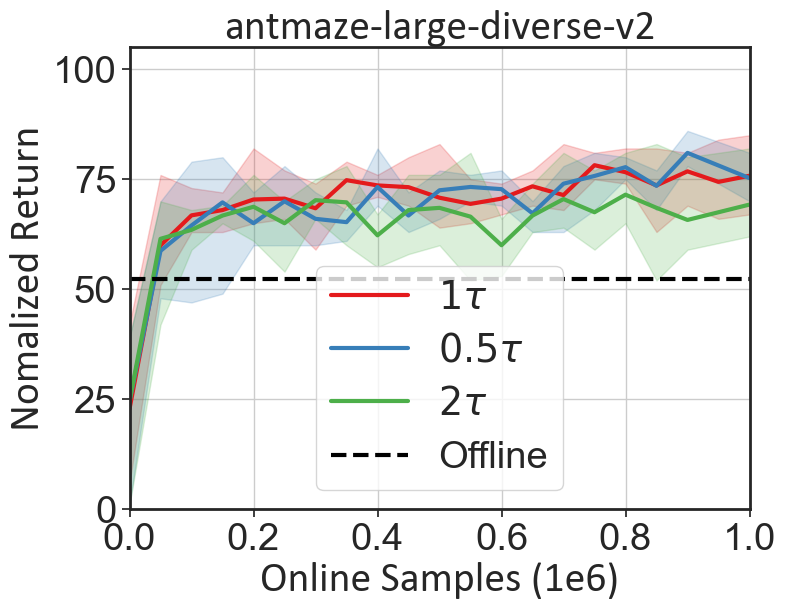}
    \includegraphics[width=0.24\textwidth]{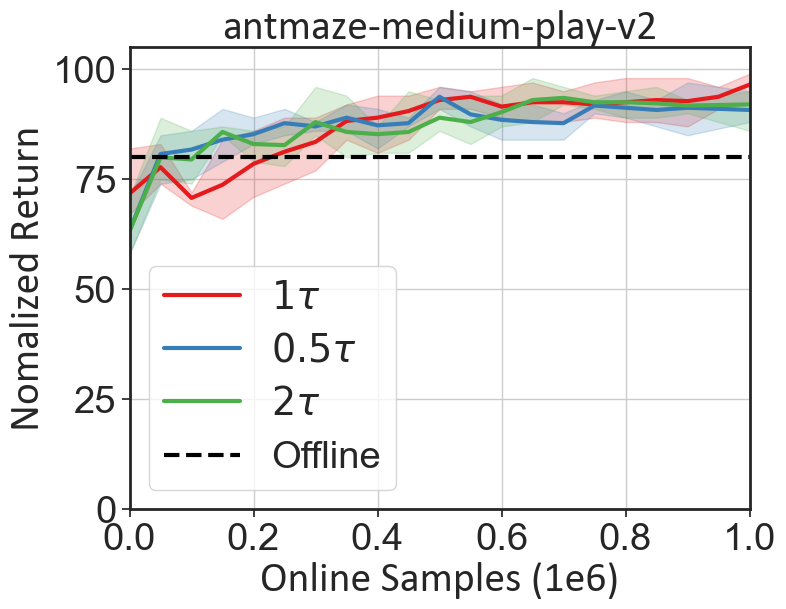}
    \includegraphics[width=0.24\textwidth]{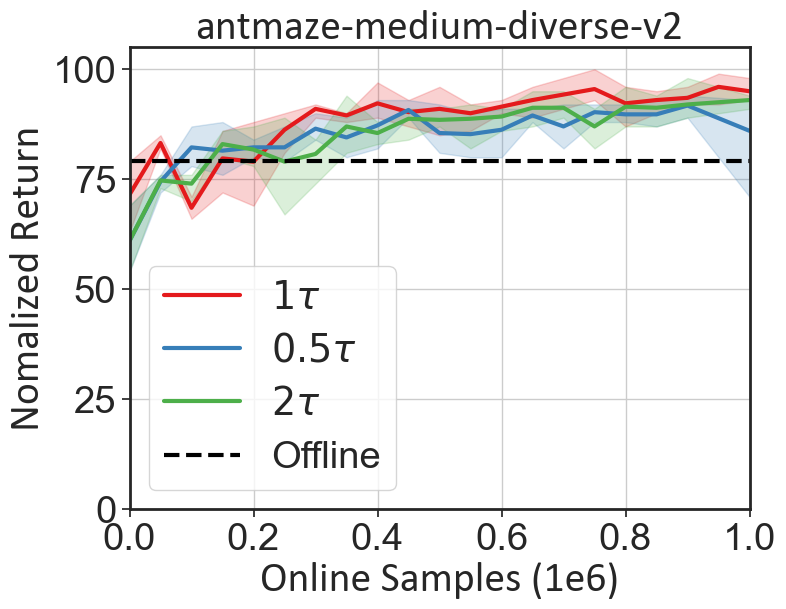}
    \includegraphics[width=0.24\textwidth]{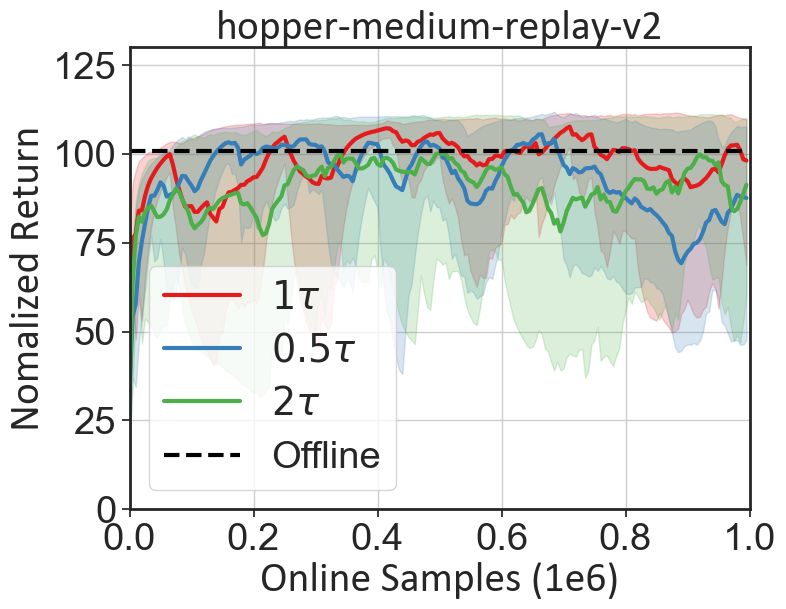}
    \includegraphics[width=0.24\textwidth]{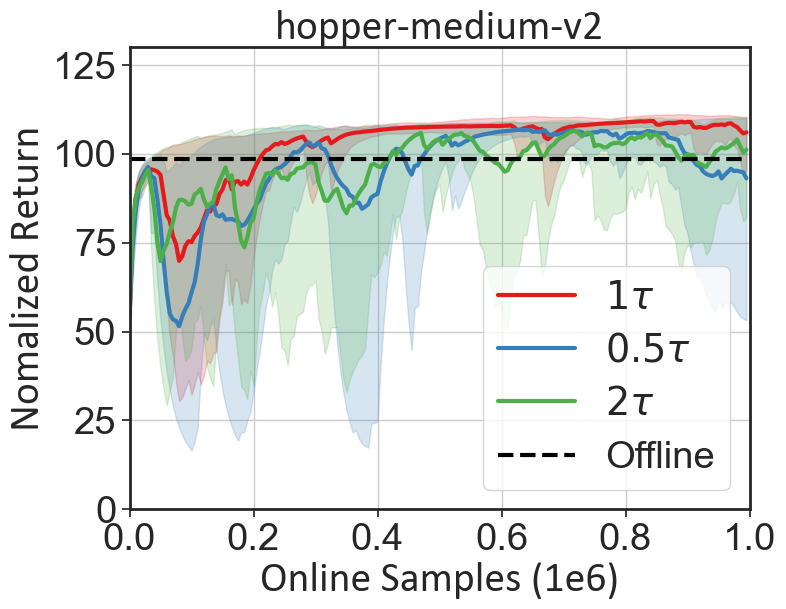}
    \includegraphics[width=0.24\textwidth]{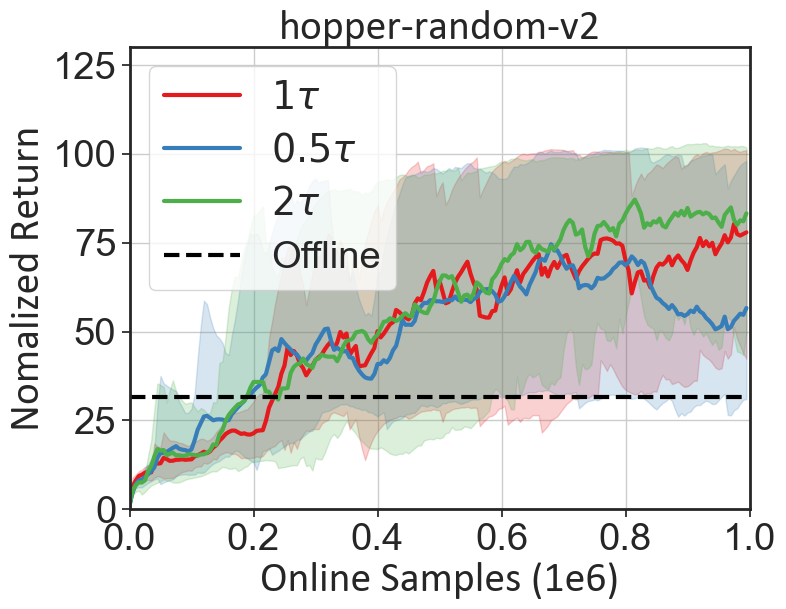}
    \includegraphics[width=0.24\textwidth]{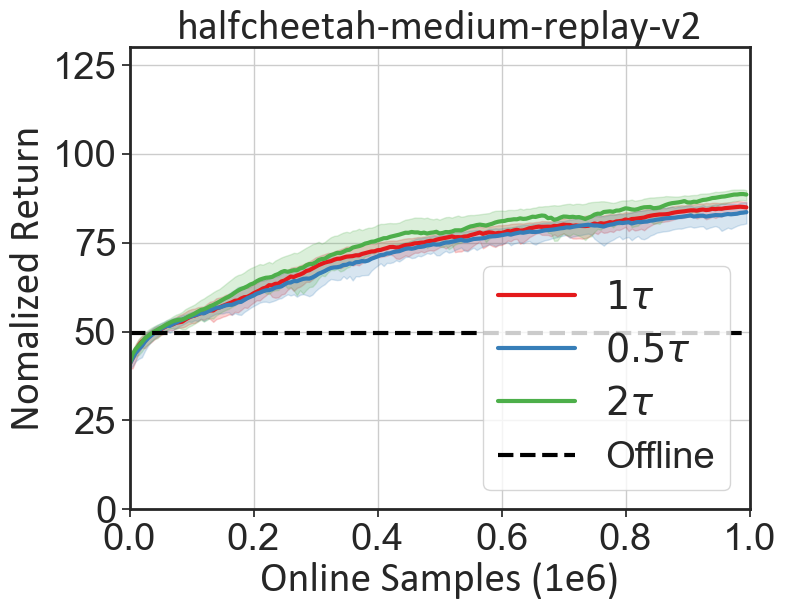}
    \includegraphics[width=0.24\textwidth]{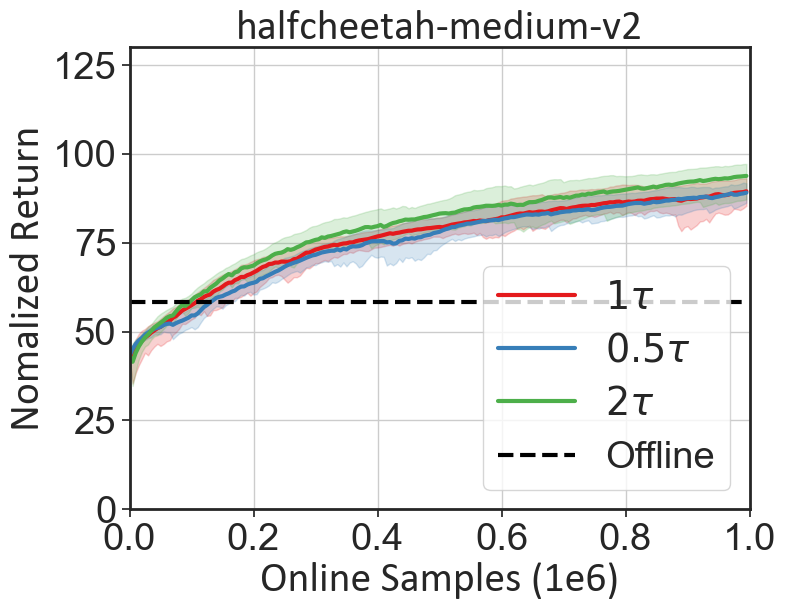}
    \includegraphics[width=0.24\textwidth]{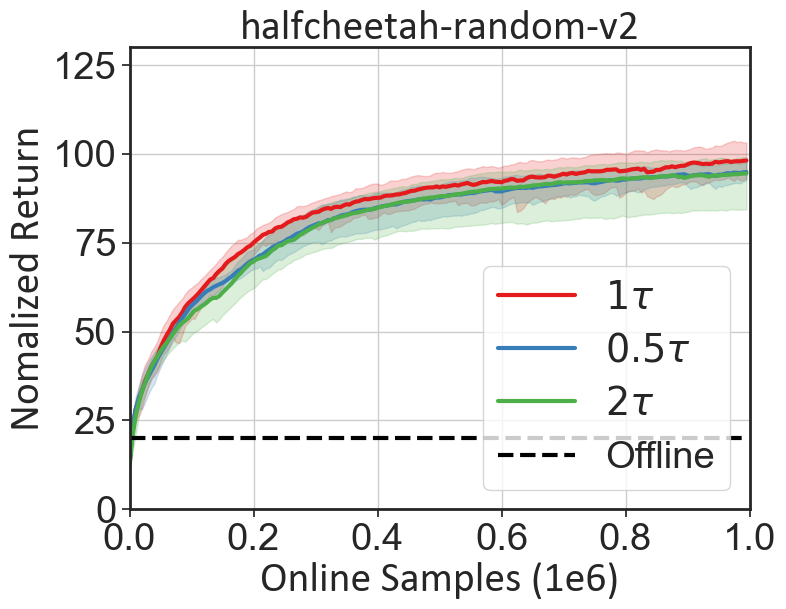}
    \includegraphics[width=0.24\textwidth]{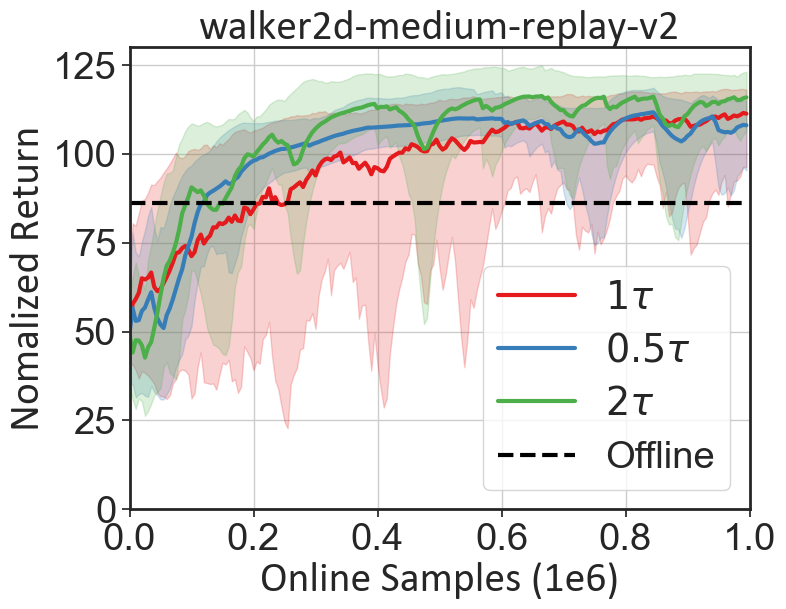}
    \includegraphics[width=0.24\textwidth]{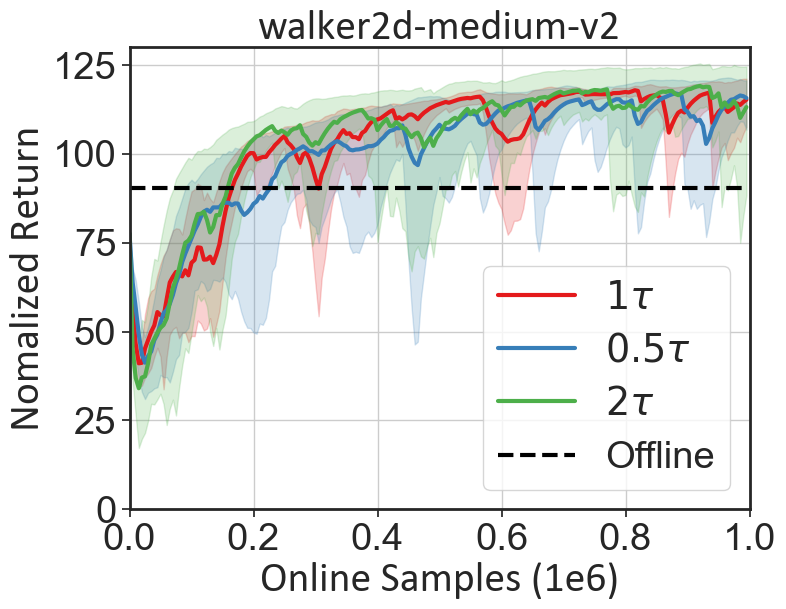}
    \includegraphics[width=0.24\textwidth]{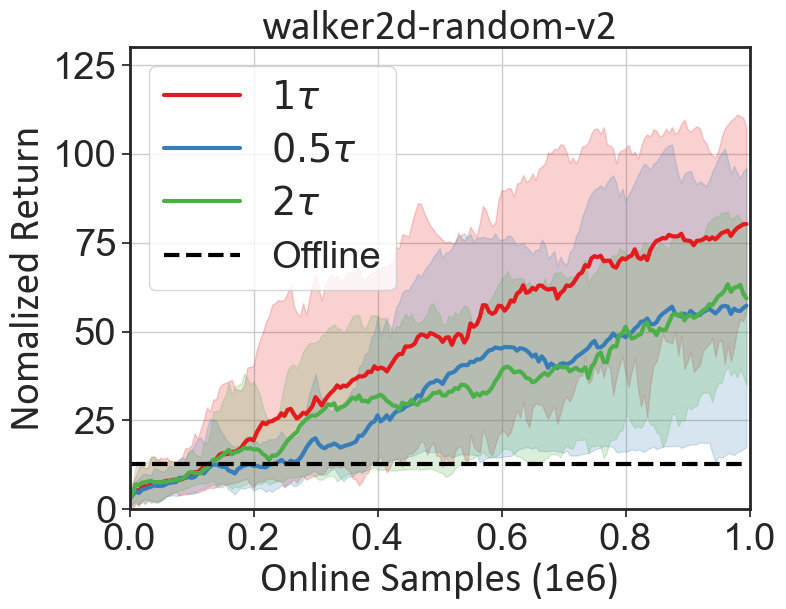}
    \caption{ Full results of ablations on the polyak averaging speed.}
    \label{fig:tau_speed}
\end{figure}

For $\tau$, we ablates on three sets of parameters: $0.5\tau, 1\tau$ and $2\tau$, where $1\tau$ is the original hyper-parameter that is used to reproduce the results in our paper including $5\times10^{-3}$ for all MuJoCo locomotion tasks and $5\times10^{-5}$ for all Adroit manipulation and Antmaze navigation tasks. $0.5\tau$ represents half the original speed and $2\tau$ denotes double the original speed. The aggregated learning curves and full results can be found in Figure~\ref{fig:tau_speed_aggregated} and Figure~\ref{fig:tau_speed}.

% Figure~\ref{fig:tau_speed_aggregated} and Figure~\ref{fig:tau_speed} demonstrate that \textit{PROTO} can obtain similar performances with the Polyak averaging speed varying from $0.5\tau$ to $2\tau$, representing its robustness to this hyper-parameter.

Figure~\ref{fig:tau_speed_aggregated} and Figure~\ref{fig:tau_speed} illustrate that \textit{PROTO} can achieve similar performances across a range of Polyak averaging speeds, including $0.5\tau$ to $2\tau$. This finding highlights the robustness of \textit{PROTO} to variations in the Polyak averaging speed ($\tau$).
\newpage

\begin{figure}[h]
    \centering
    \includegraphics[width=0.32\textwidth]{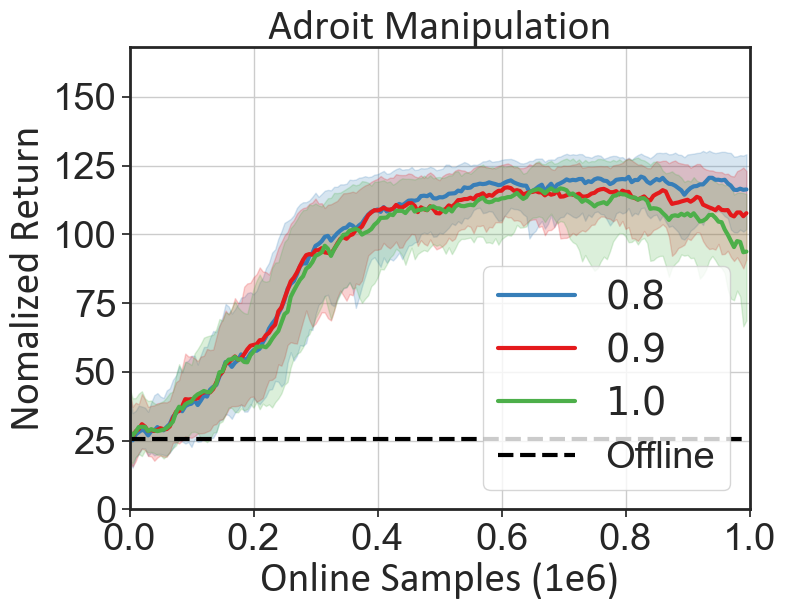}
    \includegraphics[width=0.32\textwidth]{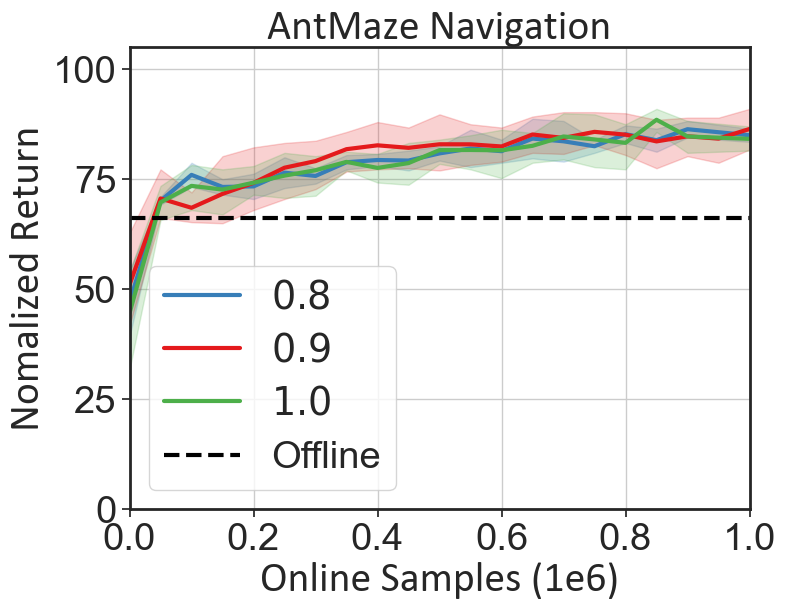}
    \includegraphics[width=0.32\textwidth]{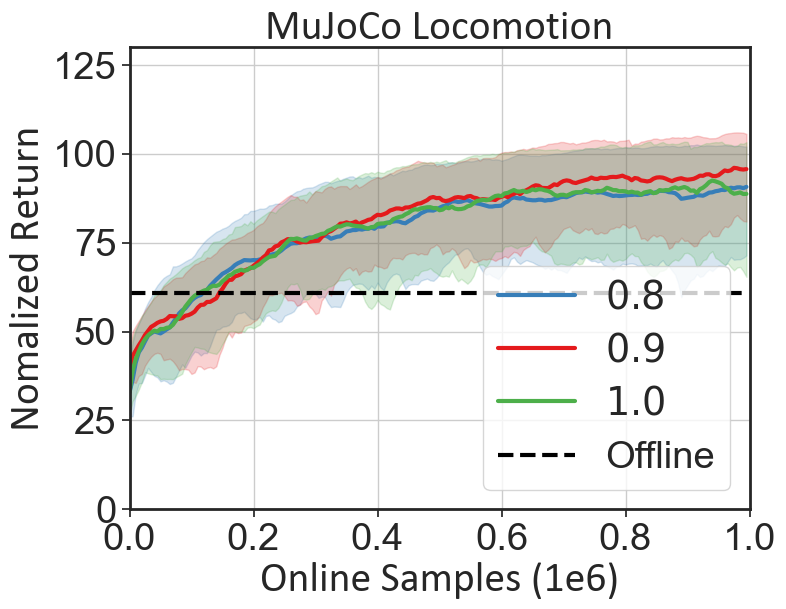}
    \caption{{Aggregated learning curves} of ablations on the conservatism annealing speed $\eta$.}
    \label{fig:annealing_speed_aggregated}
    \includegraphics[width=0.24\textwidth]{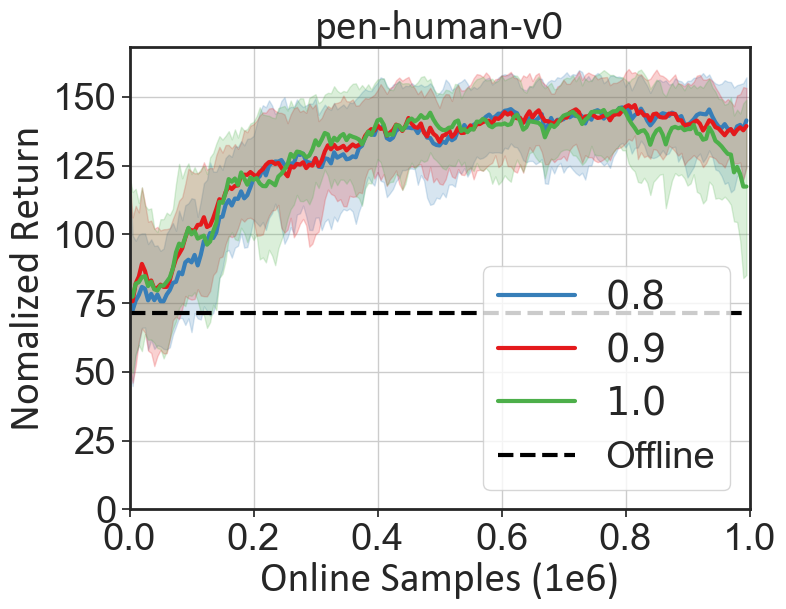}
    \includegraphics[width=0.24\textwidth]{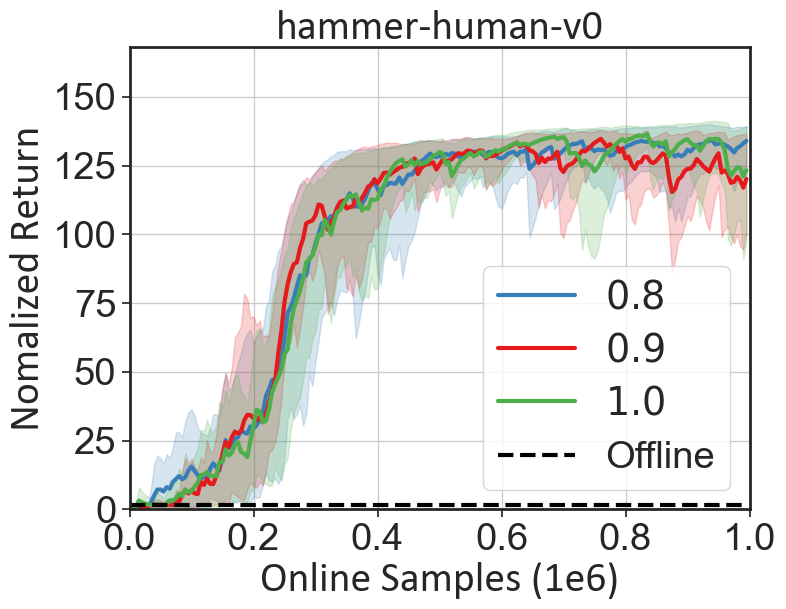}
    \includegraphics[width=0.24\textwidth]{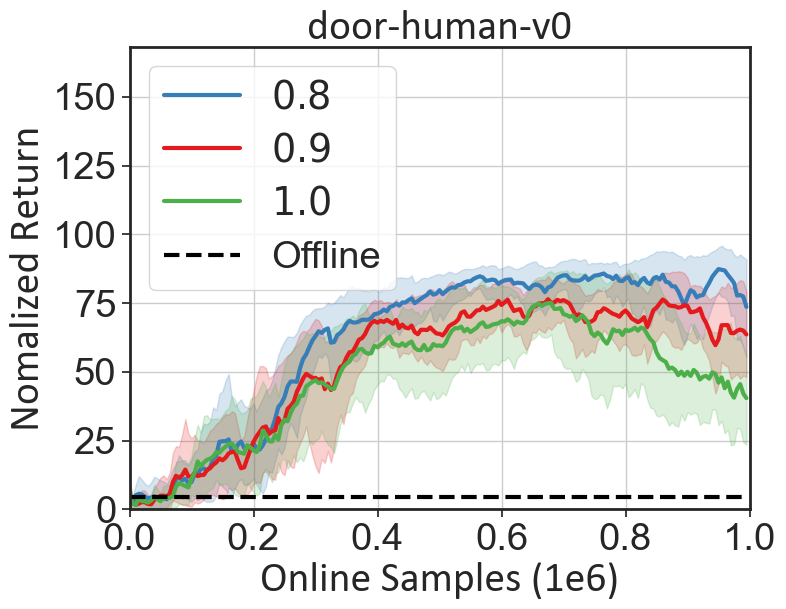}
    \includegraphics[width=0.24\textwidth]{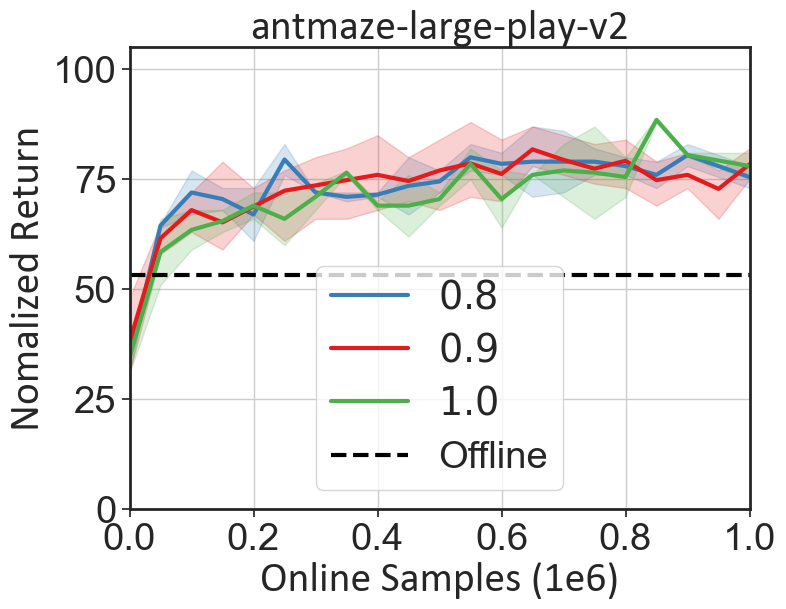}
    \includegraphics[width=0.24\textwidth]{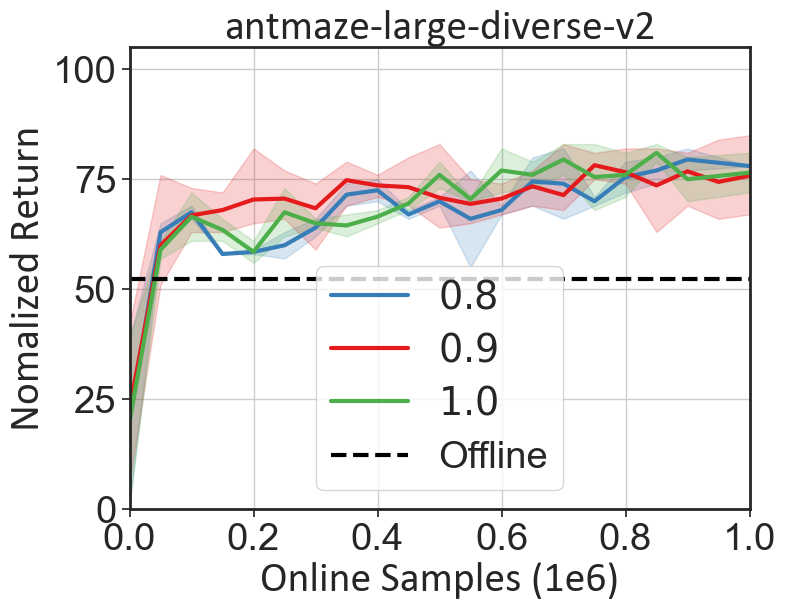}
    \includegraphics[width=0.24\textwidth]{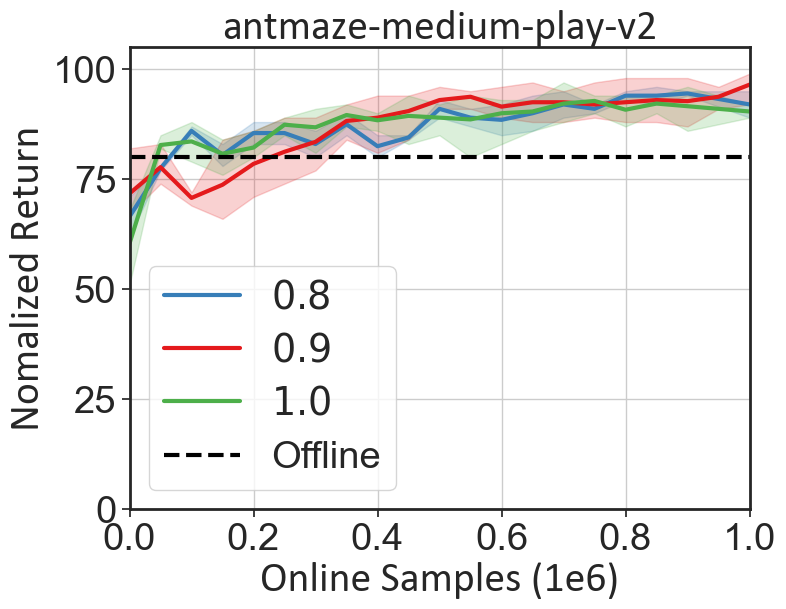}
    \includegraphics[width=0.24\textwidth]{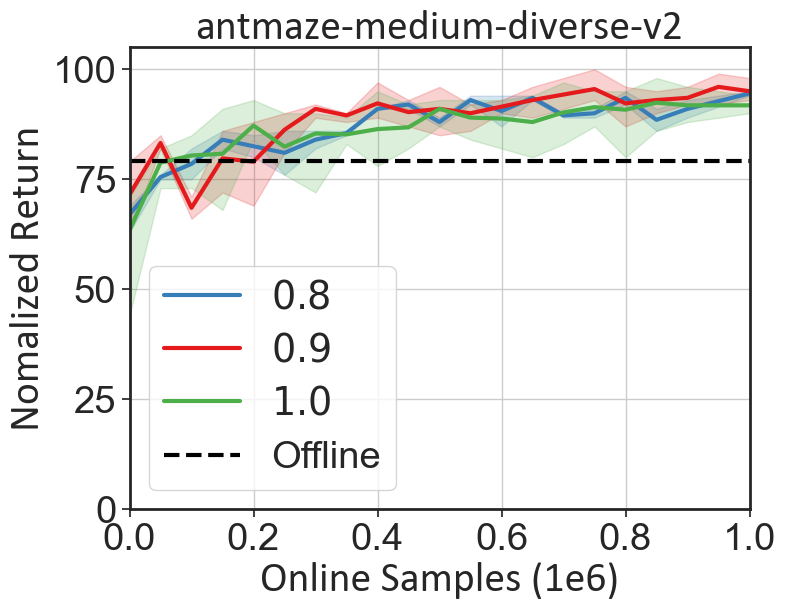}
    \includegraphics[width=0.24\textwidth]{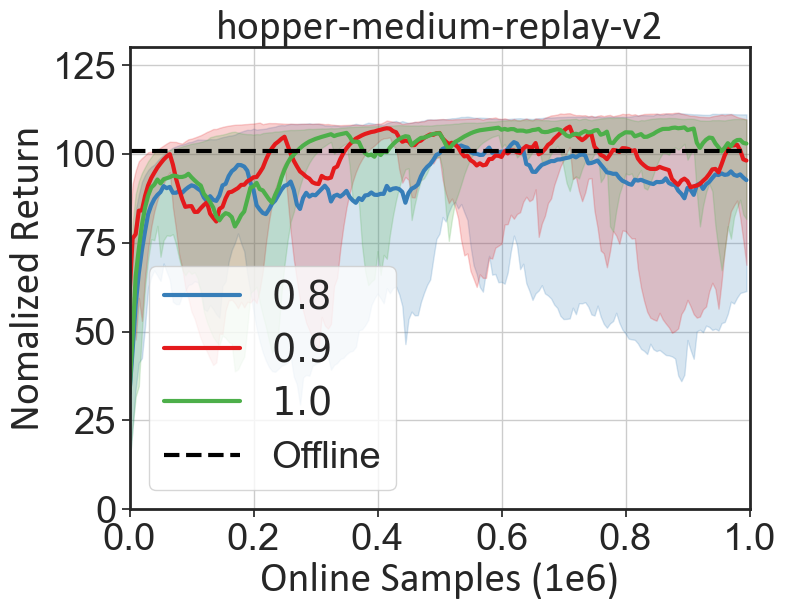}
    \includegraphics[width=0.24\textwidth]{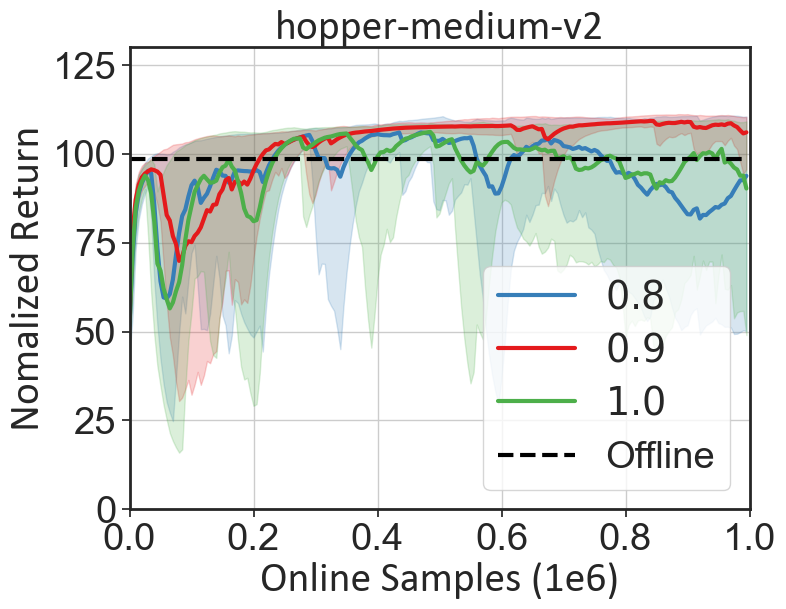}
    \includegraphics[width=0.24\textwidth]{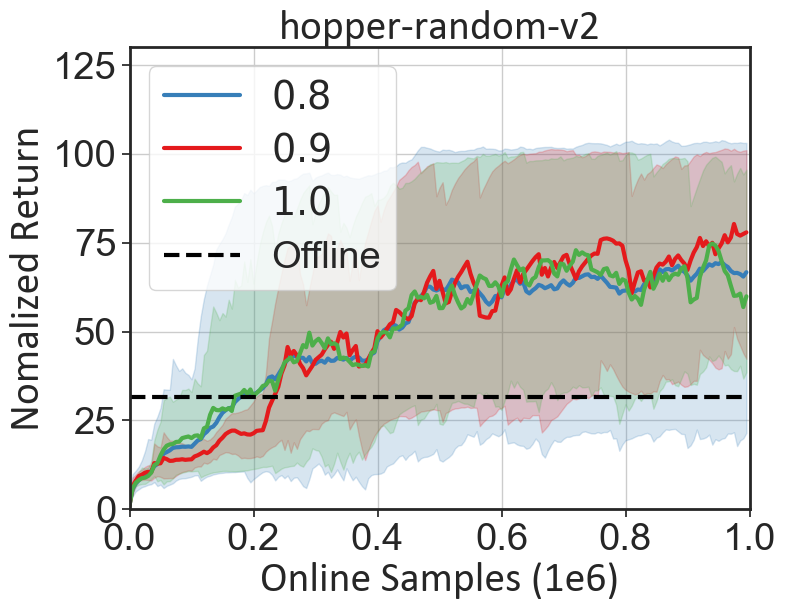}
    \includegraphics[width=0.24\textwidth]{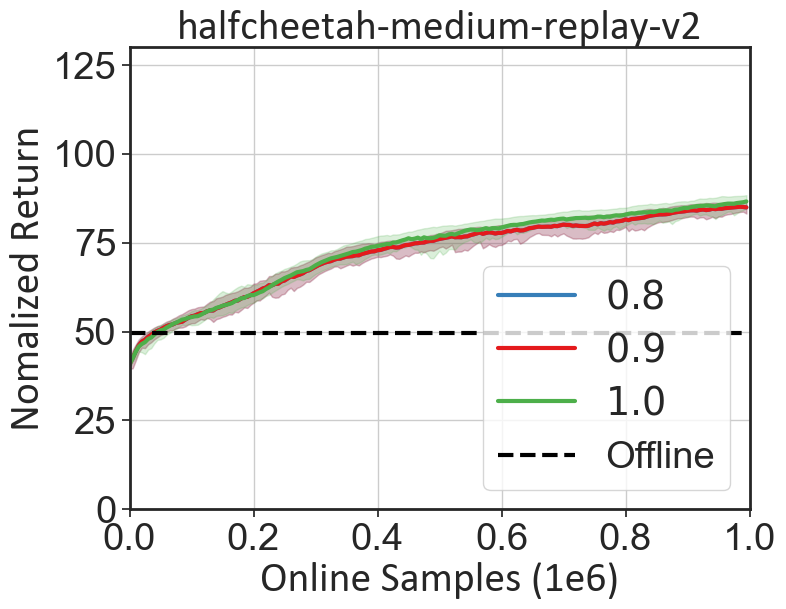}
    \includegraphics[width=0.24\textwidth]{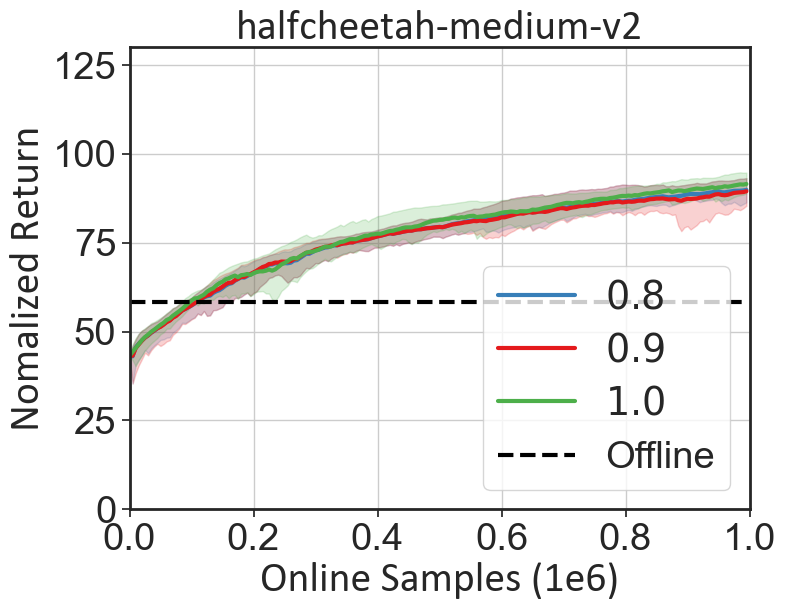}
    \includegraphics[width=0.24\textwidth]{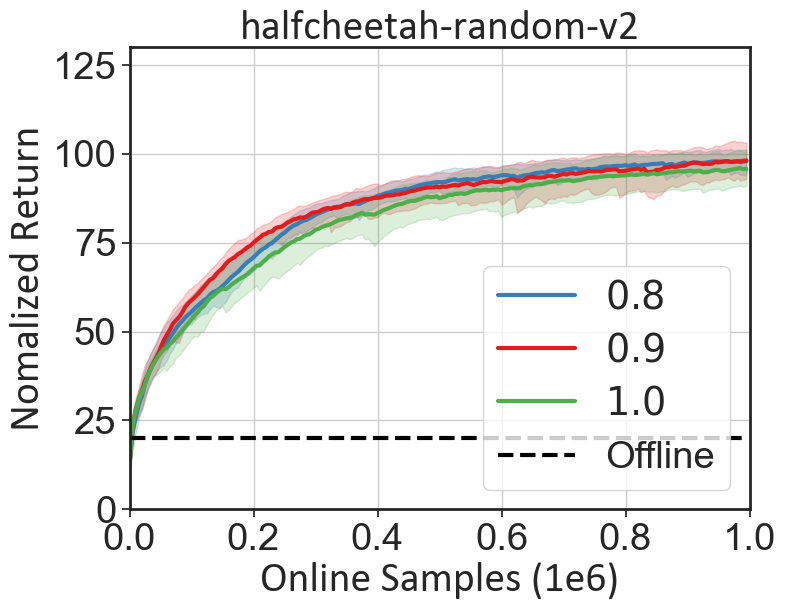}
    \includegraphics[width=0.24\textwidth]{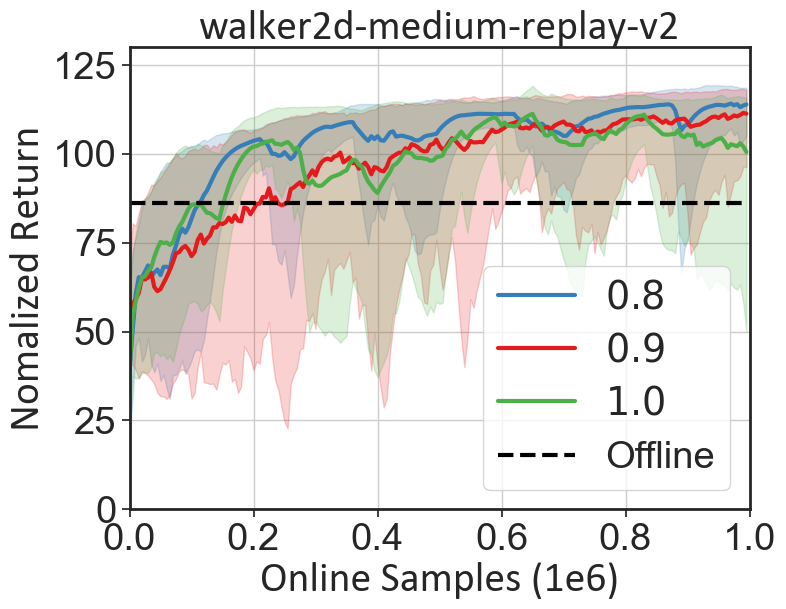}
    \includegraphics[width=0.24\textwidth]{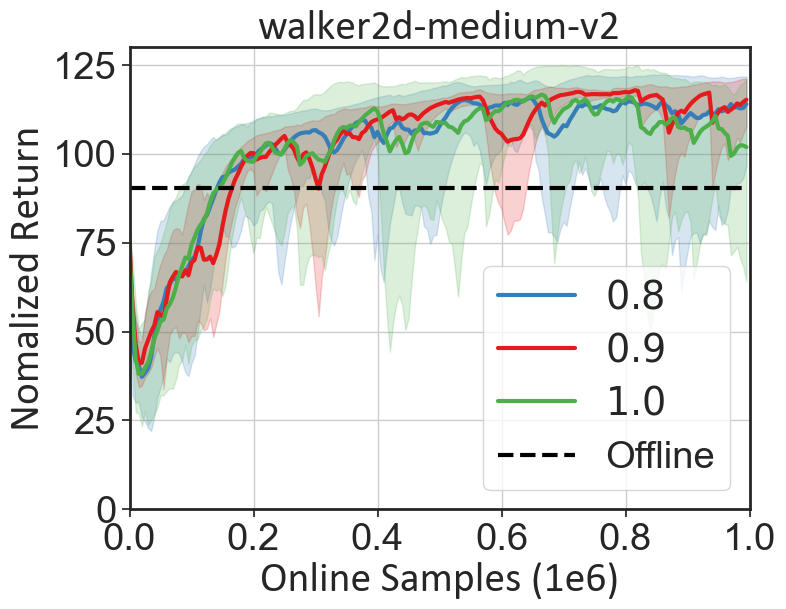}
    \includegraphics[width=0.24\textwidth]{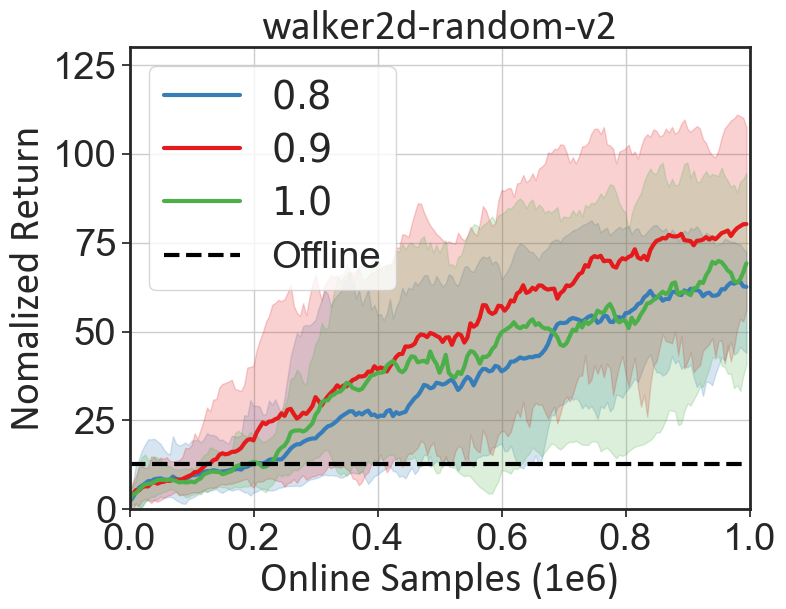}
    \caption{Full results of ablations on the conservatism annealing speed $\eta$.}
    \label{fig:annealing_speed}
\end{figure}

For the ablations on the conservatism annealing speed $\eta$, we also ablate on three sets of parameters: $0.8, 0.9$ and $1.0$, where $0.9$ is the original hyper-parameter that is used to reproduce the results in our paper and means the conservatism strength $\alpha$ anneals to 0 with $\frac{10^6}{0.9}$ online samples. $0.8$ and $1.0$ represents $\alpha$ decays to 0 with $\frac{10^6}{0.8}$ and $\frac{10^6}{1.0}$ online samples, respectively. The aggregated learning curves and full results can be found in Figure~\ref{fig:annealing_speed_aggregated} and Figure~\ref{fig:annealing_speed}.

Figure~\ref{fig:annealing_speed_aggregated} and Figure~\ref{fig:annealing_speed} demonstrate that \textit{PROTO} can obtain consistently good performances with three sets of annealing speed. In this paper, we adopt a non-parametric treatment by setting $\eta$ as 0.9 to ease the parameter tuning.

\newpage
\section{Full Results of PROTO with BC Pretraining}

\begin{figure}[h]
    \centering
    \includegraphics[width=0.7\textwidth]{Figure/Main/lc_bc/legend_bc.png}
    
    \includegraphics[width=0.32\textwidth]{Figure/Main/lc_bc/aggrated-return_adroit.png}
    \includegraphics[width=0.32\textwidth]{Figure/Main/lc_bc/aggrated-return_antmaze.png}
    \includegraphics[width=0.32\textwidth]{Figure/Main/lc_bc/aggrated-return_mujoco.png}
    \caption{{Aggregated learning curves} of online finetuning with BC pretrained policy.}
    \label{fig:appendix_bc_pretrained_aggregated}
    \includegraphics[width=0.24\textwidth]{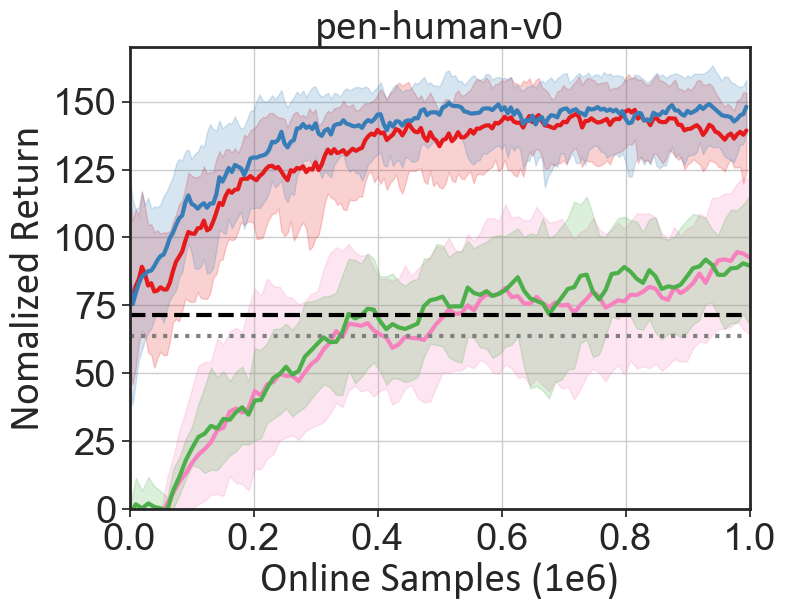}
    \includegraphics[width=0.24\textwidth]{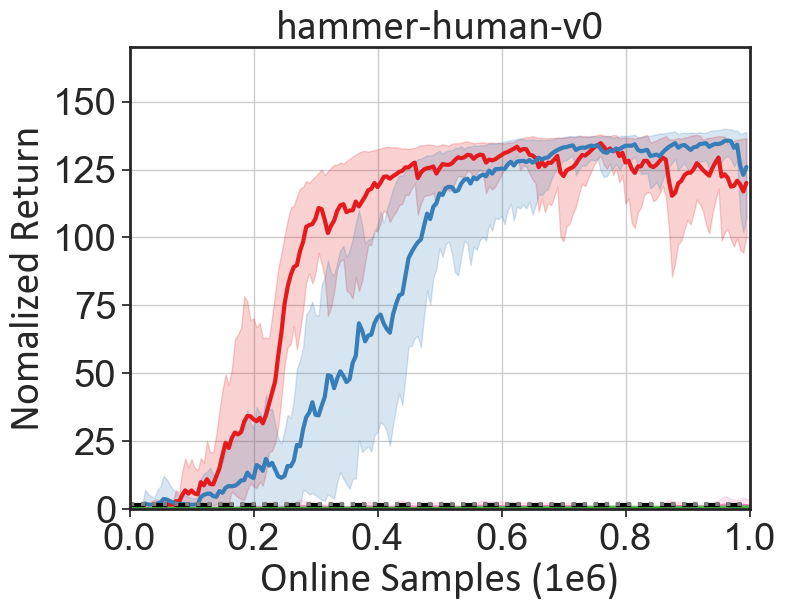}
    \includegraphics[width=0.24\textwidth]{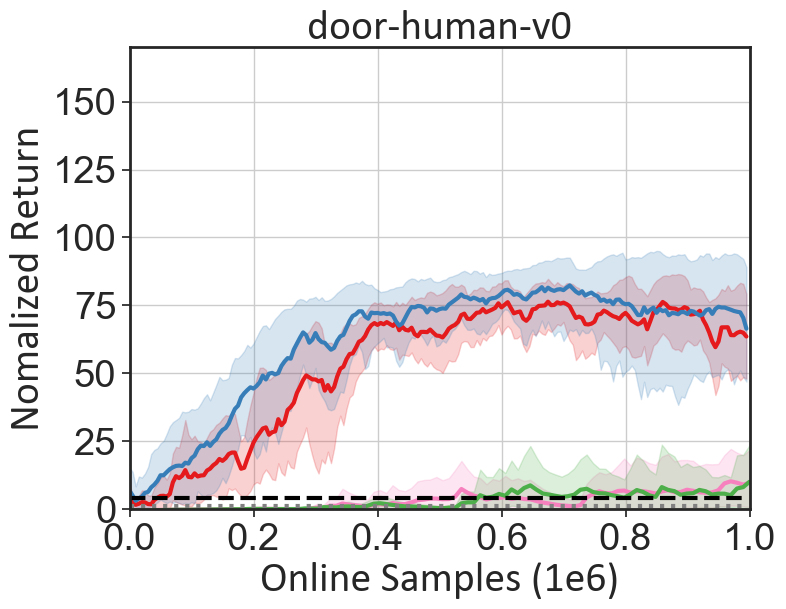}
    \includegraphics[width=0.24\textwidth]{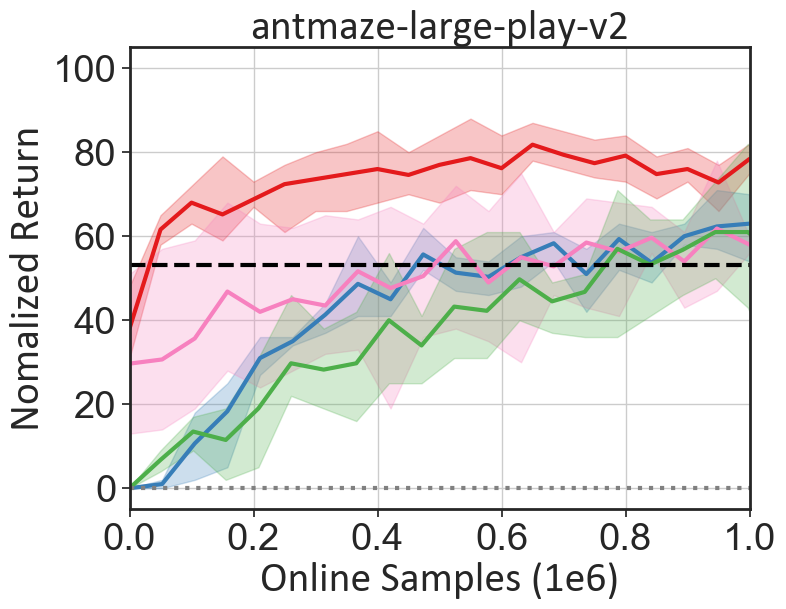}
    \includegraphics[width=0.24\textwidth]{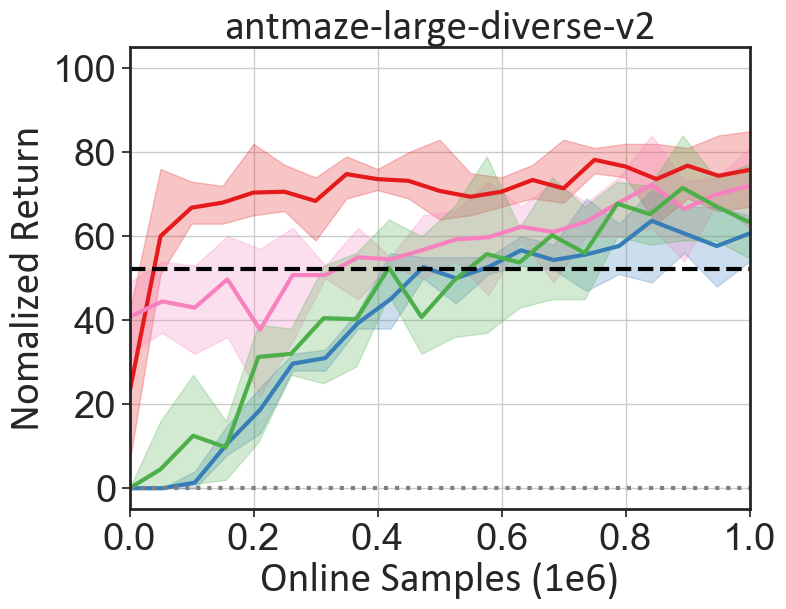}
    \includegraphics[width=0.24\textwidth]{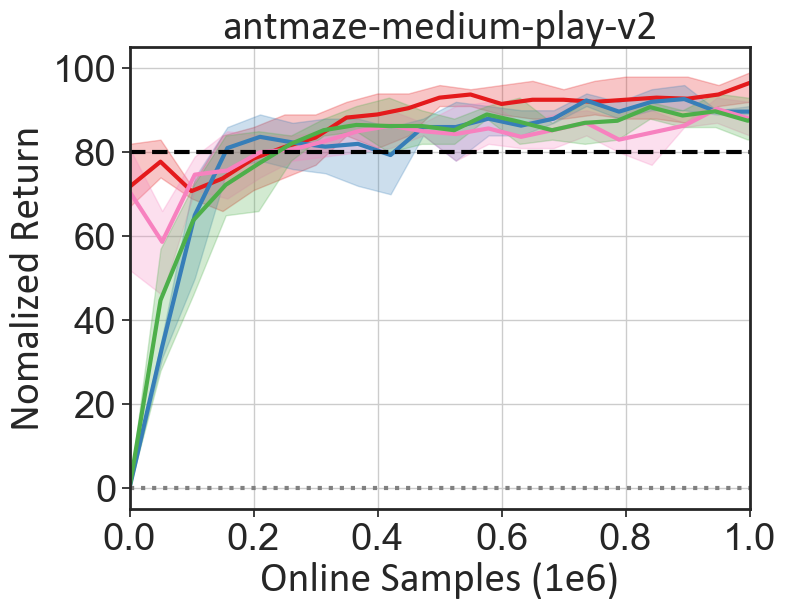}
    \includegraphics[width=0.24\textwidth]{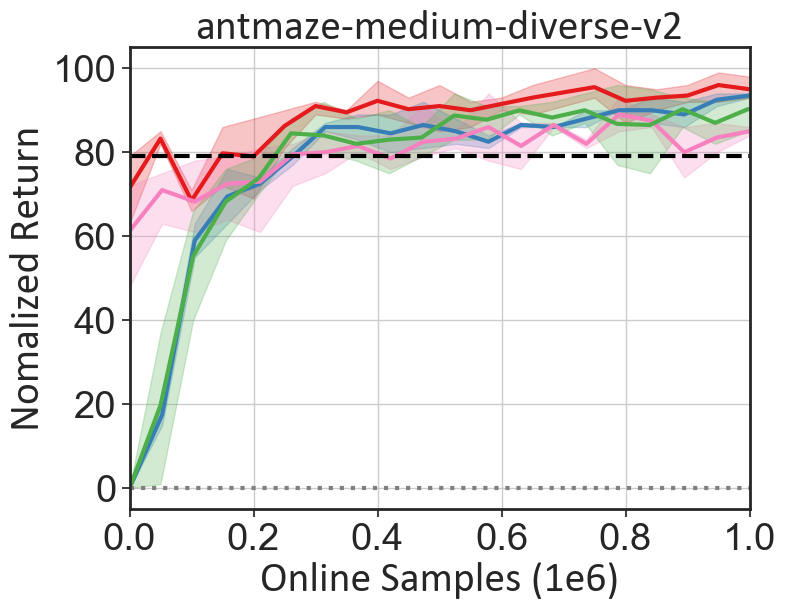}
    \includegraphics[width=0.24\textwidth]{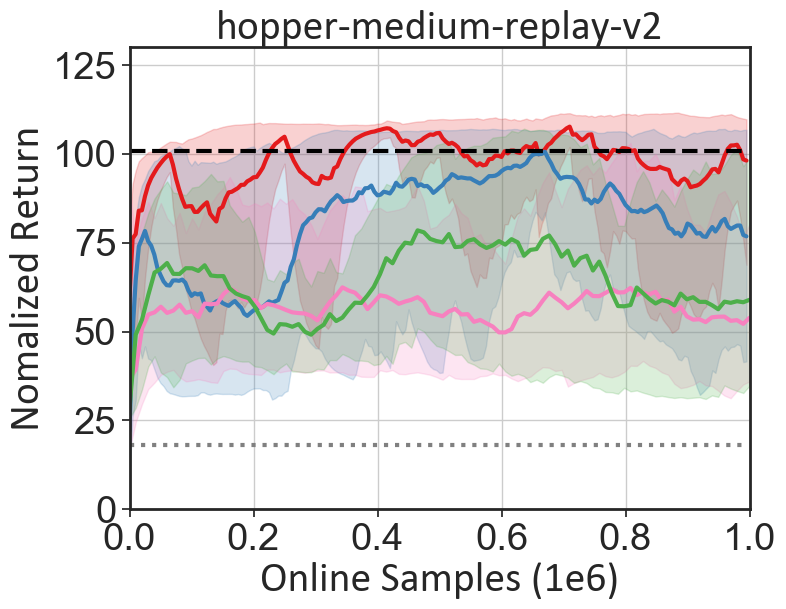}
    \includegraphics[width=0.24\textwidth]{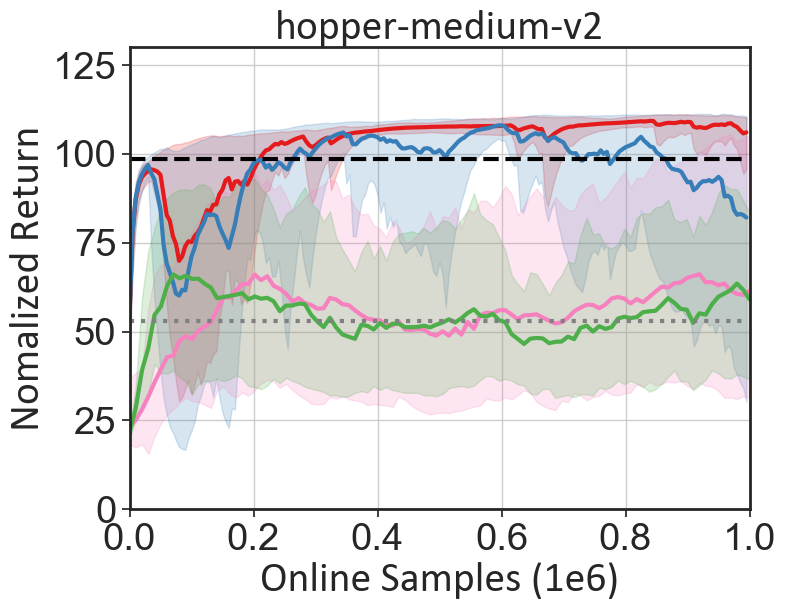}
    \includegraphics[width=0.24\textwidth]{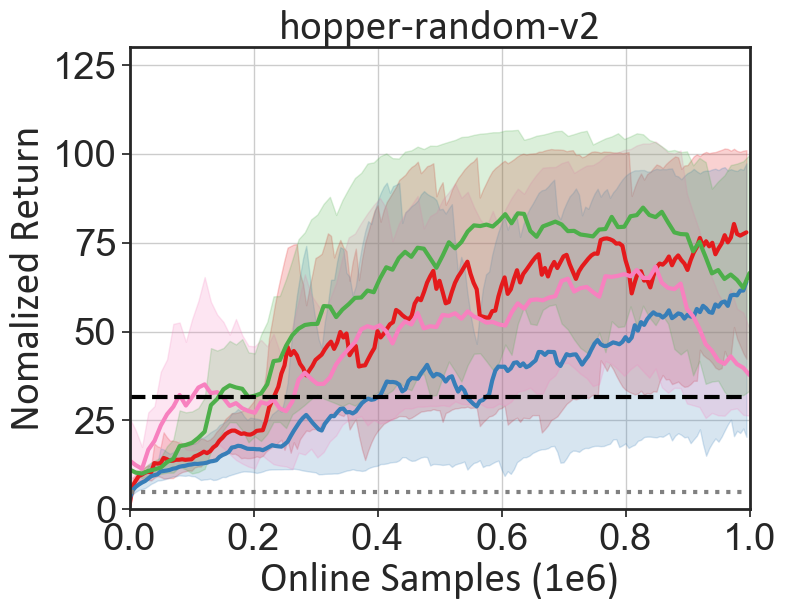}
    \includegraphics[width=0.24\textwidth]{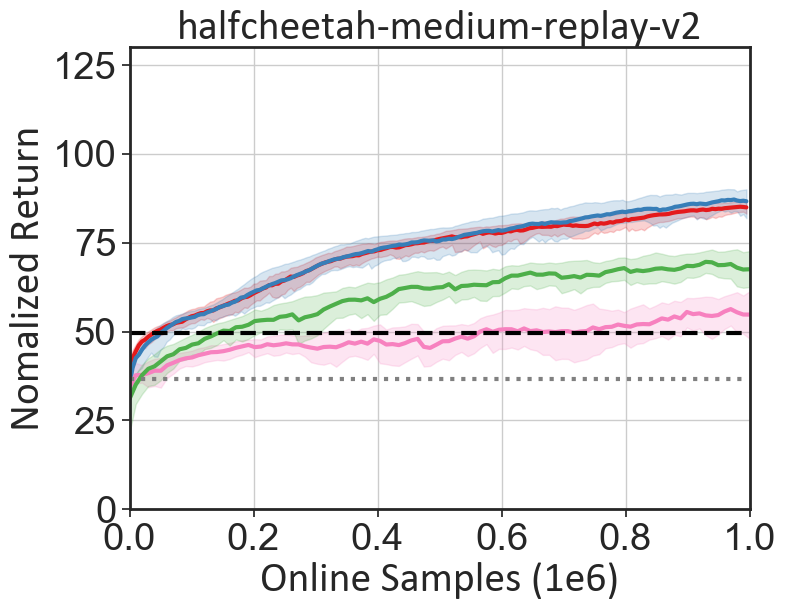}
    \includegraphics[width=0.24\textwidth]{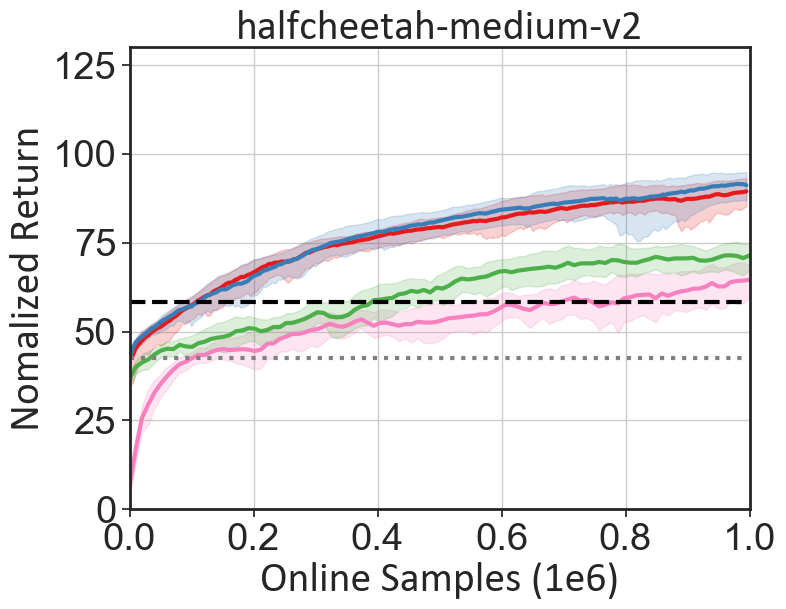}
    \includegraphics[width=0.24\textwidth]{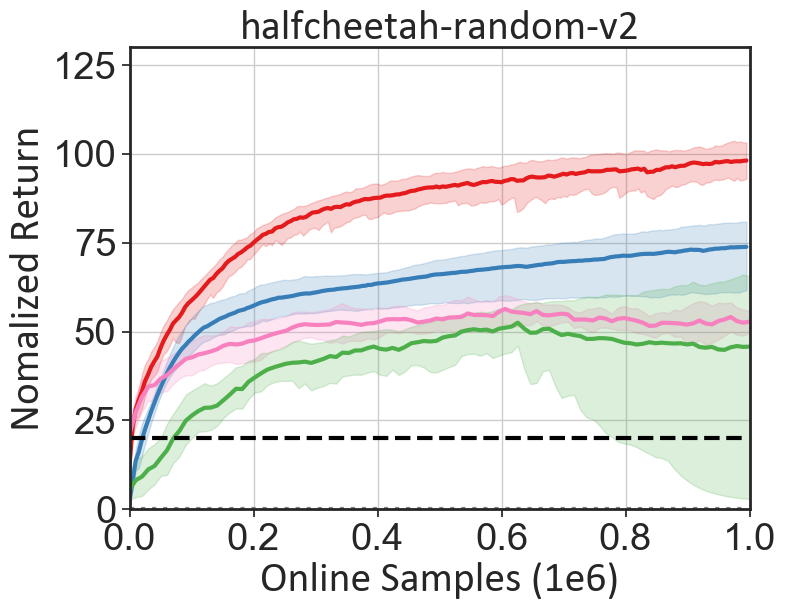}
    \includegraphics[width=0.24\textwidth]{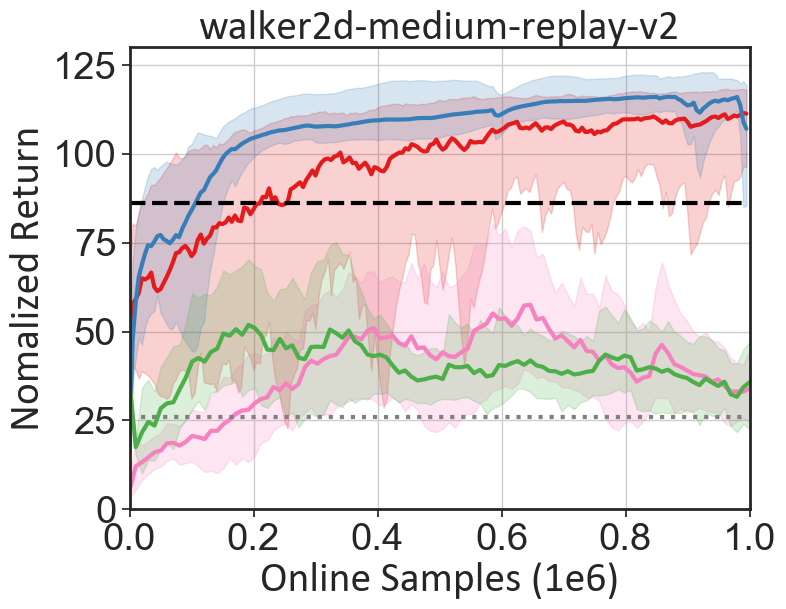}
    \includegraphics[width=0.24\textwidth]{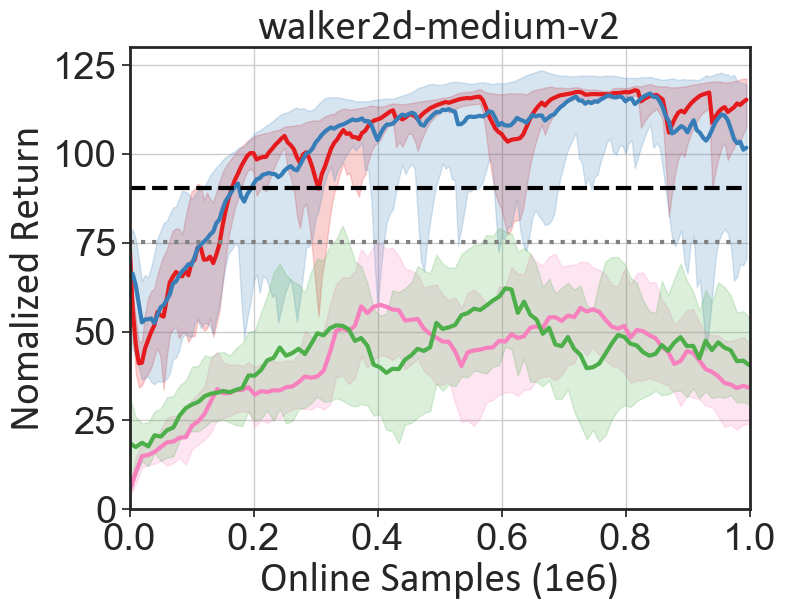}
    \includegraphics[width=0.24\textwidth]{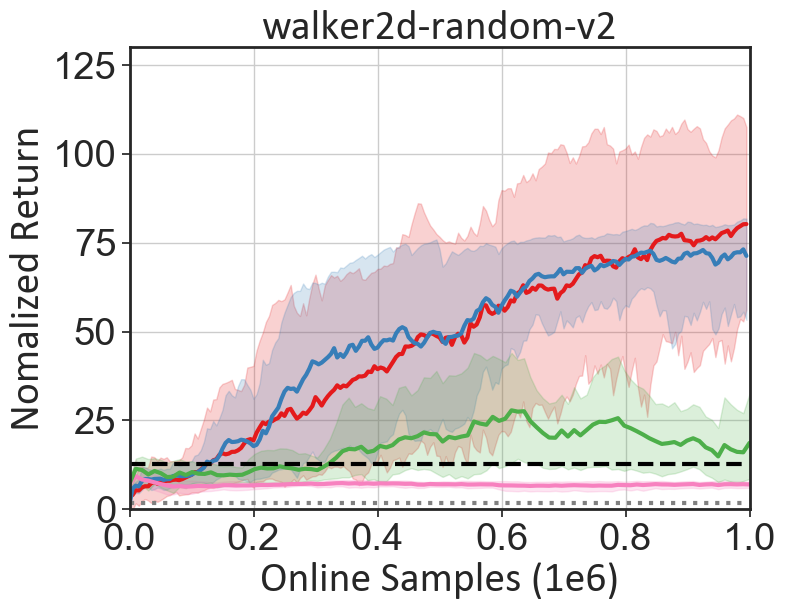}
    \caption{Full results of online finetuning with {{BC}} pretrained policy.}
    \label{fig:appendix_bc_pretrained}
\end{figure}

In this section, we present the complete results for \textit{PROTO} with BC pretraining, referred to as \textit{PROTO+BC}, to demonstrate the versatility of \textit{PROTO} for various offline pretraining approaches. The results are reported in Figure~\ref{fig:appendix_bc_pretrained_aggregated} and Figure~\ref{fig:appendix_bc_pretrained}. In the case of using BC for policy pretraining, we use Fitted Q evaluation (FQE)~\cite{le2019batch} to obtain the action-value $Q^0$ that corresponds to the BC policy. FQE is simple to train and insensitive to hyper-parameters, so pretraining $Q^0$ using FQE will not add additional parameter tuning burden or significant computational costs.

It has been observed in prior work~\cite{lee2022offline, nair2020awac} that directly initializing the value function with FQE and the policy with BC can lead to a significant performance drop during the initial finetuning stage. However, as shown in Figure~\ref{fig:appendix_bc_pretrained_aggregated} and Figure~\ref{fig:appendix_bc_pretrained}, this issue can be effectively addressed while achieving competitive finetuning performances using \textit{PROTO}. This suggests that \textit{PROTO} enables the use of the simplest pretraining method to construct state-of-the-art offline-to-online RL methods, bypassing the need for complex offline RL training.

\newpage
\section{Full Results of Online Finetuning with TD3}
\label{sec:td3_online}

In this section, we plug \textit{PROTO} into TD3~\cite{fujimoto2018addressing}, another SOTA online RL method that focuses on deterministic policy learning, dubbed as \textit{PROTO+TD3} to demonstrate the adaptability of \textit{PROTO} for diverse online finetuning approaches.
% and report the results in Figure~\ref{fig:appendix_td3_finetuning_aggragate} and Figure~\ref{fig:appendix_td3_finetuning}.
Differ from finetuning with SAC~\cite{haarnoja2018soft}, TD3 builds on top of deterministic policy and thus the log term in Eq.~(\ref{equ:PROTO_objective}) in not calculable. To solve this, we replace the KL-divergence regularization in Eq.~(\ref{equ:PROTO_objective}) with a MSE loss akin to~\cite{fujimoto2021minimalist}:

\begin{equation}
    \pi_{k+1}\leftarrow\arg\max_\pi\mathbb{E}\left[\sum_{t=0}^\infty\gamma^t\left(r(s_t,a_t)-\alpha\cdot \left({\pi(s_t)}-{\pi_k(s_t)}\right)^2\right)\right], k\in N,
    \label{equ:PROTO_TD3}
\end{equation}

then the policy evaluation operator and policy improvement step become:

\begin{equation}
    (\mathcal{T}^{\pi_k}_{\pi_{k-1}}Q)(s,a):=r(s,a)+\gamma\mathbb{E}_{s'\sim\mathcal{P}(\cdot|s,a)}\left[Q(s',\pi_k(s'))-\alpha\cdot \left({\pi_k(s')}-{\pi_{k-1}(s')}\right)^2\right], k\in N^+,
\end{equation}

\begin{equation}
    \pi_{k+1}\leftarrow\arg\max_{\pi} \left[Q^{\pi_k}(s,\pi(s))-\alpha\cdot \left({\pi(s)}-{\pi_k(s)}\right)^2\right], k\in N.
    \label{equ:td3_actor}
\end{equation}

\begin{figure}[h]
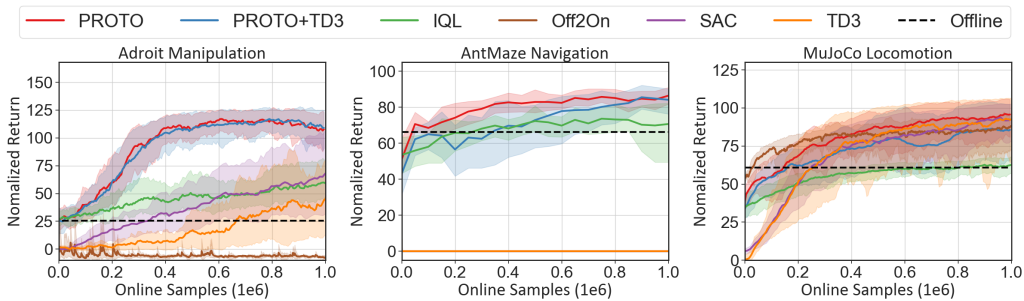

    \centering
    \includegraphics[width=0.95\textwidth]{Figure/Main/td3_finetune/legend_main_td3_finetun.png}
    
    \includegraphics[width=0.32\textwidth]{Figure/Main/td3_finetune/aggrated-return_adroit.png}
    \includegraphics[width=0.32\textwidth]{Figure/Main/td3_finetune/aggrated-return_antmaze.png}
    \includegraphics[width=0.32\textwidth]{Figure/Main/td3_finetune/aggrated-return_mujoco.png}
    \caption{{Aggregated learning curves} of {TD3} online finetuning.}
    \label{fig:appendix_td3_finetuning_aggragate}
\end{figure}

\begin{figure}[h]
\centering
    \includegraphics[width=0.95\textwidth]{Figure/Main/td3_finetune/legend_main_td3_finetun.png}

    \includegraphics[width=0.24\textwidth]{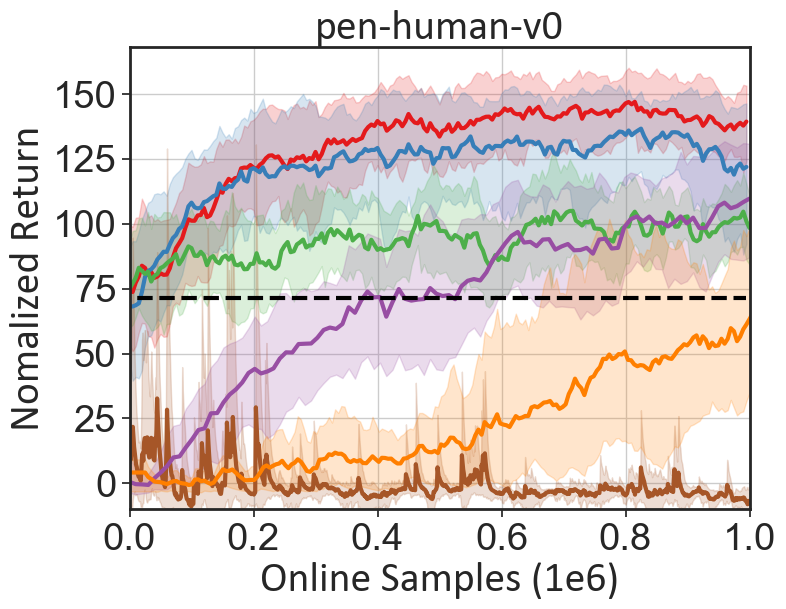}
    \includegraphics[width=0.24\textwidth]{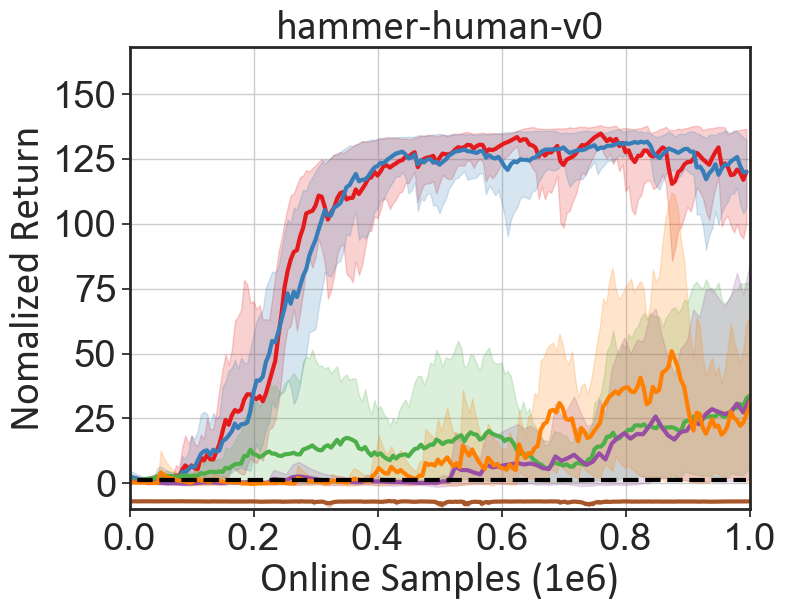}
    \includegraphics[width=0.24\textwidth]{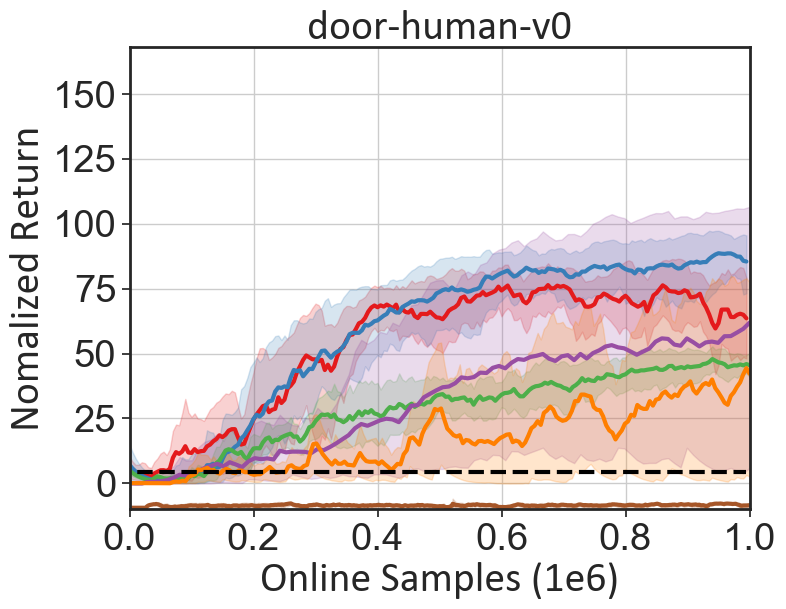}
    \includegraphics[width=0.24\textwidth]{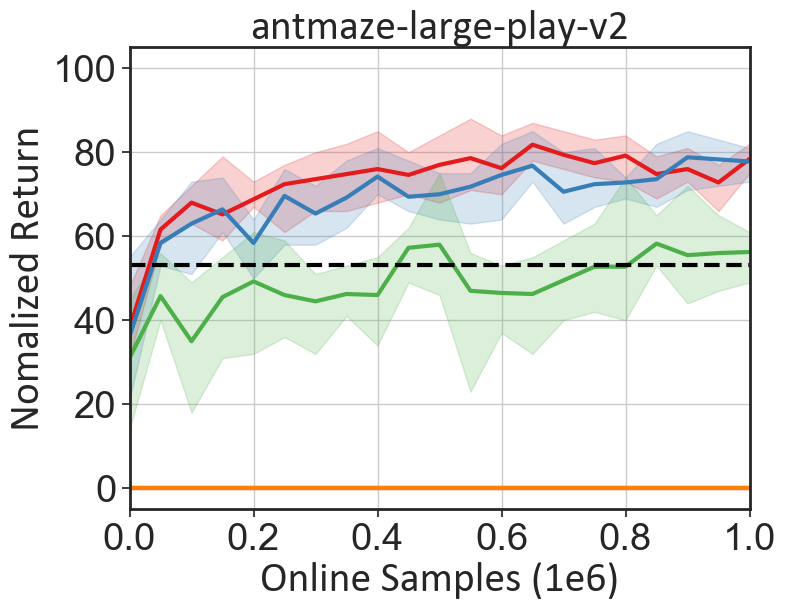}
    \includegraphics[width=0.24\textwidth]{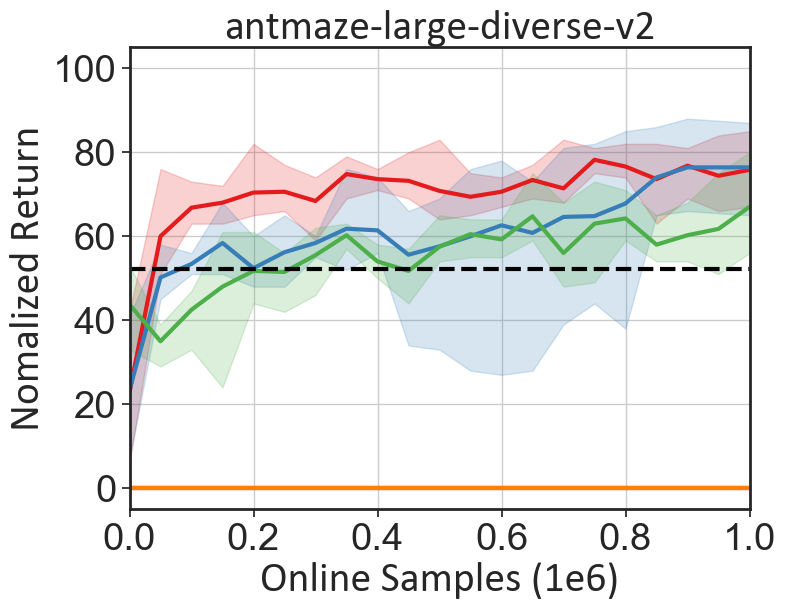}
    \includegraphics[width=0.24\textwidth]{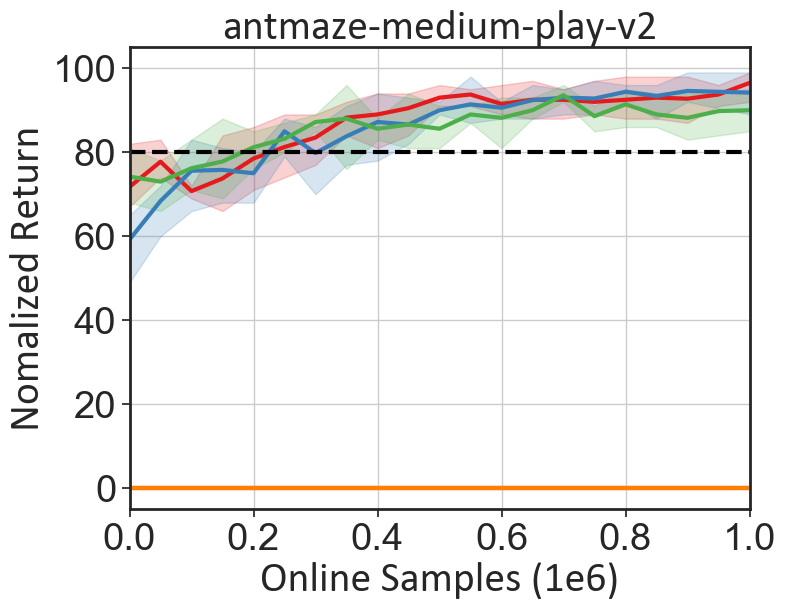}
    \includegraphics[width=0.24\textwidth]{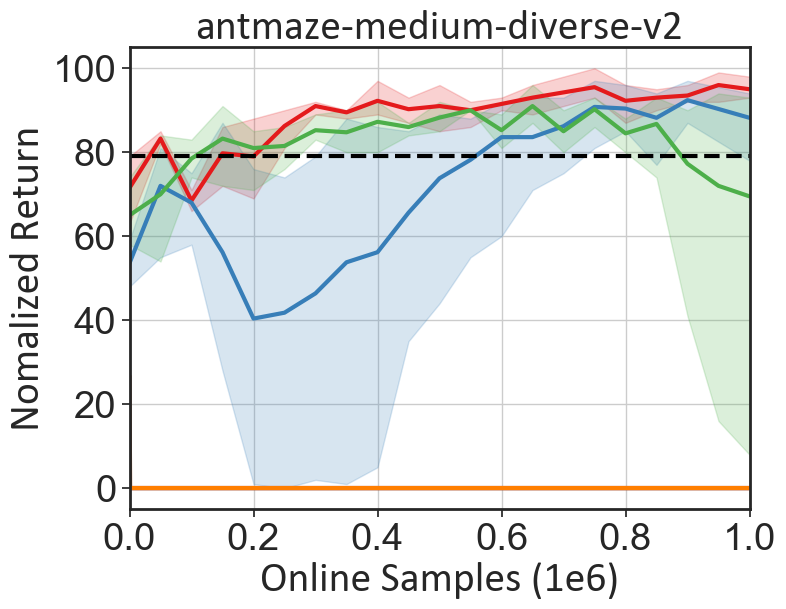}
    \includegraphics[width=0.24\textwidth]{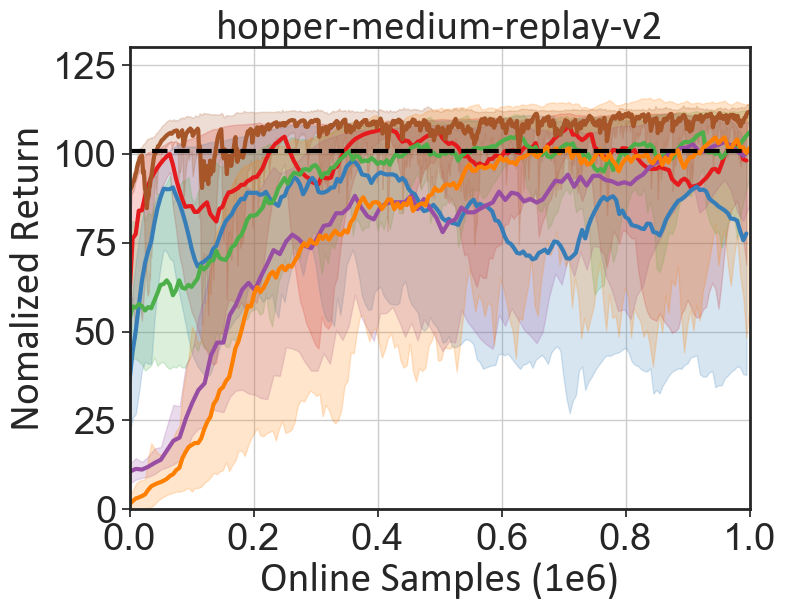}
    \includegraphics[width=0.24\textwidth]{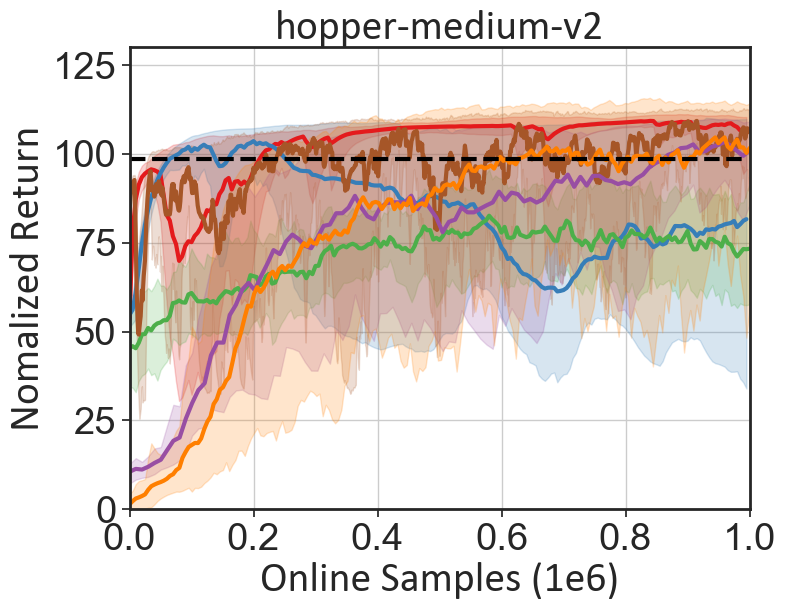}
    \includegraphics[width=0.24\textwidth]{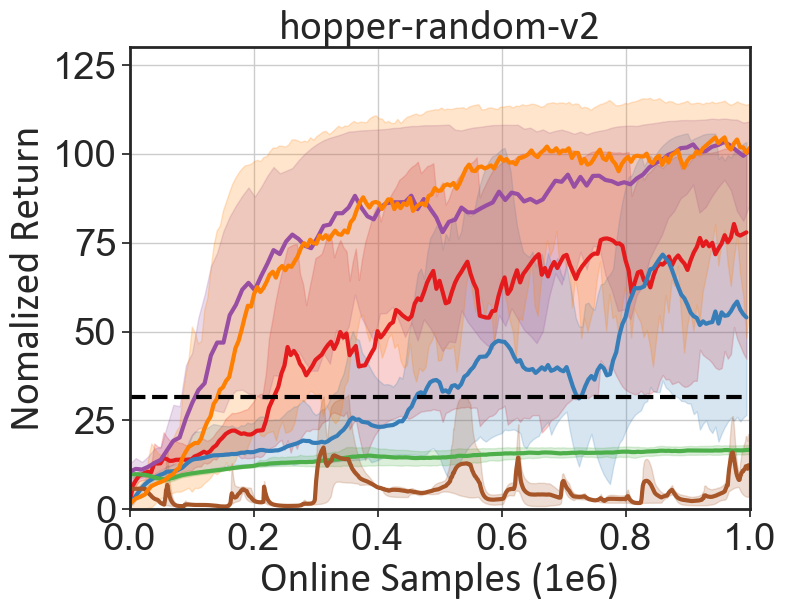}
    \includegraphics[width=0.24\textwidth]{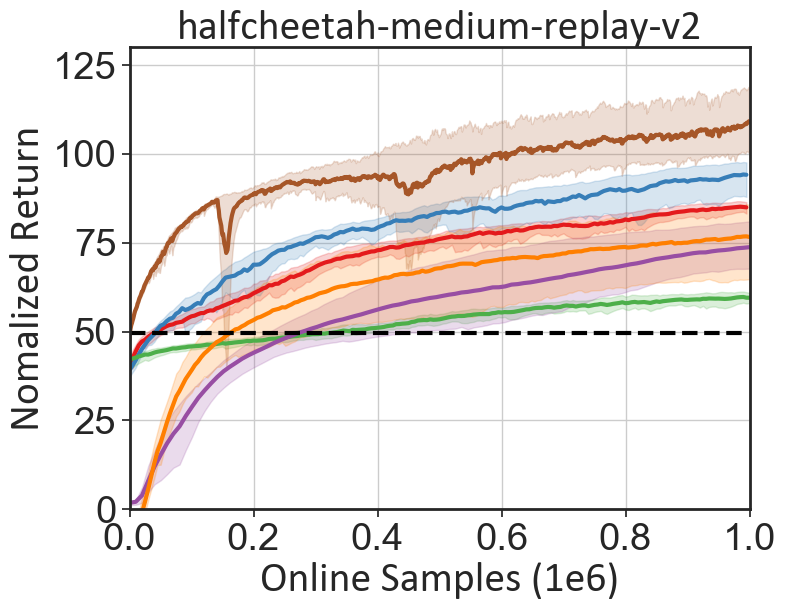}
    \includegraphics[width=0.24\textwidth]{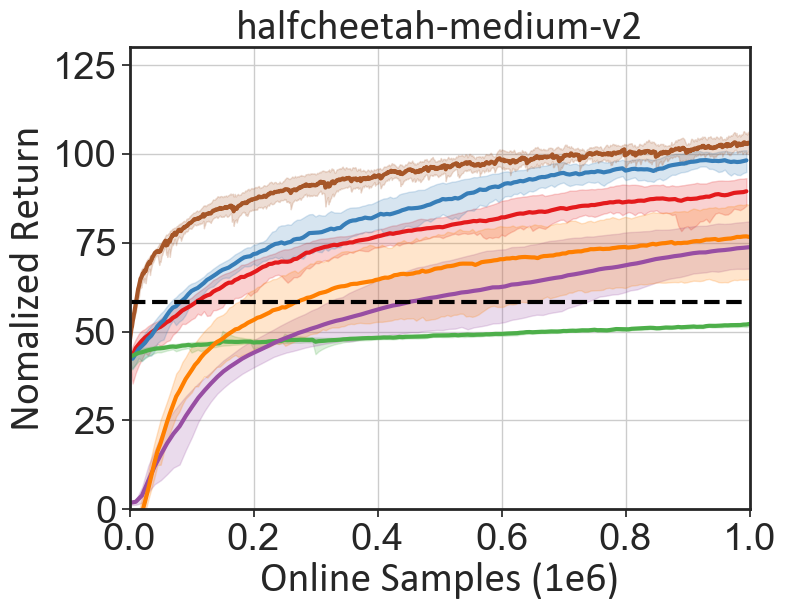}
    \includegraphics[width=0.24\textwidth]{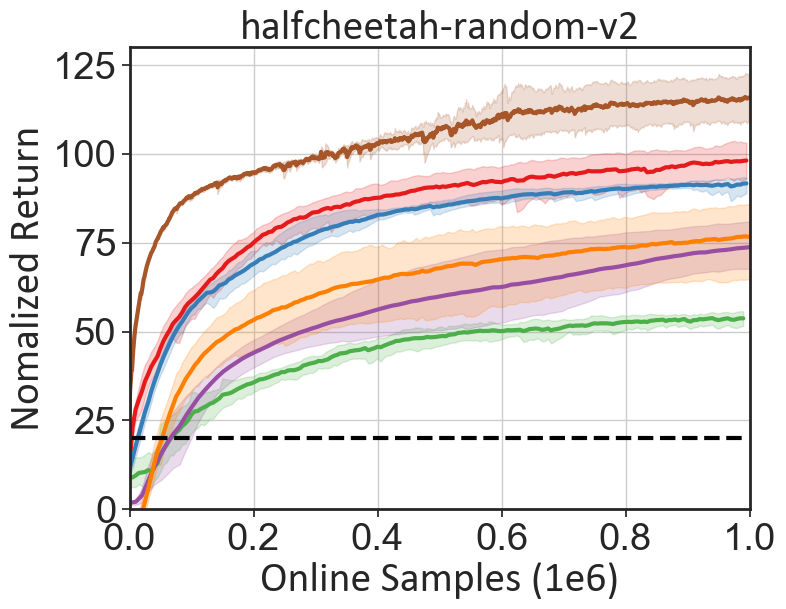}
    \includegraphics[width=0.24\textwidth]{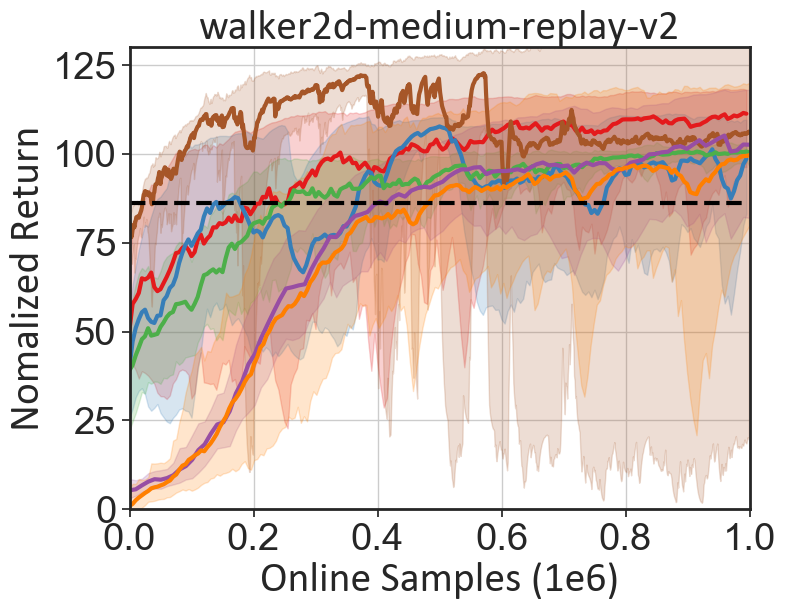}
    \includegraphics[width=0.24\textwidth]{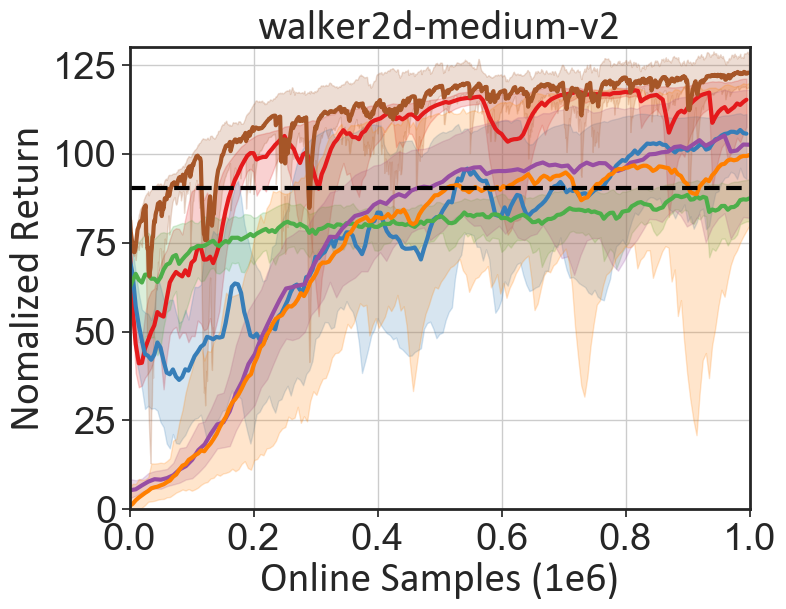}
    \includegraphics[width=0.24\textwidth]{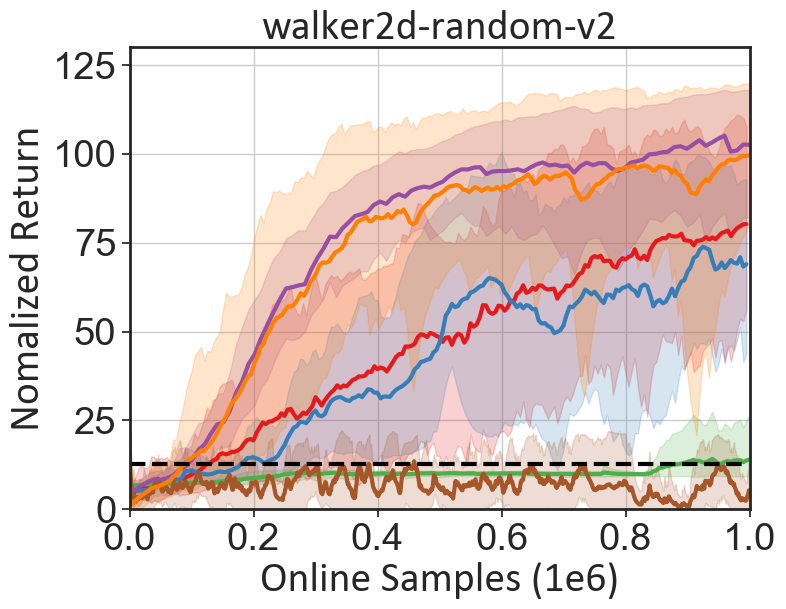}
    \caption{Full results of {TD3} online finetuning.}
    \label{fig:appendix_td3_finetuning}
\end{figure}

This objective shares the same philosophy of Eq.~(\ref{equ:PROTO_objective}) that constraining the finetuning policy \textit{w.r.t} an iteratively evolving policy $\pi_k$ instead of a fixed $\pi_0$, which degenerates to a recent related work~\cite{Yicheng2023fintuning}. However, it is worth mentioning that the constraint strength of the MSE loss in Eq.~(\ref{equ:td3_actor}) is far more weak than the log-barrier in Eq.~(\ref{equ:policy_trust}). For instance,assume the action space ranges from [-1, 1], then the MSE loss in Eq.~(\ref{equ:td3_actor}) is at most 4, which may vanish compared to the large action-value $Q$ during policy improvement Eq.~(\ref{equ:td3_actor}), while the log-barrier in Eq.~(\ref{equ:policy_trust}) may reach $\infty$. Therefore, we adopt the similar treatment in TD3+BC~\cite{fujimoto2021minimalist} that introducing a scalar $\lambda$ during policy improvement to rescale the action value to a comparable scale \textit{w.r.t} the MSE loss to stabilize training:

\begin{equation}
    \pi_{k+1}\leftarrow\arg\max_{\pi} \left[\lambda Q^{\pi_k}(s,\pi(s))-\alpha\cdot \left({\pi(s)}-{\pi_k(s)}\right)^2\right], k\in N,
\end{equation}

where

\begin{equation}
    \lambda = \frac{\beta}{\frac{1}{N}\sum_{(s_i,a_i)}|Q(s_i,a_i)|},
\end{equation}

where $N$ is batch size and $\beta$ is a hyper-parameter to control the Q scale. Although introducing one additional parameter to tune, we find that setting $\beta=4$ can achieve consistently good performance across 16 tasks. For the other parameter choice, we reuse almost all of the hyper-parameters from finetuning with SAC as reported in Appendix~\ref{sec:experimental_details} and only disable the clip-double Q trick for adroit Manipulation tasks. We report the aggregated learning curves and full results of \textit{PROTO+TD3} in Figure~\ref{fig:appendix_td3_finetuning_aggragate} and Figure~\ref{fig:appendix_td3_finetuning}. We also record the learning curves of training SAC and TD3 from scratch.


Figure~\ref{fig:appendix_td3_finetuning_aggragate} and Figure~\ref{fig:appendix_td3_finetuning} demonstrate the adaptability of \textit{PROTO} for diverse online finetuning approaches. By simply plugging \textit{PROTO} into TD3~\cite{fujimoto2018addressing}, we can form a competitive offline-to-online RL algorithm \textit{PROTO+TD3}, which also obtains SOTA performances compared to SOTA baselines. 

\newpage
\section{Full Results of Comparisons with Fixed Policy Regularization}

\begin{figure}[h]
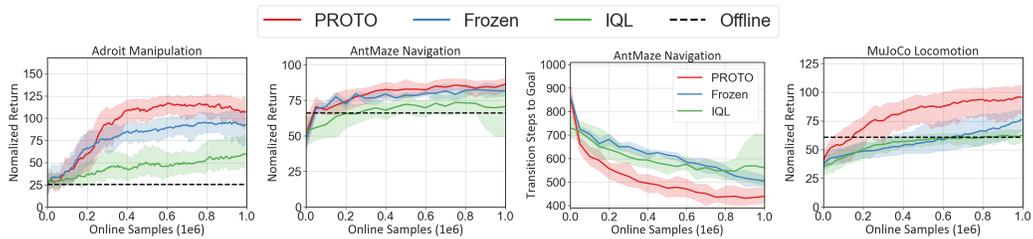

    \centering
    \includegraphics[width=0.5\textwidth]{Figure/Main/lc_frozen/legend_main_frozen.png}
    
    \includegraphics[width=0.24\textwidth]{Figure/Main/lc_frozen/aggrated-return_adroit.png}
    \includegraphics[width=0.24\textwidth]{Figure/Main/lc_frozen/aggrated-return_antmaze.png}
    \includegraphics[width=0.24\textwidth]{Figure/Main/lc_frozen/aggrated-return_antmaze_length.png}
    \includegraphics[width=0.24\textwidth]{Figure/Main/lc_frozen/aggrated-return_mujoco.png}
    \caption{{Aggregated learning curves} of iterative policy regularization ({{PROTO}}) \textit{v.s} fixed policy regularization ({{Frozen}}).}
    \label{fig:appendix_iterative_vs_fixed_aggregated}
\end{figure}

\begin{figure}[h]
\centering
    \includegraphics[width=0.5\textwidth]{Figure/Main/lc_frozen/legend_main_frozen.png}

    \includegraphics[width=0.24\textwidth]{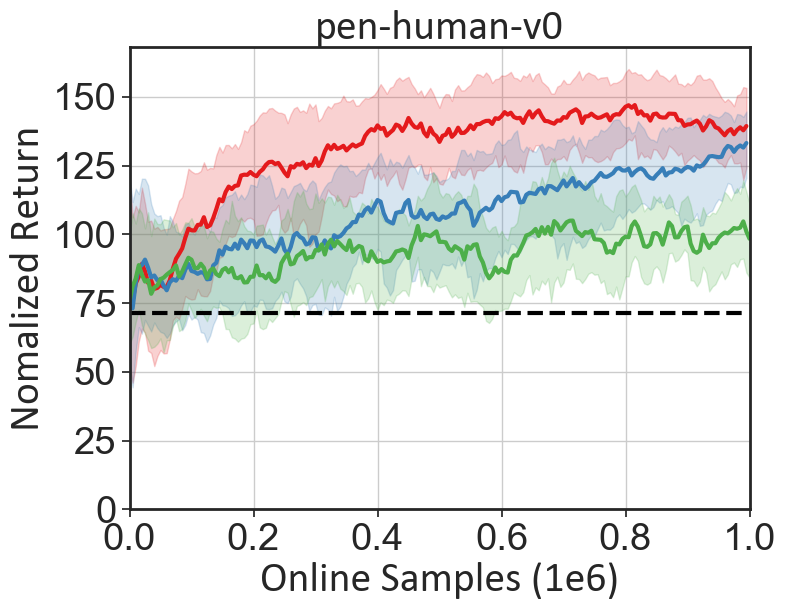}
    \includegraphics[width=0.24\textwidth]{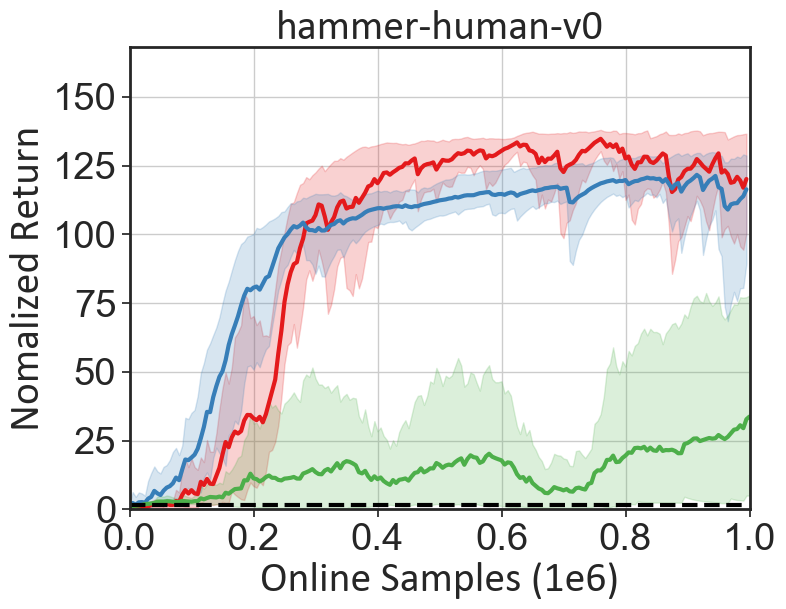}
    \includegraphics[width=0.24\textwidth]{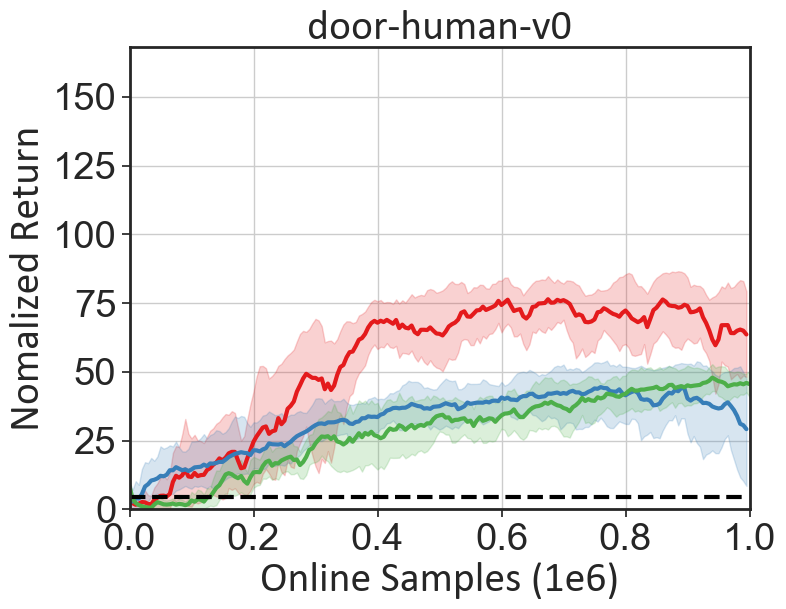}
    \includegraphics[width=0.24\textwidth]{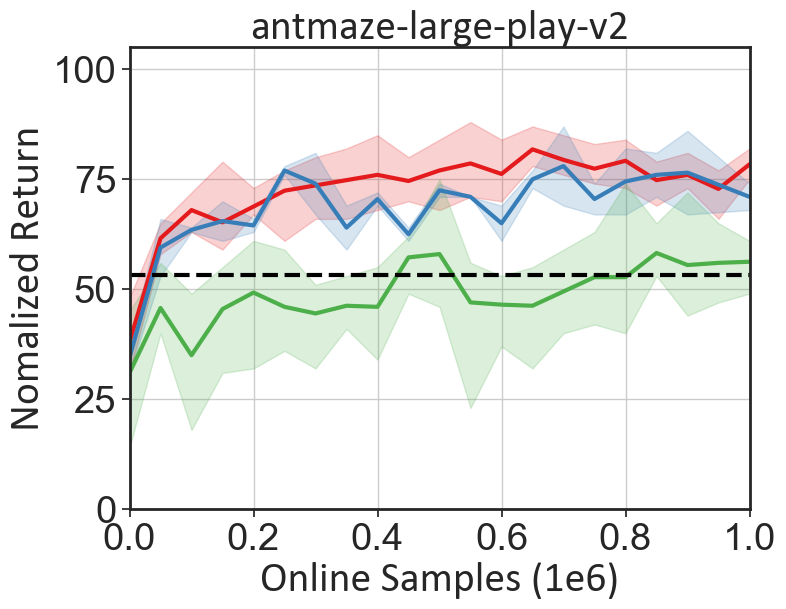}
    \includegraphics[width=0.24\textwidth]{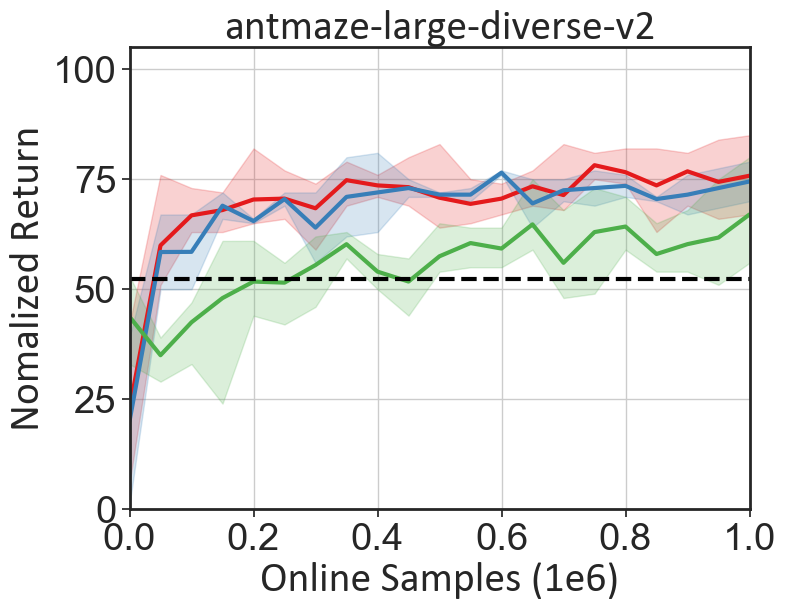}
    \includegraphics[width=0.24\textwidth]{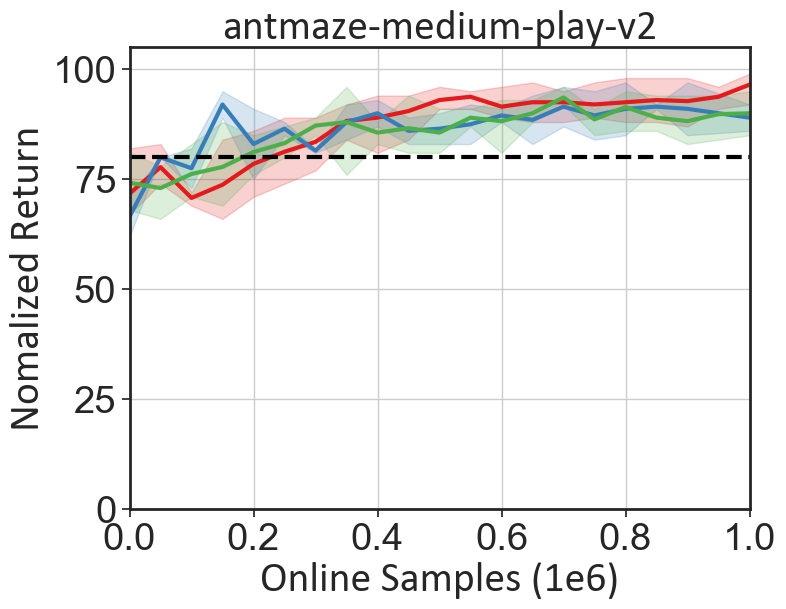}
    \includegraphics[width=0.24\textwidth]{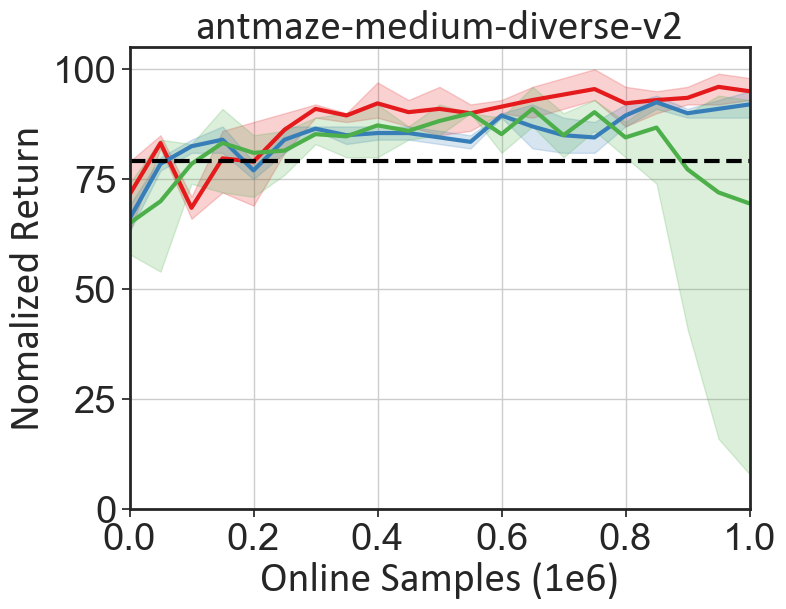}
    \includegraphics[width=0.24\textwidth]{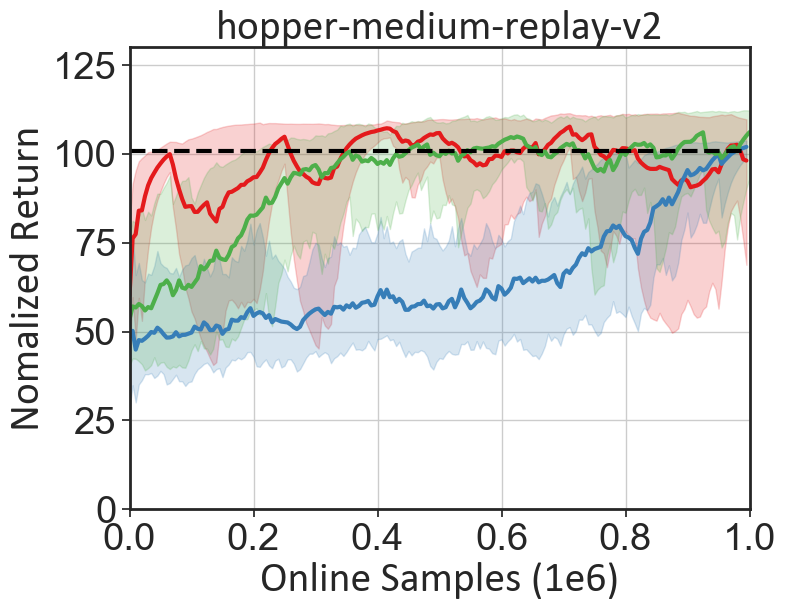}
    \includegraphics[width=0.24\textwidth]{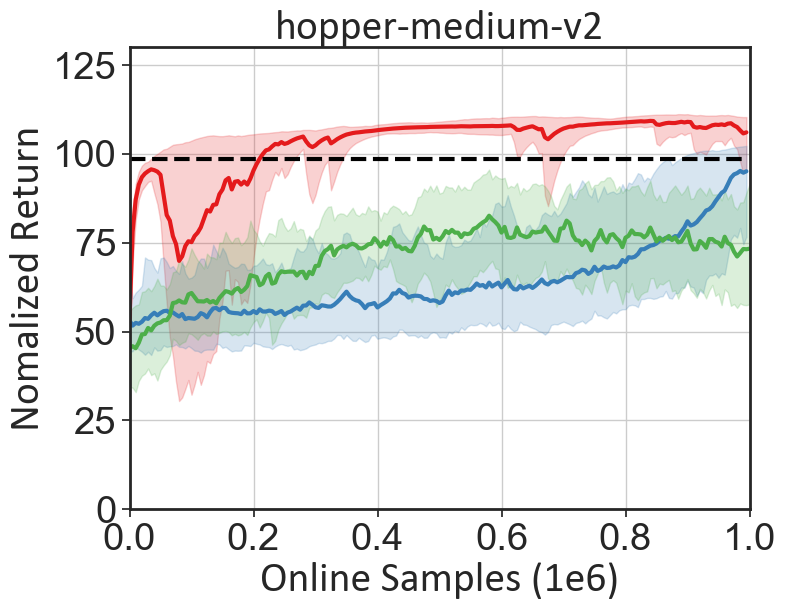}
    \includegraphics[width=0.24\textwidth]{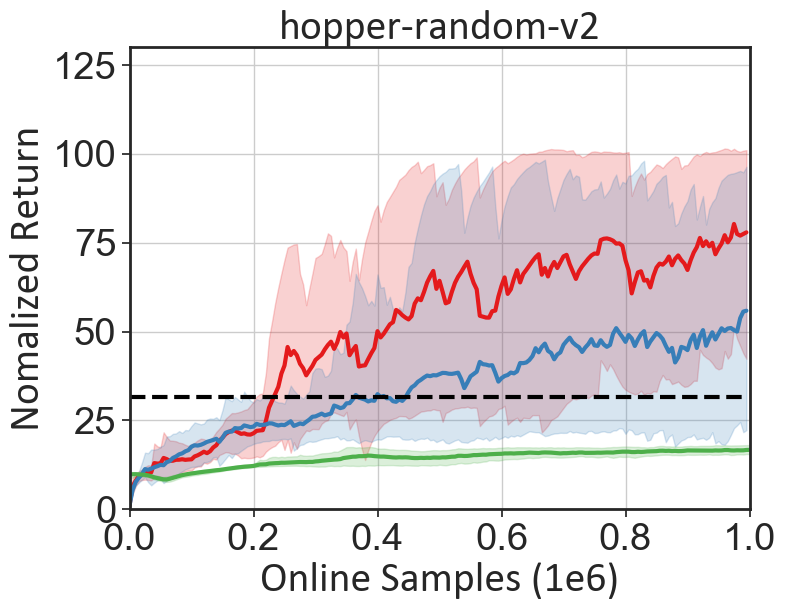}
    \includegraphics[width=0.24\textwidth]{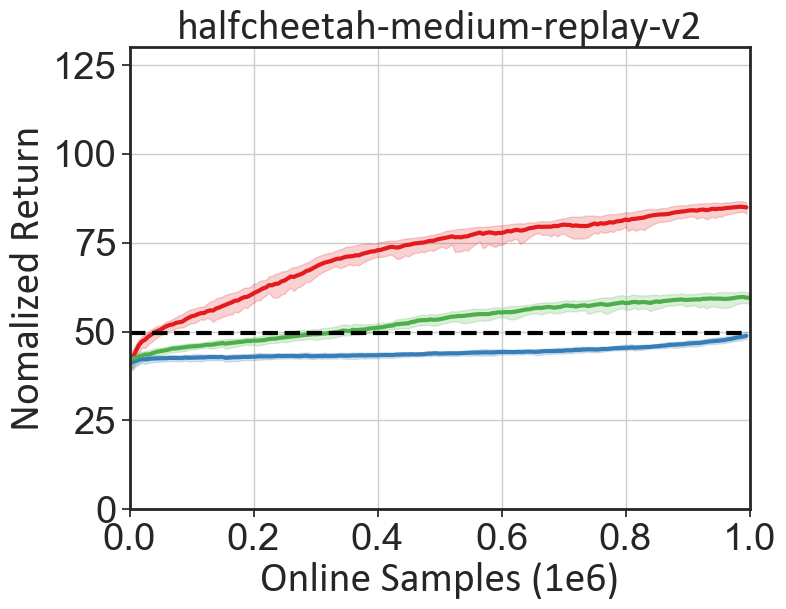}
    \includegraphics[width=0.24\textwidth]{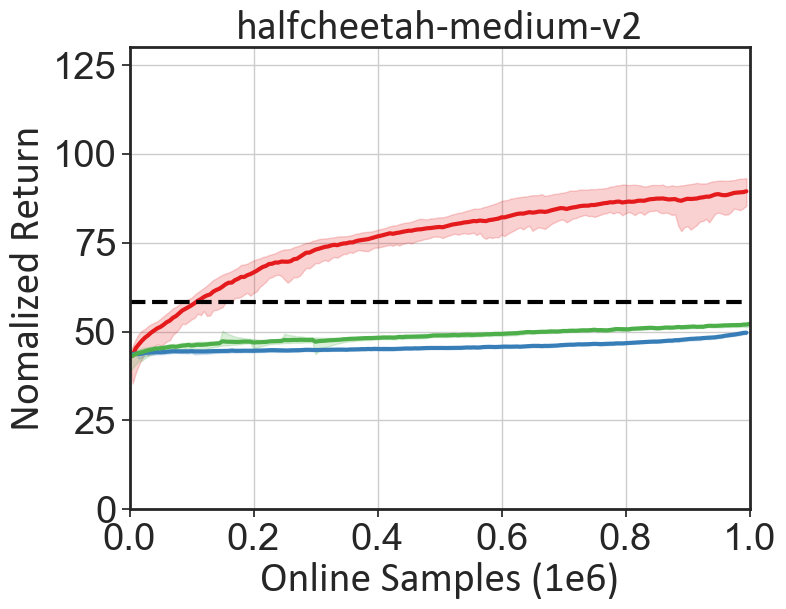}
    \includegraphics[width=0.24\textwidth]{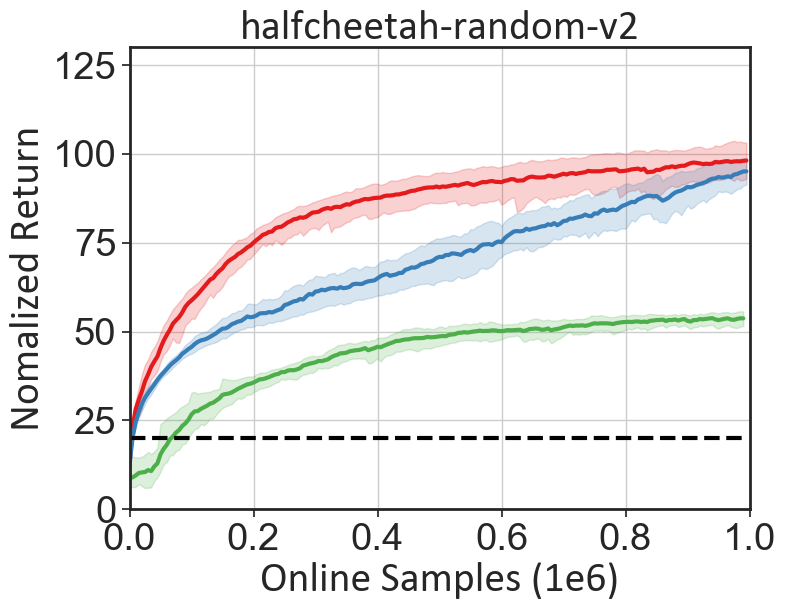}
    \includegraphics[width=0.24\textwidth]{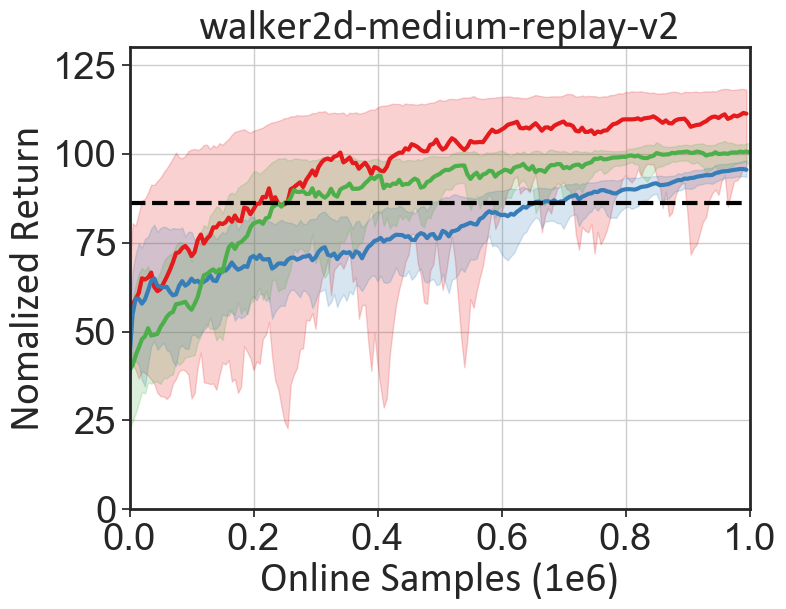}
    \includegraphics[width=0.24\textwidth]{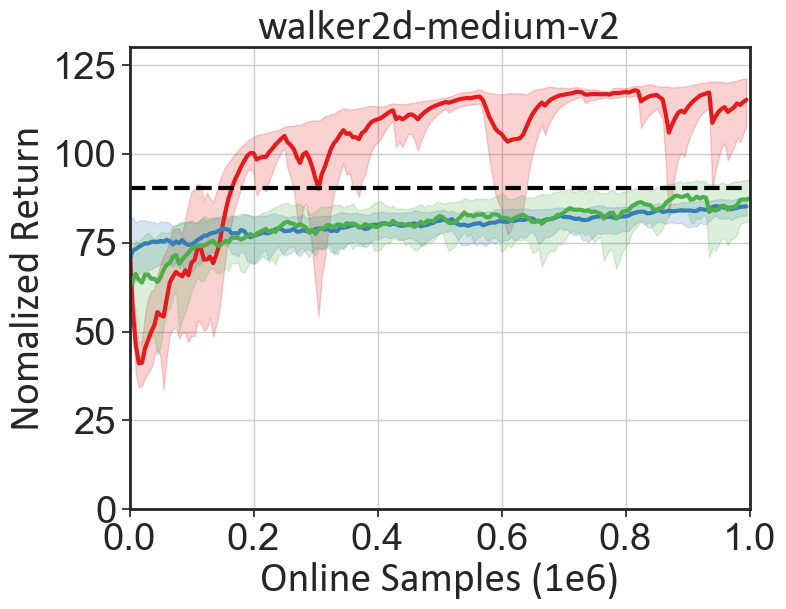}
    \includegraphics[width=0.24\textwidth]{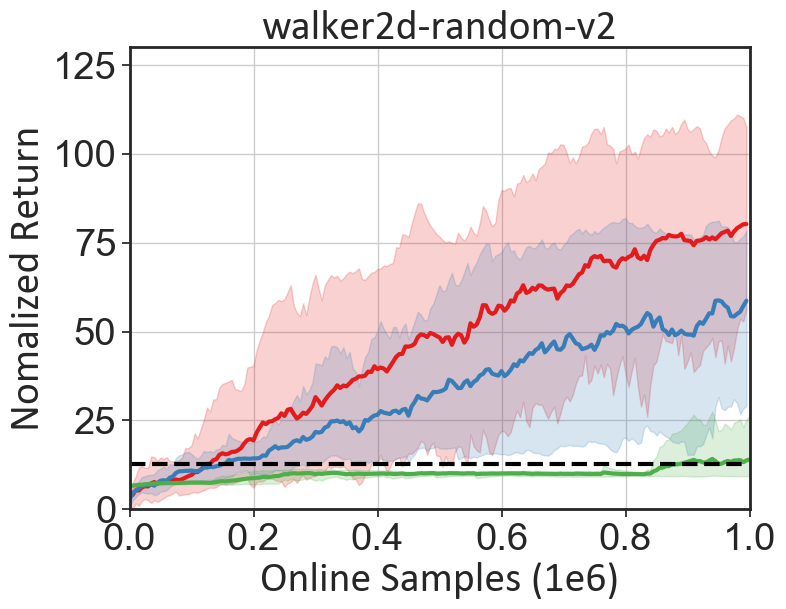}
    \includegraphics[width=0.24\textwidth]{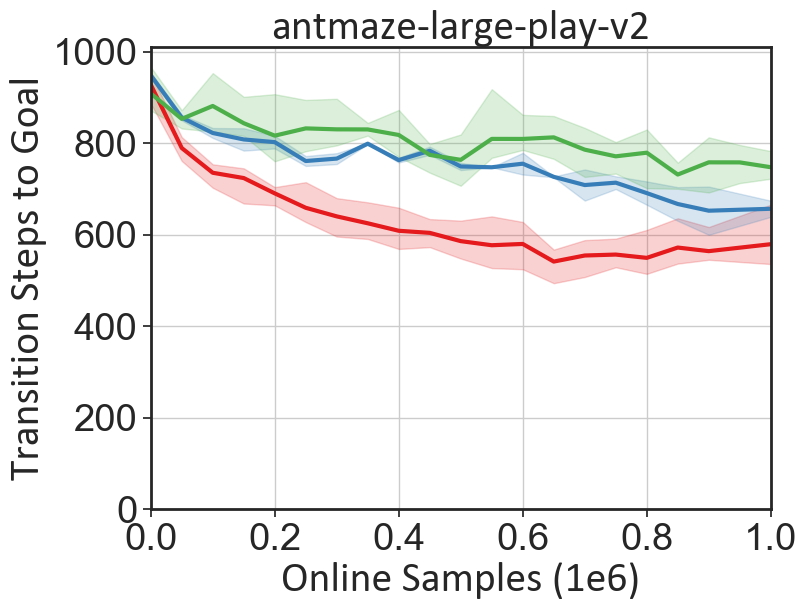}
    \includegraphics[width=0.24\textwidth]{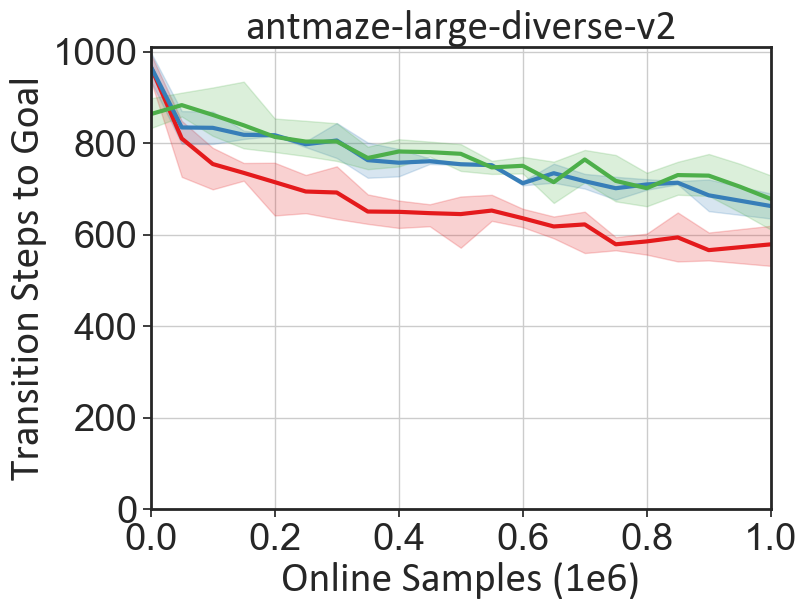}
    \includegraphics[width=0.24\textwidth]{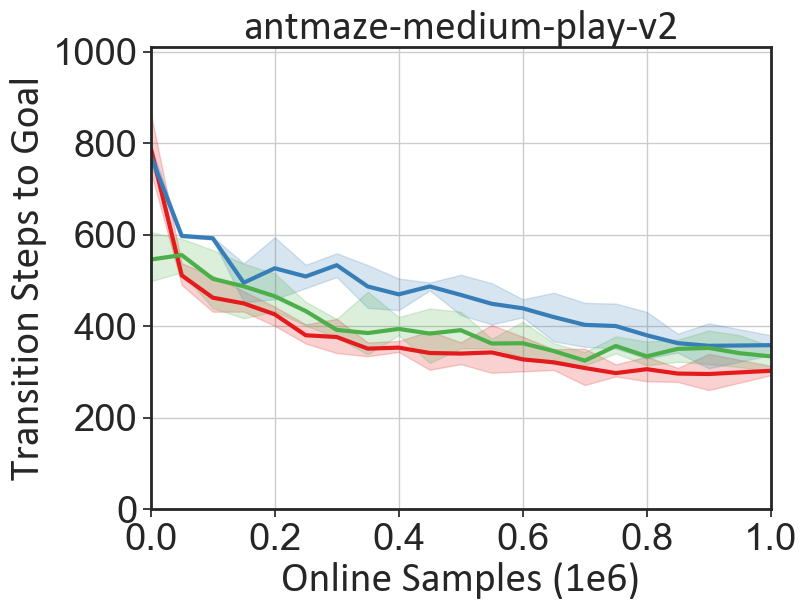}
    \includegraphics[width=0.24\textwidth]{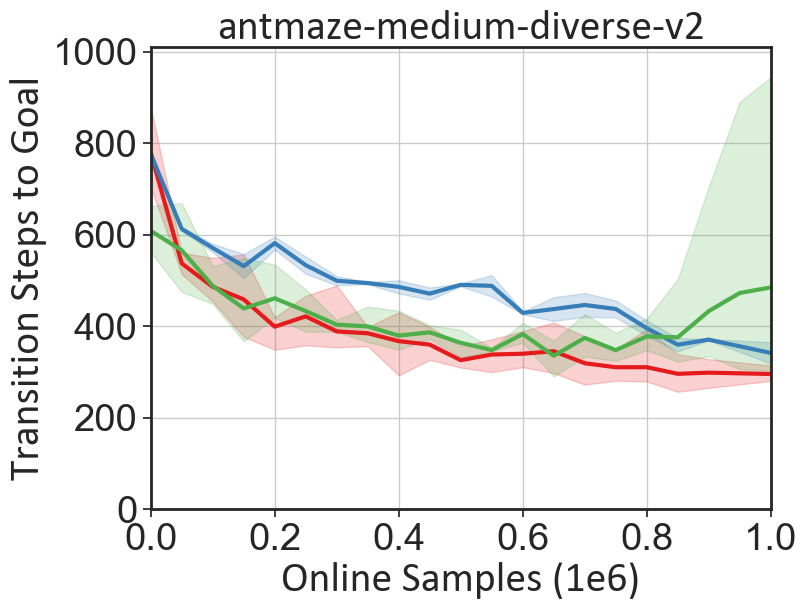}   
    \caption{Full results of iterative policy regularization ({{PROTO}}) \textit{v.s} fixed policy regularization ({{Frozen}}).}
    \label{fig:appendix_iterative_vs_fixed}
\end{figure}

In this section, we report the aggregated learning curves and full results for the comparisons of \textit{iterative policy regularization} with \textit{fixed policy regularization} in Figure~\ref{fig:appendix_iterative_vs_fixed_aggregated} and Figure~\ref{fig:appendix_iterative_vs_fixed}.


Observe from Figure~\ref{fig:appendix_iterative_vs_fixed_aggregated} and Figure~\ref{fig:appendix_iterative_vs_fixed} that although we adopt a linear schedule to anneal the conservatism strength for \textit{Frozen}, \textit{Frozen} still suffers from severe slow online finetuning and poor sample efficiency caused by the initial over-conservatism induced by fixed policy regularization.

\end{document}